%% file: main.tex
\documentclass[twoside,11pt]{article}

\usepackage{jmlr2e}

\usepackage[utf8]{inputenc} 
\usepackage[T1]{fontenc}    
\usepackage{hyperref}       
\usepackage{url}            
\usepackage{booktabs}       
\usepackage{amsfonts}       
\usepackage{nicefrac}       
\usepackage{microtype}      
\usepackage{xcolor}         

\usepackage{verbatim}

\usepackage{subcaption}

\newcommand{\E}{\mathbb{E}}
\newcommand{\R}{\mathbb{R}}

\usepackage{amsmath}

\usepackage{algorithm}
\usepackage{algorithmic}

\usepackage{lastpage}

\ShortHeadings{Dynamic Estimation of Learning Rates Using a Non-Linear Autoregressive Model}{R. Okhrati}
\firstpageno{1}

\begin{document}
\title{Dynamic Estimation of Learning Rates Using a Non-Linear Autoregressive Model}

\author{\name Ramin Okhrati \email r.okhrati@ucl.ac.uk \\
       \addr 
       University College London\\
       London, UK
       }


\maketitle

\begin{abstract}

We introduce a new class of adaptive non-linear autoregressive (Nlar) models incorporating the concept of momentum, which dynamically estimate both the learning rates and momentum as the number of iterations increases. In our method, the growth of the gradients is controlled using a scaling (clipping) function, leading to stable convergence. Within this framework, we propose three distinct estimators for learning rates and provide theoretical proof of their convergence. We further demonstrate how these estimators underpin the development of effective Nlar optimizers. The performance of the proposed estimators and optimizers is rigorously evaluated through extensive experiments across several datasets and a reinforcement learning environment. The results highlight two key features of the Nlar optimizers: robust convergence despite variations in underlying parameters, including large initial learning rates, and strong adaptability with rapid convergence during the initial epochs.
\end{abstract}


\input{sections/introduction}
\input{sections/notation}
\input{sections/nlar_aglorithms}

\input{sections/experimental_setup}
\input{sections/results_discussions}
\input{sections/conclusions_future_work}

\input{sections/declaration}

{
\small

\vskip 0.2in
\bibliography{references}  
}


\input{sections/appendix}

\end{document}

%% file: sections/introduction.tex
\section{Introduction}

Estimating a local optimum  is at the heart of any machine learning (ML) algorithm. This is typically achieved through iterative processes that require the gradient of an objective function and a mechanism to control the step size in updating the parameters. In Newton’s method, a reliable approach for approximating a local optimum, this control is implemented through a learning rate and the inverse of the Hessian matrix. However, calculating the Hessian matrix is computationally expensive, and becomes infeasible as the feature space dimension increases. As a result, gradient descent algorithms, which only require the first partial derivative of the objective function—hence also known as first-order methods—are preferred.

In a basic gradient descent method, model's parameters are updated at each iteration based on a learning rate parameter where the entire dataset is used to calculate the gradient and update the parameters. The use of the entire dataset to update the parameters is computationally intensive and may cause overfitting issues. Therefore, efficient methods have been developed that only use some randomly chosen batches of data. These methods are known as stochastic gradient descent (SGD) introduced by \citet{robbins1951stochastic}. Not only does SGD reduce the time complexity of gradient descent and accelerate convergence, but it also helps to mitigate the issue of getting stuck in local minima. The updating process continues until convergence occurs or a stopping criterion is met. 

 Both SGD and gradient descent in their basic forms suffer from oscillation, where they tend to oscillate around local minima instead of converging to them. A sophisticated and effective method to improve SGD is momentum \citep{rumelhart1986learning}, which introduces a decay parameter to penalize oscillations in successive iterations, thus leading to faster convergence. It is important to note that the impact of momentum is not uniform across dimensions; each dimension is affected differently, underscoring the importance of treating learning across dimensions individually. The momentum method can be further extended to Nesterov’s Accelerated Gradient (NAG) \citep{nesterov27method}, which also uses first-order information and can achieve better convergence rates in specific scenarios, such as smooth convex problems. One of the first papers applying this method in ML is by \citet{sutskever2013importance}.

Choosing the right learning rate is crucial for initiating these iterative approaches. Selecting a fixed learning rate throughout the iterative procedure is the simplest solution to the problem. This learning rate can be further tuned via cross validation approaches. Alternatively, one can adjust the learning rate as updating of the parameters progresses. This can be done either heuristically or using learning rate schedules  \citep{sutton1992gain, yu2005fast, krizhevsky2009learning, xu2017reinforcement, gotmare2018closer, xiong2022learning},  that may require fine-tuning of some hyperparameters. In the context of adaptive learning methods,  learning rates change and evolve as the gradient descent algorithm develops. This can be done either by adapting one single learning rate to all features such as NAG  or by personalizing training per dimension across the feature space. As mentioned earlier, the classic Newton method is one of the first adaptive learning approaches using the Hessian matrix, but its computational cost escalates with an increasing number of features. Adagrad \citep{duchi2011adaptive} is one of the pioneering adaptive ML algorithms that relies solely on first-order information.

In Adagrad, the learning rate decreases as the number of iterations increases, potentially approaching zero. Additionally, Adagrad is sensitive to the choice of initial parameters. Adagrad has been extended in two notable ways. The first extension is Root Mean Square Propagation (RMSProp) \citep{tieleman2012rmsprop}, which, instead of using the previous update, employs the previous root mean square of the gradients to scale the learning rate. The second extension is Adadelta \citep{zeiler2012adadelta}, which modifies Adagrad by using an exponentially decaying average of squared gradients rather than accumulating all past squared gradients. This adjustment addresses the issue of vanishing learning rates in Adagrad. Our estimators share structural similarities with Adagrad and Adadelta.

Moreover, the principles and characteristics of these methods can be combined to create new methodologies. For instance, the Adam optimizer \citep{kingma2014adam} can be viewed as a combination of Adagrad, RMSProp, and the momentum method. Adam is one of the most widely used optimizers in ML. It has been further extended in several ways, such as Nadam \citep{dozat2016incorporating}, which integrates Adam with Nesterov’s Accelerated Gradient (NAG). A key characteristic of most adaptive learning methods, including Adam, RMSProp, Adagrad, and Adadelta, is the per-dimension adjustment of learning rates, a feature that is absent in NAG. Other notable adaptive learning rate methods exist, such as the one proposed by \citet{agarwal2019efficient}, which is based on an efficient computation of the inverse square root of a low-rank matrix. While second-order information methods are available, they are not reviewed here since our approach focuses solely on first-order methods. 

We introduce a novel iterative method for estimating learning rates, grounded in non-linear autoregressive (Nlar) time-series models and robust mathematical principles. In our approach, we model the gradient descent iterations as a discrete time series, treating each step (or iteration) as a specific point in time. Given the non-linear nature of gradients, we employ a non-linear autoregressive time-series model. Once the basic gradient descent method is represented through this model, we apply the adaptive technique of \citet{aase1983recursive}, or as described by \citet{landau1976unbiased} and \citet{kushner1977convergence}, to dynamically estimate the learning rate. Our method relies on first-order partial derivatives, and under easily verifiable technical conditions, we provide convergence results for the learning rate estimators. Notably, these convergence results hold even if the model's loss function is non-convex.

Our approach leads to a new and innovative class of optimizers called Nlar optimizers. We begin with a general and flexible form of the Nlar optimizer, termed Nlarb, which facilitates the creation of new estimators. Specifically, we introduce two efficient variations of Nlarb, integrated with the concept of momentum, called Nlarcm and Nlarsm. In all of these algorithms, gradient growth is controlled via a function, providing theoretical support for using gradient clipping as a viable technique. Furthermore, Nlarcm and Nlarsm employ dynamic momentum estimation, rather than relying on a fixed value. We determine a specific form of these dynamic momentum values for which our experiments show that it performs quite well. The learning rates and momentum values adapt as convergence progresses, with momentum growth being controlled appropriately to ensure stable and steady convergence for any reasonable initial learning rate.

Nlarcm is based on five parameters, while Nlarsm uses six. However, compared to Nlarcm, Nlarsm requires simpler conditions and assumptions. Through sensitivity analysis, we demonstrate that, despite having multiple parameters to tune, the tuning of Nlarsm and Nlarcm can largely be reduced to finding optimal initial momentum and initial learning rate. Our experiments indicate that setting the initial momentum to one divided by the initial learning rate plus one performs adequately due to the dynamic adaptability of the momentum values. This simplifies the tuning process to primarily finding the optimal initial learning rate.

We compare our method with Adam, a prominent optimizer that has been extended to many others, and with one of its variations, AdamHD \citep{baydin}. While numerous extensions of Adam exist, such as AdaBelief \citep{zhuang2020adabelief} and $\text{Adam}^+$ \citep{liu2020adam}, we believe that a fair comparison for our proposed class of optimizers should primarily focus on the original Adam optimizer, which serves as the foundational concept for these variants. Including every extension would not only overwhelm the graphical comparisons but also shift the focus away from the core evaluation against the baseline idea, which is crucial for a clear and fair assessment of our contributions.  AdamHD was chosen as a benchmark due to its relative novelty and strong performance during the initial epochs, similar to our optimizers. We conduct a comprehensive set of experiments, including classification tasks and a reinforcement learning (RL) problem using Adam and AdamHD as benchmarks. These experiments confirm that both Nlarsm and Nlarcm demonstrate reliable convergence even with large initial learning rates up to 1. Furthermore, the experiments indicate that Nlarsm and Nlarcm exhibit strong initial adaptability, and the performance of Nlar algorithms generally surpasses the benchmarks, particularly during the early epochs.

The rest of paper is organized as follows. In Section \ref{sec:notation}, we introduce and provide some notation. The algorithms and theoretical backgrounds are provided in Section \ref{sec:Nlar}. Section \ref{sec:experimental setup} discusses the experimental setups. Section \ref{sec:discussion} is devoted to results and discussions. Finally, we conclude the paper in Section \ref{sec:conclusion}.

%% file: sections/notation.tex
\section{Notation}
\label{sec:notation}

Consider an ML model from  $\R^{n}$ to $\R$ where 
$n\geq1$ is a positive integer. Suppose that the model depends on some unknown trainable parameters (or weights), whose local optimal values are determined by minimizing a loss function $L=L_X$ on a dataset $X\subset\R^{n}$. It is assumed that $L$ is a function of the trainable parameters and admits first-order partial derivatives with respect to them. 

Corresponding to each trainable parameter, there is a unique tuple, also called dimension, that identifies the parameter. For instance, in a deep learning model, if $\Theta^{(1)}$ represents the matrix of weight parameters from layer $l$ (that includes a bias term)  to layer $l+1$, then each entry of the matrix, which is a parameter, is uniquely determined by the triplet $(l, i, j)$ for $0\leq i\leq n_r$ and $0\leq j\leq n_c$ where $n_r$ and $n_c$ are respectively the number of rows and columns of matrix $\Theta^{(1)}$. The set of all dimensions is denoted by $\mathcal D$. 

Estimating local optimal points of $L$ is normally done through iterative schemes.
For a dimension $d$ in $\mathcal D$, we use $\theta_{t}(d)$ to denote the estimated value of the local optimal point of $L$, along the dimension $d$ at iteration (or step) $t$. We also use $L_t$ to represent the value of the loss function at this iteration, which is based on $\theta_{t}(d)$ along all dimensions $d$. 

We recall that in its basic form, gradient descent estimates the local optimal points by a random initialization of parameters and then updating them through the following recursive formula (also known as the update rule) for all $t=0,1,2,\dots,$
\begin{equation}
\label{eq:gd_basic}
\theta_{t+1}(d) = \theta_{t}(d) - \gamma \dfrac{\partial L_t}{\partial \theta_{t}(d)},\text{ for all }d \text{ in } \mathcal D,    
\end{equation}
where $\dfrac{\partial L_t}{\partial \theta_{t}(d)}$ is the partial derivative of $L_t$ along the dimension $d$ evaluated on all the updated parameters at step $t$, and $\gamma>0$ is a constant learning rate.

The data or information\footnote{This is a loose term for a mathematical concept known as sigma-algebra.} generated by $(\theta_{s}(d))_{0\leq s\leq t}$ up until $t$, for all $d\in \mathcal D$ is denoted by $\mathfrak F_{t}$. When we say that a process is $\mathfrak{F} = (\mathfrak{F}_t)_{t \geq 0}$-adapted, we mean that the value of the process at time $t$ is fully determined by the information available up to that time. Additionally, in what follows, due to the random nature of our approach, all equalities, inequalities, and limiting behaviors involving random variables or stochastic processes should be interpreted as holding almost surely.

%% file: sections/nlar_aglorithms.tex
\section{Nlar estimators and algorithms}
\label{sec:Nlar}

As mentioned earlier, we treat gradient descent as a time-series model, where the iterations are interpreted as time. Furthermore, at each time step, some form of noise is injected into the system. An advantage of our estimator is that at each step, the growth of the partial derivatives is controlled through $f(\dfrac{\partial L_t}{\partial\theta_{t}(d)})$, where $f$ is a function. Specifically, for all $t=0,1,2,\dots$, the iterations of the gradient descent algorithm are described by the following stochastic process:
\begin{equation}
\label{eq:Nlar_main}
\theta_{t+1}(d) = \theta_{t}(d) - \gamma(d) f(\dfrac{\partial L_t}{\partial\theta_{t}(d)}) + \sigma_{t}(d) \epsilon_{t+1}(d),\text{ for all } d\text{ in }\mathcal D,
\end{equation}
where $L_t$ the loss function at step $t$, $f$ is any real-valued function, $\theta_{t}(d)$ is the parameter of the model to be estimated, $\gamma(d)$ is the learning rate,  
$\epsilon_{t+1}(d)$ is the injected noise, and $\sigma_{t}(d)$ is the magnitude of the noise. Equation \eqref{eq:Nlar_main} is basically a stochastic process that models the evolution of the  parameters. 

Note that Equation \eqref{eq:Nlar_main} can also be interpreted as the noisy observation of the basic gradient descent. Injecting noise into ML systems is a method that under the right circumstances can improve the performance of the model. For instance, \citet{neelakantan2015adding} use a specific normal distribution  to perturb the gradients, and they use fix learning rates of either 0.1 or 0.01. The noise $\epsilon_{t+1}(d)$ can follow any distribution, if part (b) of the following technical assumption is satisfied.
\begin{assumption}
\label{assump:epsilon}
\begin{itemize}
    \item[(a)] The iterations of the gradient descent are described by Equation \eqref{eq:Nlar_main}.
    \item[(b)] We assume that for all $t=0,1,2,\dots,$ random variables $\epsilon_{t+1}(d)$ are centered and have equal finite second moments (that is $\mathbb{E}[\epsilon_{t+1}(d)]=0$ and $e(d)=\mathbb{E}[\epsilon_{t+1}(d)^2]<\infty$, where $\mathbb E$ is the expected value and $e(d)$ is independent of $t$), and they are independent from  the information generated by all parameters along all dimensions up until $t$, i.e., $\mathfrak F_{t}$.
    \item[(c)] For all $d\in\mathcal D$ and $t=0,1,2,\dots,$ for other than a finite number of iterations, the partial derivatives of $L_t$ with respect trainable parameters are non-zero along dimension $d$.
\end{itemize}

\end{assumption}
Any type of centered noise with finite second moments (such as those following a normal or uniform distribution), generated independently at each iteration, satisfies part (b) of Assumption \ref{assump:epsilon}.

\begin{assumption}
\label{assump:grad}
The magnitude of the noise, $(\sigma_{t}(d))_{t\geq0}$, is defined as follows:
\begin{equation}\label{eq:sigmajs}
    \begin{aligned}
\sigma_{t}(d) = 
&\begin{cases}
\min(c_{t}(d), |f(\dfrac{\partial L_t}{\partial\theta_{t}(d)})|) & \text{if } \dfrac{\partial L_t}{\partial\theta_{t}(d)} \neq 0 \\
c_{t}(d) & \text{otherwise }  \\
\end{cases}
\end{aligned},
\end{equation}
where $(c_{t}(d))_{t\geq 0}$, $c_{t}(d)>0$ is an $\mathfrak F = (\mathfrak{F}_t)_{t\geq0}$-adapted process controls the scale of the noise.

For all $d$, and all $t=0,1,2, \dots,$ we assume that $|f(\dfrac{\partial L_t}{\partial\theta_{t}(d)})|\leq |b_{t}(d)|$, for an $\mathfrak F$-adapted process $(b_{t}(d))_{t\geq0}$  such that 
\begin{equation}
\label{eq:m}
    \sum_{s=1}^\infty \left(\dfrac{c_{s}(d)^{-1}b_{s}(d)}{s}\right)^2<\infty.
\end{equation}
\end{assumption}
Note that in Assumption \ref{assump:grad}, we do not require for the partial derivatives to be bounded directly, but for their mapping by $f$ to be bounded by the process $(b_t(d))_{t\geq0}$ satisfying \eqref{eq:m}. It is crucial that the values of $c_{t}(d)$ and hence $\sigma_t(d)$, are small enough not to destabilize the convergence. There are various setups that can satisfy Assumption \ref{assump:grad}. For example, we can take $b_t(d)=|f(\dfrac{\partial L_t}{\partial\theta_{t}(d)})|$, $c_t(d)=|b_t(d)|/(\min(|b_t(d)|, 10^{20})$ then $c_t(d)^{-1} |b_t(d)|=(\min(|b_t(d)|, 10^{20}) \leq 10^{20}$. This ensures that Equation \eqref{eq:m} is satisfied without imposing any restriction on $f$ or the partial derivatives. However, in our implementation, we have taken a simpler approach by considering a clipping function $f$ that puts an upper bound of $b_{t}(d)=b$ on $|f(\dfrac{\partial L_t}{\partial\theta_{t}(d)})|$, and $c_{t}(d)=c$, for constants $c>0$, $b>0$.

Now, we have all the ingredients to introduce the main estimator in its general form. The learning rate at step $t$ is estimated by:
\begin{equation}\label{eq:ktheta}
\hat\gamma_{t+1}(d) = \dfrac{k^{'}_t(d) - \sum_{s=0}^t \left((\sigma_{ s}(d))^{-2}f(\dfrac{\partial L_s}{\partial\theta_{s}(d)}) (\Delta_{s}^{(\theta)}(d))\right)}{k_t(d) +\sum_{s=0}^t \left((\sigma_{ s}(d))^{-1}f(\dfrac{\partial L_s}{\partial\theta_{s}(d)})\right)^2},
\end{equation}
where $\Delta_{t}^{(\theta)}(d)=\theta_{t+1}(d) - \theta_{t}(d)$, $(k^{'}_t(d))_{t\geq0}$ and $(k_t(d))_{t\geq0}$ can be any two $\mathfrak F = (\mathfrak{F}_t)_{t\geq0}$-adapted processes satisfying the following condition:
\begin{assumption}
\label{assump:kprime}
For all $t=0,1,2, \dots,$ we assume that $k_t(d) +\sum_{s=0}^t \left((\sigma_{s}(d))^{-1}f(\dfrac{\partial L_s}{\partial\theta_{s}(d)})\right)^2\neq0$, and 
\begin{equation}
    \label{eq:kprime}
  \lim_{t\rightarrow\infty} k^{'}_t(d)/t=\lim_{t\rightarrow\infty}k_t(d)/t=0.
\end{equation}
\end{assumption}
Assumption \ref{eq:kprime} is easily satisfied if $k_t(d)$ and $k^{'}_t(d)$ are constants and $k_t(d)> 0$. 

Before proving the convergence of the estimator \eqref{eq:ktheta}, a few points are worth noting regarding this estimator. Estimator \eqref{eq:ktheta} can be viewed as a generalization of Theorem 1 of \citet{aase1983recursive}. In fact, if $k_t(d) = k$ for a constant $k$; $k^{'}_t(d) = k \gamma_0(d)$, where $\gamma_0(d)>0$ is an initial value for the learning rate; $f(x)=x$ for all $x$, then estimator \eqref{eq:ktheta} would have the same form as that of \citet{aase1983recursive} though in a completely different context. Moreover, our assumptions are more tractable and much easier to verify and implement. 

In its general form, theoretically, the learning rates \eqref{eq:ktheta} can potentially take negative values in some dimensions or iterations. Negative learning rates are not uncommon in adaptive methods, for instance in the approach of \citep{baydin}.  However, in our experimental analysis, these negative learning rates did not cause any convergence issues or performance degradation. They appear to occur rarely and only in specific dimensions and iterations, without adversely affecting overall performance.  The following theorem which is inspired by Theorem 1 of \citet{aase1983recursive} and its proof is provided in Section \ref{proof:th-main} of the appendix,  guarantees the strong consistency of the estimator given by \eqref{eq:ktheta}.
\begin{theorem}
\label{th:main}
    Suppose that Assumptions \ref{assump:epsilon}, \ref{assump:grad}, and \ref{assump:kprime} hold. Then estimator $\hat\gamma_{t+1}(d)$ given by \eqref{eq:ktheta} converges to $\gamma(d)$ in \eqref{eq:Nlar_main}, as $t$ approaches infinity, i.e. $\lim_{t\rightarrow\infty}\hat\gamma_{t+1}(d) =  \gamma(d) $.
\end{theorem}

The estimator given by Equation \eqref{eq:ktheta} has certain similarity to Adagrad and Adadelta learning rate estimators respectively given by Equations \eqref{eq:adagrad} and  \eqref{eq:adadelta} in Section \ref{sec:app-overview} of the appendix, in the sense that the denominator of these estimators contains the accumulations of the square of the previous gradients. However, there are some key differences as well. First, the accumulation of the gradients is discounted by the magnitude of the noise at each step, and the whole sum is further controlled by $k_t(d)$, $k^{'}_t(d)$, and $f$. Second, note that in the numerator of Equation \eqref{eq:ktheta}, all the previous gradients and the updated parameters up until the last step are present. Unlike Adagrad for which the estimators gradually decrease, here, the behavior is completely different. 


\subsection{Nlarb: the basic form of the Nlar algorithm}

We present the first version of the Nlar algorithm in a general form, called Nlarb (``b'' stands for basic), which is based on Theorem \ref{th:main} and the estimator given by \eqref{eq:ktheta}. 
The details of this algorithm are provided in Algorithm \ref{alg:Nlarb} where $\sigma_{t}(d)$ (which is based on $c_{ t}$), $b_{t}(d)$, $k_t(d)$, and $k^{'}_{t}(d)$,   are the same as those in Equations \eqref{eq:sigmajs}, \eqref{eq:m}, \eqref{eq:ktheta}, and \eqref{eq:kprime}. Furthermore, $\epsilon_{t+1}(d)$ as in \eqref{eq:Nlar_main}, is the noise, and $N$ is the total number of iterations.

\begin{algorithm}[!t]
\tiny
    \caption{Nlarb}
    
    \textbf{Input:} $L$, $k_t(d)$, $k^{'}_t(d)$, $N$, $b_{t}(d)$, $c_{t}(d)$, $\lambda_0$, $\epsilon_{t+1}(d)$, for all  $t=0,1,2,\dots,N$ and $d$ in $\mathcal D$. The inputs must satisfy Assumptions \ref{assump:epsilon}, \ref{assump:grad}, and \ref{assump:kprime}.\\
     \textbf{Output:} estimated learning rates $\gamma(d)$ and parameters $\theta(d)$, for all $d\in\mathcal D$\\ 
    \vspace{-1em}
    \begin{algorithmic}[1]
        \STATE Initialise parameters: $\theta_{ 0}(d)$ for all $d\in\mathcal D,$
        \STATE $S_{0} (d)= 0$, $G_{0}(d)=0$, $\hat\gamma_0(d)=\lambda_0$, for all $d\in\mathcal D,$
        \FOR {$t=0,1,2,\dots,N$}
		\STATE compute $\dfrac{\partial L_t}{\partial\theta_{t}(d)}$ and $f(\dfrac{\partial L_t}{\partial\theta_{t}(d)})$, for all $d\in\mathcal D$,
            \STATE for all $d\in\mathcal D$, compute $\sigma_{ t}(d)$ using the following equation:
\begin{equation*}
    \begin{aligned}
\sigma_{t}(d) = 
\begin{cases}
\min(c_{t}(d), |f(\dfrac{\partial L_t}{\partial\theta_{t}(d)})|) & \text{if } \dfrac{\partial L_t}{\partial\theta_{t}(d)} \neq 0 \\
c_{t}(d) & \text{otherwise }  \\
\end{cases}
\end{aligned},
\end{equation*}
            \STATE for all $d\in\mathcal D$, update parameters:  
\begin{equation*}
\begin{aligned}
\theta_{t+1}(d) \gets \theta_{t}(d) - & \hat\gamma_{t}(d)f(\dfrac{\partial L_t}{\partial\theta_{t}(d)})                                                                     +  \sigma_{t}(d) \epsilon_{t+1}(d),
\end{aligned}
\end{equation*}
            \STATE Compute $\Delta^{(\theta)}_{t}(d)=\theta_{t+1}(d) - \theta_{t}(d)$, for all $d\in\mathcal D$,
			\STATE for all $d\in\mathcal D$: 
\begin{equation*}
\begin{aligned}
S_{t+1}(d) \gets  S_{t}(d)+(\sigma_{t}(d))^{-2} f(\dfrac{\partial L_t}{\partial\theta_{t}(d)}) (\Delta^{(\theta)}_{t}(d)), 
\end{aligned}
\end{equation*}
		  \STATE for all $d\in\mathcal D$: 
\begin{equation*}
\begin{aligned}
G_{t+1}(d) \gets  G_{t}(d)+(\sigma_{t}(d))^{-2}\left(f(\dfrac{\partial L_t}{\partial\theta_{t}(d)})\right)^2,
\end{aligned}
\end{equation*}
				\STATE $\hat\gamma_{t+1}(d) \gets \dfrac{k^{'}_{t}(d) - S_{t+1}(d)}{k_t(d)+G_{t+1}(d)}, $ for all $d\in\mathcal D$,
			\ENDFOR
					\STATE $\theta(d) \gets \theta_{N+1}(d) $ and $\gamma(d) \gets \hat\gamma_{N+1}(d)$, for all $d\in\mathcal D$
    \end{algorithmic}
    \label{alg:Nlarb}
\end{algorithm}

Algorithm \ref{alg:Nlarb} requires as inputs $L$, $k_t(d)$, $k^{'}_t(d)$, $N$ (number of iterations), $b_{t}(d)$, $c_{t}(d)$, $\lambda_0$ (the initial value of the learning rates), and $\epsilon_{t+1}(d)$. In comparison to a basic vanilla gradient descent algorithm, $k_t(d)$, $k^{'}_{t}$, $b_{t}(d)$,  $c_{t}(d)$, and $\epsilon_{t+1}(d)$ are the extra terms that we have control in choosing as long as Assumptions \ref{assump:epsilon}, \ref{assump:grad}, and \ref{assump:kprime}
 are satisfied. In Line 2, the main components of Nlarb are initialized. Note that all learning rates are initialized by $\lambda_0$, but each dimension could, in principle, have a different initialization. In Line 4, first the gradients of the model are computed, and then they are mapped using the function $f$. Line 5 calculates the magnitude of the noise based on the mapped gradients and $c_t(d)$. The parameters are updated in Line 6. In Line 7, the difference between the updated parameters and the previous ones is calculated and stored in $\Delta^{(\theta)}_{t}(d)$. Lines 8 and 9 respectively implement the numerator and denominator of Equation \eqref{eq:ktheta} apart from $k_t(d)$ and $k^{'}_t(d)$. Note that the algorithm directly gets feedback from the updated parameters. In Line 10, the learning rates are updated according to \eqref{eq:ktheta}, and then the process renews itself until the maximum number of iterations is reached. 

Note that based on Algorithm \ref{alg:Nlarb}, for any  model, corresponding to each trainable parameter of the model, there is also a learning rate. During the backpropagation process, as these parameters are updated, the learning rates are also adjusted in Line 10 of this algorithm. Algorithm \ref{alg:Nlarb} can be  thought of as an update rule where the convergence of the adaptive learning rates is guaranteed by Theorem \ref{th:main}. 

We present all the algorithms such as Algorithm \ref{alg:Nlarb} in a per-dimensional format. In practice, however, the updates are performed simultaneously, leveraging parallel processing capabilities. While Nlarb offers significant flexibility in constructing estimators, practical implementations require additional assumptions. Throughout the rest of this paper, we define $k^{'}_t(d) = k \lambda_{0}$ and $k_t(d) = k$, where $k$ is a constant and $\lambda_0$ is the initial value of the learning rates. In the following sections, we introduce two extensions of Nlarb that are suitable for practical implementation.


\subsection{Integration of the Nlar algorithm with momentum}

In this section, we discuss an integration of Nlar with the concept of momentum. Traditionally, in a vanilla SGD setting for a fixed index $d$ and momentum $\rho$, the velocity updates are given by
\[
v_{t+1}(d) = \rho v_{t}(d) - \gamma(d)\frac{\partial L_t}{\partial\theta_{t}(d)},
\]
and the parameter update rule is
\[
\theta_{t+1}(d) = \theta_{t}(d) + v_{t+1}(d).
\]
Here, the momentum $\rho$ is constant across all dimensions $d$ and time steps $t$. In contrast, our integration introduces a dynamic momentum that evolves along different dimensions and time periods. Specifically, for each time step $t=0,1,2,\dots$, we modify the update rule to accommodate this dynamic momentum as follows:
\begin{equation}
\label{eq:mu_update_rule}
\theta_{t+1}(d) = \theta_{t}(d) + \rho_t(d) v_t(d) - \gamma(d) f( \dfrac{\partial L_t}{\partial\theta_{t}(d)}) + \sigma_{t}(d) \epsilon_{t+1}(d),
\end{equation}
where $(\rho_t(d))_{t\geq0}$ is a non-negative process (i.e. $\rho_t(d)\geq0$ for all $t$ and $d$) that models the momentum,  $\sigma_{t}(d)$ is given by \eqref{eq:sigmajs}, and the velocities evolve based on the following dynamic:
\begin{equation}
    \label{eq:vels}
    v_{t+1}(d) = \rho_t(d) v_{t}(d) - \gamma(d) f(\dfrac{\partial L_t}{\partial\theta_{t}(d)}),\; t=0,1,2,\dots.
\end{equation}
The next theorem provides the convergence of the learning rates in this setup. See Section \ref{proof:th-momentum} of the appendix for the proof. 
\begin{theorem}
\label{th:momentum}
    Suppose that Assumptions \ref{assump:epsilon}, \ref{assump:grad}, and \ref{assump:kprime} hold, and for all $t$ and $d$, the update rule and velocities are respectively given by 
    \eqref{eq:mu_update_rule}, \eqref{eq:vels}, for a process $(\rho_t(d))_{t\geq0}$ that satisfies
    \begin{equation}
    \label{eq:cbd}
\sum_{s=1}^\infty\dfrac{  (\sigma_{s}(d))^{-2} |\rho_s(d)||v_s(d)||b_{s}(d)|}{s}<\infty.
    \end{equation}
Let $\Delta^{(\theta)}_{t}(d)=\theta_{t+1}(d) - \theta_{t}(d)$,    then for a constant $k> 0$,  the estimator 
    \begin{equation}
    \label{eq:ktheta3}
\hat{\zeta}_{t+1}(d) = \dfrac{{k}\hat{\zeta}_{0}(d) - \sum_{s=0}^t \left((\sigma_{ s}(d))^{-2}f(\dfrac{\partial L_s}{\partial\theta_{s}(d)}) (\Delta^{(\theta)}_{s}(d))\right)}{k +\sum_{s=0}^t \left((\sigma_{ s}(d))^{-1}f(\dfrac{\partial L_s}{\partial\theta_{s}(d)})\right)^2},
\end{equation}
 converges to $\gamma(d) $ as $t$ approaches infinity, i.e. $\lim_{t\rightarrow\infty}\hat{\zeta}_{t+1} =  \gamma(d) $.
\end{theorem}

\emph{Discussions on the assumptions of Theorem \ref{th:momentum}}. Equation \eqref{eq:cbd}, is the additional condition compared to Nlarb, and while it looks complicated, there are several setups for which the condition is easily satisfied.  We begin by setting  the constant stochastic processes $c_{t}(d)=c$ and $b_{t}(d)=b$. Although in Theorem \ref{th:momentum}, the values for the momentum are not restricted, in practical applications, they are usually kept within the range $[0,1]$ to ensure the stability and efficiency of the optimization process, which is the approach we adhere to. Specifically, we let $m_t(d)=c^{-2}\sigma_{t}(d)^2/(t+1)$, then condition \eqref{eq:cbd} is satisfied for  $\rho_t(d)=\rho{\min(|v_t(d)|, m_t(d))}/{|v_t(d)|}$ assuming $v_t(d)\neq0$, $\rho$ constant. Alternatively,  for $0\leq \rho\leq 1$, one can set 
\begin{equation}
\label{eq:rho_t}
m_t(d)=c^{-2}\frac{(\sigma_{t}(d))^2}{(t+1)},\qquad
    \rho_t(d)=\frac{\rho}{1+|\hat{\zeta_t}(d)|} \frac{m_t(d)}{(m_t(d) + |v_t(d)|)}.
\end{equation} 
The first fraction in $\rho_t(d)$ has a sort of reciprocal relationship with the learning rates, that is, for a given $d$, as the learning rate increases, the fraction decreases causing the momentum to adjust itself accordingly slowing down the pace of convergence. We use \eqref{eq:rho_t} in our numerical implementation, and as our experiments confirm, this form of $\rho_t(d)$ provides a stable and efficient convergence. Next, we develop a learning algorithm based on Theorem \ref{th:momentum}.

In Theorem \ref{th:momentum}, Equation \eqref{eq:ktheta3} is similar to Equation \eqref{eq:ktheta} of Theorem \ref{th:main}, where we have assumed that $k^{'}_t(d) = k \hat{\zeta}_0(d)$, $k_t(d)=k$, for a constant $k$. We impose $k>0$ to ensure that Assumption \ref{assump:kprime} holds, especially, this guarantees that the denominator in \eqref{eq:ktheta3} is non-zero.
To carry out numerical implementations based on Theorem \ref{th:momentum}, we need to make further simplifications as follows. In Equation \eqref{eq:sigmajs} that determines $\sigma_{t}(d)$, we let $c_{t}(d)=c$ and $f=f^\star$ where
\begin{equation}
\label{eq:global_clipping}
 f^{\star}(\dfrac{\partial L_t}{\partial\theta_{t}(d)})=b\dfrac{\dfrac{\partial L_t}{\partial\theta_{t}(d)}}{||\nabla_\Theta(L_t)||_2},
\end{equation}
$b$ is a constant ($|b|$ is called the global clipping norm), and $||\nabla_\Theta(L_t)||_2$ is the $L_2$-norm of the gradient of the loss function with respect to all the trainable variables, that is we flatten the set of all trainable variables in a one dimensional array, and then calculate the $L_2$-norm of the array. If we let $b_{t}(d)=b$, then  Assumption \eqref{assump:grad} is satisfied. Now, we have all the elements to define estimator Nlarcm (where ``c'' and ``m'' in Nlarcm respectively refer to constant and momentum),  formally presented in the following definition. 

\begin{definition}[\textbf{Nlarcm}]
\label{def:nlarcm}
Suppose that for all $d\in\mathcal D$, for other than a finite number of iterations, partial derivatives are non-zero along dimension $d$, and for all $t=0,1,2,\dots,$ random variables $\epsilon_{t+1}(d)$ are pairwise independent, independent from all $\mathfrak{F}_s$, $s\geq0$, and uniformly distributed with zero mean and variance of 1. Furthermore, we let $\hat{\zeta_0}(d)=\lambda_0$ for a positive constant $\lambda_0$, that is we apply the same initial learning rate across all dimensions.

Then the Nlarcm algorithm is  based on the estimator given by Equation  \eqref{eq:ktheta3}, for $m_t(d)=c^{-2}(\sigma_{t}(d))^2/(t+1)$, $c_{t}(d)=c$, $\rho_t(d)=\frac{\rho}{1+|\hat{\zeta}_t(d)|} m_t(d) / (m_t(d) + |v_t(d)|)$, and $b_{t}(d)=b$ where $c>0$, $0\leq \rho\leq 1$, $b$, $k>0$ are constants, and $f=f^{\star}$ is given by \eqref{eq:global_clipping}.
\end{definition}

Definition \ref{def:nlarcm} leads to the Nlarcm algorithm summarized in Algorithm \ref{alg:nlarcm} which satisfies the assumptions outlined in Theorem \ref{th:momentum}, and so the convergence of the learning rates are guaranteed. Note that $\rho$ and hence $\rho_t(d)$ can be zero. Algorithm \ref{alg:nlarcm} steps are similar to those of Algorithm \ref{alg:Nlarb}, but it is very important to note that in Line 7 of Nlarcm, we have used the velocity of the last step, i.e. $v_{t}(d)$ instead of $v_{t+1}(d)$.

\begin{algorithm}[!ht]
\tiny
    \caption{Nlarcm}
    \label{alg:nlarcm}
    \textbf{Input:} $L$, $k$, $N$, $b$, $c$, $\rho$, $\lambda_0$, $\epsilon_{t+1}(d)$, for all  $t=0,1,2,\dots,N$ and $d\in\mathcal D$. The inputs must satisfy Definition \ref{def:nlarcm}.\\
     \textbf{Output:} estimated learning rates $\gamma(d)$ and parameters $\theta(d)$, for all $d\in\mathcal D$\\
    \begin{algorithmic}[1]
        \STATE Initialize parameters $\theta_{ 0}(d)$, for all $d\in\mathcal D,$

        \STATE $S_{0}(d) =0$, $G_{0}(d)=0$, $v_{0}(d)=0$, $\hat\zeta_0(d)=\lambda_0$ for all $d\in\mathcal D,$
        \FOR {$t=0,1,2,\dots,N$}
    \STATE compute $\dfrac{\partial L_t}{\partial\theta_{t}(d)}$ and $f^{\star}(\dfrac{\partial L_t}{\partial\theta_{t}(d)})$, for all $d\in\mathcal D$, where $f^{\star}(\dfrac{\partial L_t}{\partial\theta_{t}(d)})$ is defined based on $b$ by \eqref{eq:global_clipping},
                        \STATE for all $d\in\mathcal D$, compute $\sigma_{ t}(d)$ using the following equation:
\begin{equation*}
    \begin{aligned}
\sigma_{t}(d) = 
\begin{cases}
\min(c, |f^{\star}(\dfrac{\partial L_t}{\partial\theta_{t}(d)})|) & \text{if } \dfrac{\partial L_t}{\partial\theta_{t}(d)} \neq 0 \\
c & \text{otherwise }  \\
\end{cases}
\end{aligned},
\end{equation*}
            \STATE for all $d\in\mathcal D$, update velocities using $m_t(d)=c^{-2}\sigma_{t}(d)^2/(t+1)$, $\rho_t(d)=\frac{\rho}{1+|\hat{\zeta}_t(d)|} m_t(d) / (m_t(d) + |v_t(d)|)$, and $$v_{t+1}(d) \gets \rho_t(d) v_{t}(d) - \hat\zeta_{t}(d)f^{\star}(\dfrac{\partial L_t}{\partial\theta_{t}(d)}),$$
            \STATE             for all $d\in\mathcal D$, update parameters:  
$$\theta_{t+1}(d) \gets \theta_{t}(d) + \rho_t(d)v_t(d) - \hat{\zeta}_{t}(d) f^{\star}(\dfrac{\partial L_t}{\partial\theta_{t}(d)}) + \sigma_{t}(d) \epsilon_{t+1}(d),$$
            \STATE $\Delta^{(\theta)}_{t}(d)=\theta_{t+1}(d) - \theta_{t}(d)$, for all $d\in\mathcal D$,
				\STATE for all $d\in\mathcal D$: $$S_{t+1}(d) \gets S_{t}(d)+(\sigma_{t}(d))^{-2} f^{\star}(\dfrac{\partial L_t}{\partial\theta_{t}(d)}) (\Delta^{(\theta)}_t(d)),$$
				\STATE for all $d\in\mathcal D$: $$G_{t+1}(d) \gets G_{t}(d)+(\sigma_{t}(d))^{-2}\left(f^{\star}(\dfrac{\partial L_t}{\partial\theta_{t}(d)})\right)^2,$$
				\STATE $\hat\zeta_{t+1}(d) \gets \dfrac{k\lambda_0 - S_{t+1}(d)}{k+G_{t+1}(d)},$ for all $d\in\mathcal D,$
		\ENDFOR
					\STATE $\theta(d) \gets \theta_{N+1}(d) $ and $\gamma(d) \gets \hat\zeta_{N+1}(d)$, for all $d\in\mathcal D$

    \end{algorithmic}
\end{algorithm}

Next, we obtain an alternative estimator of learning rates which is self-contained and requires fewer Assumptions leading to a different estimator. See Section \ref{proof:prop-Nlarsm} of the appendix for the proof. 

\begin{proposition}
\label{prop:Nlarsm}
Suppose that Assumption \ref{assump:epsilon} is satisfied, and there are positive constants $B$, $B^{'}$, and a function $f$, such that  $0<B^{'}<|f(\dfrac{\partial L_t}{\partial\theta_{t}(d)})|\leq B$. Let $\theta_{t+1}(d) - \theta_{t}(d)=\rho_t(d) v_{t}(d) - \gamma(d)  f(\dfrac{\partial L_t}{\partial\theta_{t}(d)}) + \sigma_{t}^{'}(d)\epsilon_{t+1}(d),\; t=0, 1, 2,\dots$, where $v_{t}(d)$ is given by \eqref{eq:vels}, $(\sigma_{t}^{'}(d))_{t\geq0}$ is a positive $\mathfrak F$-adapted processes, $(\rho_t(d))_{t\geq0}$ is an $\mathfrak F$-adapted processes  such that 
\begin{equation}
\label{eq:cbd2}
    \sum_{s=1}^\infty\E[(\sigma_s^{'}(d))^{2}]/s^2<\infty,\qquad \sum_{s=1}^\infty |\rho_s(d)v_{s}(d)|/s<\infty.
\end{equation} 
Then for a constant $k> 0$, the estimator
\begin{equation}
\label{eq:zetastar}
    \hat{\zeta}_{t+1}^\star(d) = \dfrac{k\hat{\zeta}_{0}^\star(d) - \sum_{s=0}^t \left(f(\dfrac{\partial L_s}{\partial\theta_{s}(d)}) (\Delta^{(\theta)}_{s}(d))\right)}{k +\sum_{s=0}^t \left(f(\dfrac{\partial L_s}{\partial\theta_{s}(d)})\right)^2},
\end{equation}
converges to $\gamma(d) $. 
\end{proposition}

\emph{Discussions on the assumptions of Proposition \ref{prop:Nlarsm}}. Note that Proposition \ref{prop:Nlarsm} is self-contained and apart from Assumption \ref{assump:epsilon}, no other external assumptions are required. If we let $m_t(d)=1/(t+1)$, $\rho_t(d)=\frac{\rho}{1+|\hat{\zeta}_t(d)|} m_t(d) / (m_t(d) + |v_t(d)|)$, and $\sigma^{\prime}_{t}(d)=c^{'}$, for a positive constant $c^{'}$ and $0\leq \rho\leq 1$, then $\sum_{s=0}^\infty\E[|\rho_s(d) v_{s}(d)|]/s<\infty$ and $\sum_{s=0}^\infty\E[(\sigma_s^{'}(d))^2]/s^2<\infty$, so the conditions of Proposition \ref{prop:Nlarsm} are satisfied. Note that $k>0$, ensures that \eqref{eq:zetastar} is well-defined. Also, the inequality $0<B^{'}<|f(\dfrac{\partial L_t}{\partial\theta_{t}(d)})|$ does not enforce for the gradients to be bounded from below, and it can be easily maintained by considering a clipping function $cl^{(B^\prime)}$ composed with $f^\star$ in \eqref{eq:global_clipping}: $f^\star$ globally scale the gradients, thereby bounding them from above, and the clipping function $cl^{(B^\prime)}$ further clips the absolute values of these truncated gradients, in an element-wise manner, to a small value, $B'$, when they fall below $B'$. Specifically,  for a scalar $x$, we have that
\begin{equation}
\label{eq:clbprime}
    cl^{(B^\prime)}(x) = x1_{\{|x|>B^\prime\}}+\text{sign}(x)B^\prime 1_{\{|x|<B^\prime\}},
\end{equation} 
where $\text{sign}(x)=1_{\{x\geq0\}}-1_{\{x<0\}}$. We use $B' = 10^{-150}$, which has been shown to perform well across the experiments we have conducted. 
 \begin{definition}[\textbf{Nlarsm}]
 \label{def:nlarsm}
     Suppose that for all $d\in\mathcal D$, for other than a finite number of iterations, partial derivatives are non-zero along dimension $d$, and for all $t=0,1,2,\dots,$ random variables $\epsilon_{t+1}(d)$ are pairwise independent, independent from all $\mathfrak{F}_s$, $s\geq0$, and uniformly distributed with zero mean and variance of one. Furthermore, we let $\hat{\zeta}_{0}^\star(d)=\lambda_0$ for a positive constant $\lambda_0$, that is we apply the same initial learning rate across all dimensions.

     Then the Nlarsm (where ``s'' and ``m'' respectively refer to ``simplified'' and ``momentum'') algorithm is based on  estimator $\hat{\zeta}_{t+1}^\star(d)$ given by Equation \eqref{eq:zetastar} for $k>0$, $m_t(d)=1/(t+1)$, $\rho_t(d)=\frac{\rho}{1+|\hat{\zeta}_{t}^\star(d)|} m_t(d) / (m_t(d) + |v_t(d)|)$, and $\sigma^{\prime}_{t}(d)=c^{'}$, where $0\leq \rho\leq 1$, $c^{'}$ is a positive constant, $f=cl^{(B^\prime)}\circ f^{\star}$, $cl^{(B^\prime)}$, given by \eqref{eq:clbprime} for a scalar input, is a clipping function that truncates the absolute value of its inputs element-wise from below by $B^\prime>0$,  and $f^\star$ is given by \eqref{eq:global_clipping} based on a constant $b$ such that  $|b|>B^\prime>0$.  
 \end{definition}

It is important to note that Nlarsm satisfies the assumptions outlined in Proposition \ref{prop:Nlarsm}, and other alternative distributions than uniform could be used. When compared to Nlarcm, Nlarsm admits a simpler form and requires fewer conditions and assumptions. Especially, the magnitude of the noise in Nlarsm (the constant $c^\prime$) is much simpler than Nlarcm ($\sigma_t(d)$ as in \eqref{eq:sigmajs}).  An algorithm based on Proposition \ref{prop:Nlarsm} can be developed which we call it, Nlarsm, presented in  Algorithms \ref{alg:nlarsm}. This algorithm follows similar steps to those in Algorithm \ref{alg:nlarcm}. The main differences lie in its conditions and simpler components, such as $m_t(d)$ and $\sigma^{\prime}_t(d)$, which make it less sensitive to floating-point precision.

 \begin{algorithm}[!ht]
 \tiny
    \caption{Nlarsm}\label{alg:nlarsm}
    \textbf{Input:} $L$, $N$, $\lambda_0$, $b$, $k$, $\rho$, $B^\prime$, $c^{\prime}$, $\epsilon_{t+1}(d)$, for all  $t=0,1,2,\dots,N$ and $d\in \mathcal D$. The inputs must satisfy Definition \ref{def:nlarsm}.\\
     \textbf{Output:}      estimated learning rates $\gamma(d)$ and parameters $\theta(d)$, for all $d\in\mathcal D$\\

    \begin{algorithmic}[1]
    \STATE Initialize parameters $\theta_{ 0}(d)$, for all $d\in\mathcal D,$
        \STATE $S_{0}(d) = 0$, $G_{0}(d)=0$, $v_0(d)=0$, $\hat\zeta^\star_{0}(d)=\lambda_0$, for all $d\in\mathcal D,$
        \FOR {$t=0,1,2,\dots,N$}
            \STATE compute $\dfrac{\partial L_t}{\partial\theta_{t}(d)}$ and $f(\dfrac{\partial L_t}{\partial\theta_{t}(d)})$, for all $d\in\mathcal D$, where $f=cl^{(B^\prime)}\circ f^{\star}$, $f^{\star}$ is defined based on $b$ by \eqref{eq:global_clipping},  and $cl^{(B^\prime)}$ is a clipping function given by \eqref{eq:clbprime} for a scalar input,
            \STATE for all $d\in\mathcal D$, update velocities using $m_t(d)=1/(t+1)$,  $\rho_t(d)=\frac{\rho}{1+|\hat\zeta^\star_{t}(d)|} m_t(d) / (m_t(d) + |v_t(d)|)$, and $$v_{t+1}(d) \gets \rho_t(d) v_{t}(d) - \hat\zeta^\star_{t}(d)f(\dfrac{\partial L_t}{\partial\theta_{t}(d)}),$$
            \STATE for all $d\in\mathcal D$, update parameters using:  $$\theta_{t+1}(d) \gets \theta_{t}(d) + \rho_t(d)v_t(d) - \hat\zeta^\star_t(d)  f(\dfrac{\partial L_t}{\partial\theta_{t}(d)}) + c^{\prime} \epsilon_{t+1}(d),$$

            \STATE $\Delta^{(\theta)}_{t}(d)=\theta_{t+1}(d) - \theta_{t}(d)$, for all $d\in\mathcal D$,
				\STATE $S_{t+1}(d) \gets S_{t}(d)+f(\dfrac{\partial L_t}{\partial\theta_{t}(d)}) (\Delta^{(\theta)}_{t}(d))$, for all $d\in\mathcal D,$
				\STATE $G_{t+1}(d) \gets G_{t}(d)+\left(f(\dfrac{\partial L_t}{\partial\theta_{t}(d)})\right)^2$, for all $d\in\mathcal D,$
				\STATE $\hat\zeta^\star_{t+1}(d) \gets \dfrac{k\lambda_0 - S_{t+1}(d)}{k+G_{t+1}(d)},$ for all $d\in\mathcal D,$
			
		\ENDFOR
	\STATE $\theta(d) \gets \theta_{N+1}(d) $ and $\gamma(d) \gets \hat\zeta^\star_{N+1}(d)$, for all $d\in\mathcal D$

    \end{algorithmic}
\end{algorithm}

\begin{remark}
\label{remark:nlars-nlarc}
There are a few important points worth of noting:
\begin{itemize}
\item[(a)] In both Theorem \ref{th:momentum} and Proposition \ref{prop:Nlarsm}, one can let $\rho_t(d)=0$  for all $t$ and $d$. In this case, Equations \eqref{eq:cbd} and \eqref{eq:cbd2} are satisfied, \eqref{eq:vels} becomes redundant, and the update rule \eqref{eq:mu_update_rule} simplifies to
$\theta_{t+1}(d) = \theta_{t}(d) - \gamma(d) f( \dfrac{\partial L_t}{\partial\theta_{t}(d)}) + \sigma_{t}(d) \epsilon_{t+1}(d).$ Assuming the same setups as Definitions of Nlarsm and Nlarcm, but with the extra condition $\rho_t(d)=0$, one can come up with two new optimizers called Nlars and Nlarc. Nlars and Nlarc could be also seen as special cases of Nlarb, and they provide a steady but slow convergence due to the absence of momentum. 
\item[(b)] Although we have proved the convergence of the learning rate estimators in Nlarb, Nlarcm, and Nlarsm, these algorithms for estimating the optimal weights $\theta_{t}(d)$ remain experimental. The convergence to the optimal weights depends not only on the quality of the learning rate estimators but also on the update rules, such as \eqref{eq:mu_update_rule}. Therefore, only an approximation of the local optimal points is achieved, and convergence to these points may not be guaranteed. Nonetheless, in all our experiments, the algorithms performed well, producing a decent approximation of the local optima and outperforming the benchmark methods in most cases.

\end{itemize}

\end{remark}

%% file: sections/experimental_setup.tex
\section{Experimental setup}
\label{sec:experimental setup}

We begin by explaining the framework and setup of our experiments. The experiments\footnote{The source code for the Nlar optimizers and experiments is available at \url{https://github.com/raminokhrati/Nlar}.} were conducted using the Keras API within TensorFlow. We utilized the built-in Adam optimizer from Keras and implemented AdamHD, Nlarsm, Nlarcm, and a double deep Q-learning network (DDQN) agent in TensorFlow Keras.

\paragraph{Datasets, environment, and preprocessing.} In our experiments, we use the datasets MNIST, CIFAR10, and the CartPole-v0 environment. The details of these datasets and the environment are provided in Section \ref{sec:app-data} of the appendix.  

\paragraph{Models.} In all the models, we have applied a bias term and an $L_2$ regularization term of 0.0001. We employed four models for classification problems. The first one is a Logistic regression model on MNIST. The second model is an MLP (Multi-Layer Perceptron) on MNIST with two hidden layers, as in \citet{kingma2014adam}, each with 1,000 nodes, ReLU activation functions for all nodes in the hidden layers, and softmax activation for the outer layer. We call this MLP2h. The third model is an MLP with seven hidden layers each made of 512 nodes with ReLU activation functions for all nodes in the hidden layers and softmax activation for the outer layer. We call this MLP7h. MLP7h is used in \citet{zhou2020towards} for a generalization performance comparison of Adam and SGD. The fourth model is a specific Visual Geometry Group (VGG\footnote{This is the name of the research group at the University of Oxford that developed the VGG models: \url{https://www.robots.ox.ac.uk/~vgg/research/very_deep/}.}) model,   \citep{simonyan2014very}. Our VGG model is made up 8 convolutional layers followed by three flattened hidden layers. We call this model VGG11, see Section \ref{sec:vgg11-CIFAR10} of the appendix for the component details. Although more complex models like VGG16 could be considered, VGG11 was chosen to reduce time complexity while still preserving the key characteristics of VGG networks.

For the RL part in our experiments, we use a DDQN as in \cite{van2016deep}. For both the target and the main network, we use an MLP with a single hidden layer of 128 neurons with a ReLU activation function at each node, and linear activation function on the last layer. We call this MLP model, MLP1h.

\paragraph{Sensitivity and parameter choice of Nlarsm and Nlarcm.} We continue by discussing the sensitivity of Nlarsm and Nlarcm with respect to their underlying parameters $c$, $c^\prime$, $k$, $b$, $\rho$, $B^\prime$, and $\lambda_0$ (the initial value of the learning rate). Despite the presence of multiple parameters, our analysis reveals that $\lambda_0$ is the primary factor influencing performance. Then, subsequent experiments demonstrate that the models exhibit robust and stable convergence across a wide range of $\lambda_0$ values, thereby simplifying the tuning process and enhancing overall usability.

In comparison, the sensitivity analysis of Adam has been explored in \citet{kingma2014adam} and further examined in subsequent work, such as \citet{zhuang2020adabelief} and \citet{yuan2020eadam}. Apart from $\lambda_0$, an important parameter of Adamh-HD is $\beta$, called hypergradient learning rate \citep{baydin}. From the experiments of Section \ref{sec:discussion}, out of the two choices $\beta=10^{-4}$ and $\beta=10^{-7}$ as suggested by the authors (for the VGG11 we also consider $\beta=10^{-8}$), we pick the $\beta$ with the best performance. As an example, comparing Figure \ref{fig:exp-mlp7h-CIFAR10-nlar-adam-adamhd-1e-4} (with $\beta=10^{-4}$ in Section \ref{sec:app-plots} of the appendix) and Figure \ref{fig:exp-mlp7h-CIFAR10-nlar-adam-adamhd-1e-7} illustrates the sensitivity of Adamh-HD with respect to the right choice of $\beta$. 

Nlarcm primarily depends on five parameters: $k$, $b$, $c$, $\rho$, and $\lambda_0$. In contrast, Nlarsm relies on $k$, $b$, $c^{\prime}$, $\rho$, $B^\prime$, and $\lambda_0$. Large values of $c$ and $c^\prime$ can cause instability in $\theta_{t+1}(d) - \theta_{t}(d)$ in Lines 7, 6 of Algorithms \ref{alg:nlarcm} and \ref{alg:nlarsm}, respectively. Figures \ref{fig:exp-mnist-mlp2h-nlar-sensitivity}, \ref{fig:exp-mnist-mlp7h-nlar-sensitivity}, and \ref{fig:sensitivity-four_figures} in Section \ref{sec:app-sensitivity} of the appendix show that the parameters $c$, $c^\prime$, and $B^\prime$ can be selected as small as possible as long as they do not cause overflow or underflow, for instance in the square of $1/\sigma_t(d)$. Also, in our assumptions, we have $k>0$, but these figures show that $|k|$ can be chosen as any non-zero value as long as it is not extremely large. Caution should be exercised with lower floating precision accuracies such as float32, as very small values of $c$, $c^\prime$ might cause instability. In general, Nlar algorithms are sensitive to small numbers, so setting the precision of floating-point numbers to 64 bits or higher is advised.

Nlarsm and Nlarcm are sensitive with respect to $b$ and $\rho$. The parameter $\rho$ would function similarly to momentum in SGD, providing additional flexibility. Our two experiments summarized in Figures \ref{fig:exp-mnist-nlars-nlarc-different-rho} and \ref{fig:exp-cifar10-nlars-nlarc-different-rho} in Section \ref{sec:app-sensitivity} of the appendix indicate that lower values of $\rho$ shall be applied with higher values of $\lambda_0$ and vice versa. Overall, our experiments show that setting $\rho=1$ performs adequately due to the dynamic adaptability of $\rho_t(d)$. Therefore, to minimize the number of parameters and reduce the optimizer's complexity, we set $\rho=1$. As for $b$, this parameter is essentially multiplied in the updated learning rates in Lines 6, 7 (respectively Lines 5, 6) of Algorithm \ref{alg:nlarcm} (respectively Algorithm \ref{alg:nlarsm}). Hence, it is practical to fix the standardized $b=1$ and search for an optimal $\lambda_0$. Therefore, while hyperparamter tuning could be carried out for Nlarsm and Nlarcm, as we have argued, one can choose $c=c^\prime=10^{-30}$ (for float32 or lower precision accuracy, we advise larger values of $c$ and $c^\prime$ such as $10^{-19}$), $k=1$, $b=1$, $\rho=1$,  $B^\prime=10^{-150}$,  and we have observed quite a robust and consistent performance across different values $\lambda_0$.

\paragraph{Training details.} 
Batch sizes of 300 and epochs of 50 are used in all classification problem experiments. For the RL problem, we consider 1,000 episodes, and for the gradient descent step of the DDQN algorithm, a batch size of 64 is applied. 

In all our experiments, a global clip norm of 1 is applied to all optimizers. For Nlarsm and Nlarcm, this means to let $b=1$ in \eqref{eq:global_clipping}. For classification tasks, we set the initial learning rates $\lambda_0$ across a range of  $[10^{-6}, 10^{-5}, 10^{-4},10^{-3},10^{-2},10^{-1},0.5, 1]$ for all optimizers, which include Adam, AdamHD, Nlarcm, and Nlarsm. To reduce the time complexity of the RL problem, we restrict the value of $\lambda_0$ to the range $[10^{-4}, 10^{-3}, 10^{-2}, 10^{-1}, 0.5, 1]$ for all optimizers.

For Adam, we adhere to the default settings: $\beta_1=0.9$, $\beta_2=0.999$, and $\epsilon=10^{-7}$, leaving the learning rate as the sole parameter to adjust. For AdamHD, besides the learning rate, there are four additional parameters to configure: $\beta_1$, $\beta_2$, $\epsilon$, and $\beta$. Following the recommendations in \cite{baydin}, we set $\beta_1=0.9$, $\beta_2=0.999$, $\epsilon=10^{-8}$, and choose $\beta$ values to be either $10^{-4}$ or $10^{-7}$ (for the VGG11 model, in addition, we also try $\beta=10^{-8}$). As it is discussed in sensitivity analysis, for the Nlar optimizers, we let $c=c^\prime=10^{-30}$,  $k=1$, $b=1$, $\rho=1$, and $B^\prime=10^{-150}$. 

For the training of each classification experiment, we have considered a distinct set of three random seeds. We use a three-fold cross-validation as follows: For each seed, we merge the default training and testing sets provided by the data provider, reshuffle them randomly, and then divide the datasets into training and test sets with the same sizes as the original datasets. We do not consider dropout in our models, and we use the entire training data to train the models, subsequently using the test dataset as a validation set. The final performance is calculated as the average over three seeds.

In our RL experiment, we use three distinct sequences of random seeds, with each sequence corresponding in length to the number of episodes, which is 1,000. This approach enhances the rigor of our method by preventing reliance on a single seed throughout the experiment, particularly during the exploration phase. Additionally, it ensures that all optimizers operate under the same random seed for each episode. Then, similar to the classification problems, we average over the three seeds. We use a discount rate of 0.95, an initial exploration rate of 1, an exploration decay of 0.995, a minimum exploration rate of 0.01, a maximum memory buffer size of 10,000, 1,000 episodes, a maximum of 200 steps per episode, and we update the main network model at each episode. The target model is updated every 20 episodes. The model is evaluated on each episode using a newly loaded environment for 10 episodes, and MLP1h is used for both the main and target networks with a batch size of 64. Replay starts only when the memory buffer size exceeds 1,000, allowing for initial explorations. At each step of an episode, the agent is rewarded after taking an action; however, if the game terminates before step 200, the agent is penalized by -1.

\begin{remark}
    For a fair comparison, in any given model and experiment, all optimizers use the same initialization of trainable parameters, the same data configurations, and the same batches, ensuring identical initial conditions and equal bases. 

\end{remark}


\subsection{Experiments}

In our experiments, we have focused on comparing Nlarsm and Nlarcm with Adam and AdamHD optimizers. However, we have provided the results of two experiments in Figures \ref{fig:exp-mlp-mnist-nlar-mu=none-adam-adamhd-1e-7} and \ref{fig:exp-mlp7h-CIFAR10-nlar-mu=none-adam-adamhd-1e-7} of the appendix using Nlars and Nlarc (see part (a) of Remark \ref{remark:nlars-nlarc} for definitions), for interested readers. Overall, Nlarsm and Nlarcm strike a better balance between a steady and fast convergence compared to Nlars and Nlarc.

As mentioned earlier, the experiments include classification tasks (Logistic regression, MLP2h on MNIST, and MLP7h, VGG11 on CIFAR10), and the DDQN RL model using MLP1h on CartPole-v0 environment. For the classification tasks, we graph the training and validation accuracy curves for different initial learning rate values and compare the performance of Nlarsm and Nlarcm with Adam and AdamHD. It is a common practice to use learning curves, such as training and validation losses, instead of accuracy metrics. However, this approach is not suitable in our case. In most of the experiments, the loss values for Adam and AdamHD are very large, especially for large initial learning rate values. This would overshadow the other curves in the figures, making comparisons difficult.
 
For the RL experiment, we measure performance using the cumulative moving average (CMA) of rewards. 

%% file: sections/results_discussions.tex
\section{Results and discussion}
\label{sec:discussion}

\begin{figure*}[!ht]
  \centering
  \includegraphics[width=0.22\linewidth, trim=0 10 30 0]{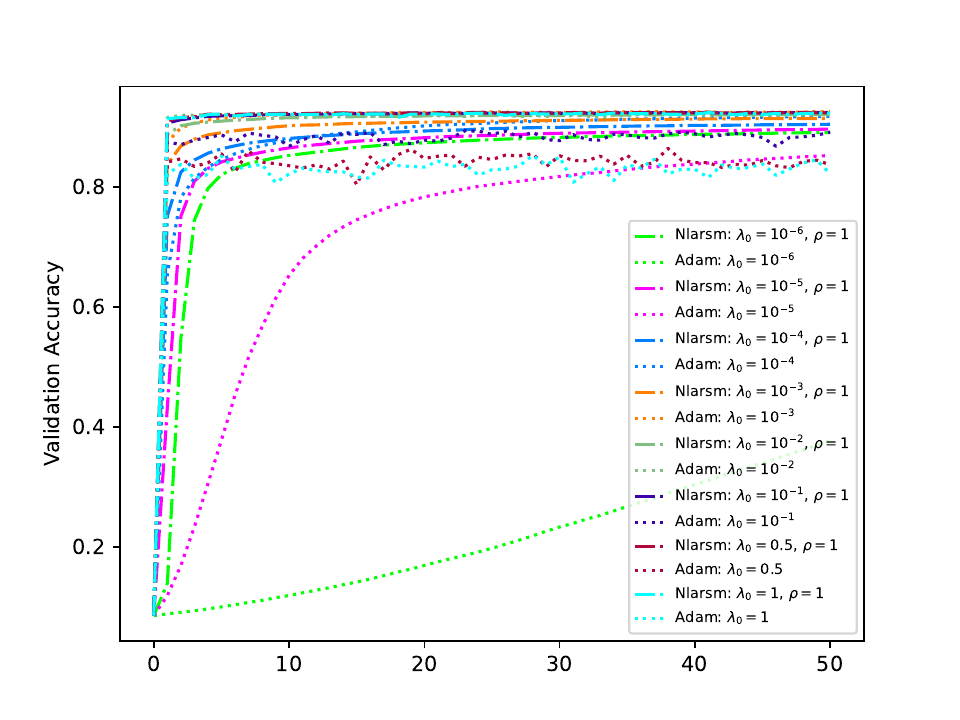}\hspace{-0.5mm}%
  \includegraphics[width=0.22\linewidth, trim=0 10 30 0]{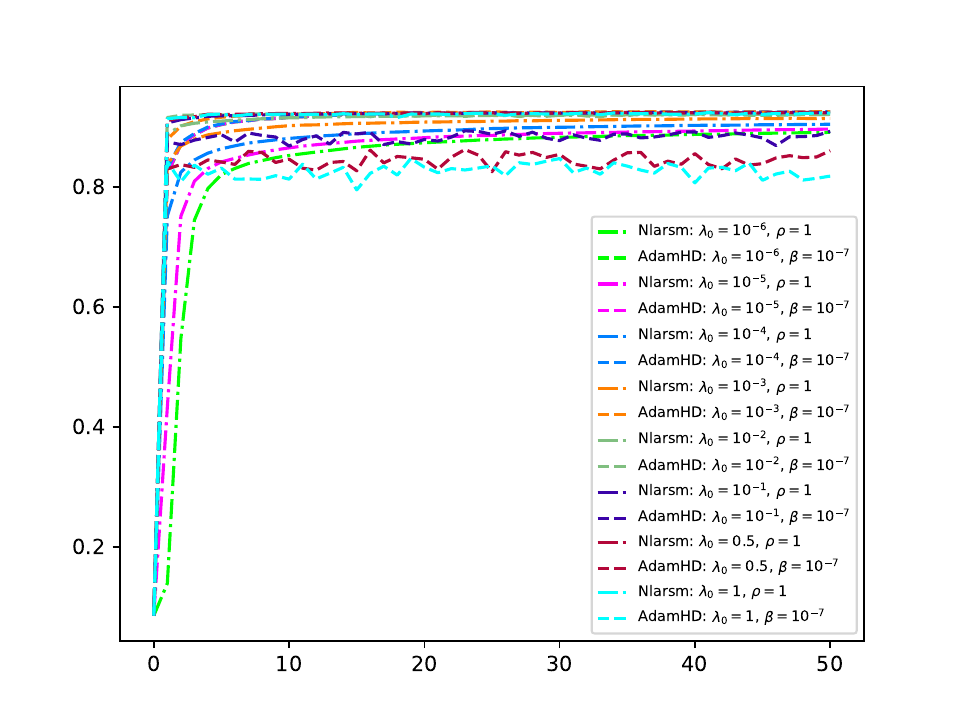}\hspace{-0.5mm}%
  \includegraphics[width=0.22\linewidth, trim=0 10 30 0]{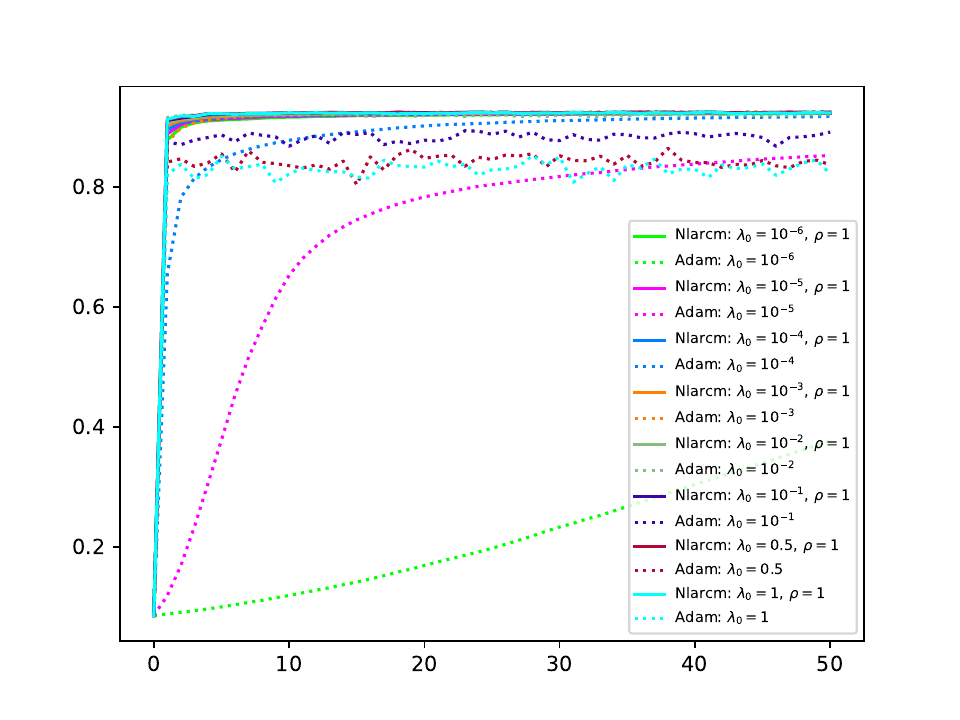}\hspace{-0.5mm}%
  \includegraphics[width=0.22\linewidth, trim=0 10 30 0]{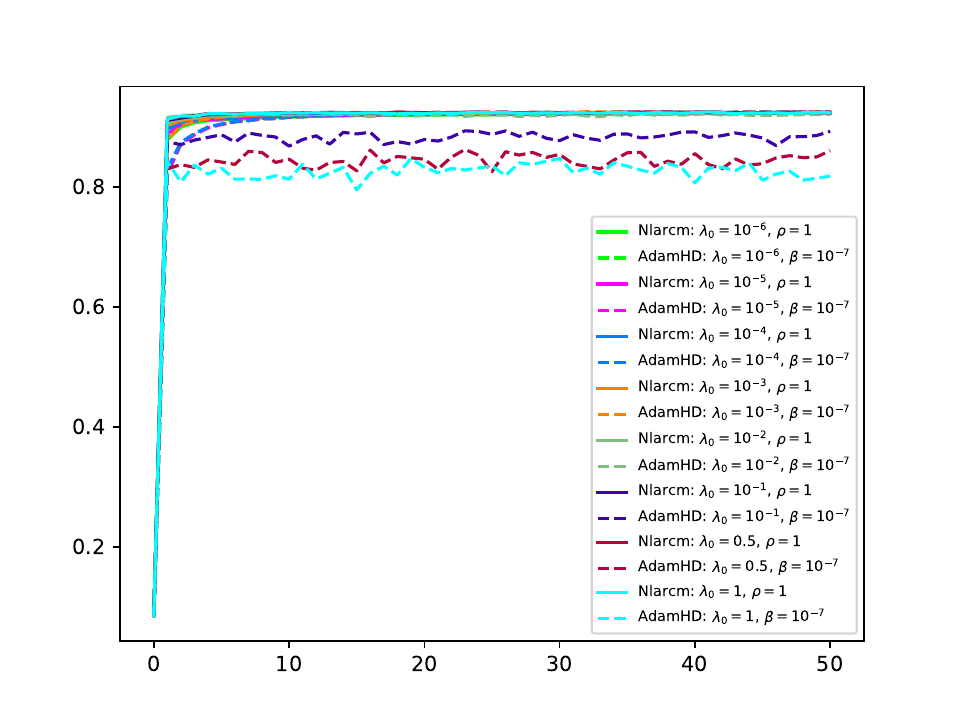}

  \includegraphics[width=0.22\linewidth, trim=0 10 30 0]{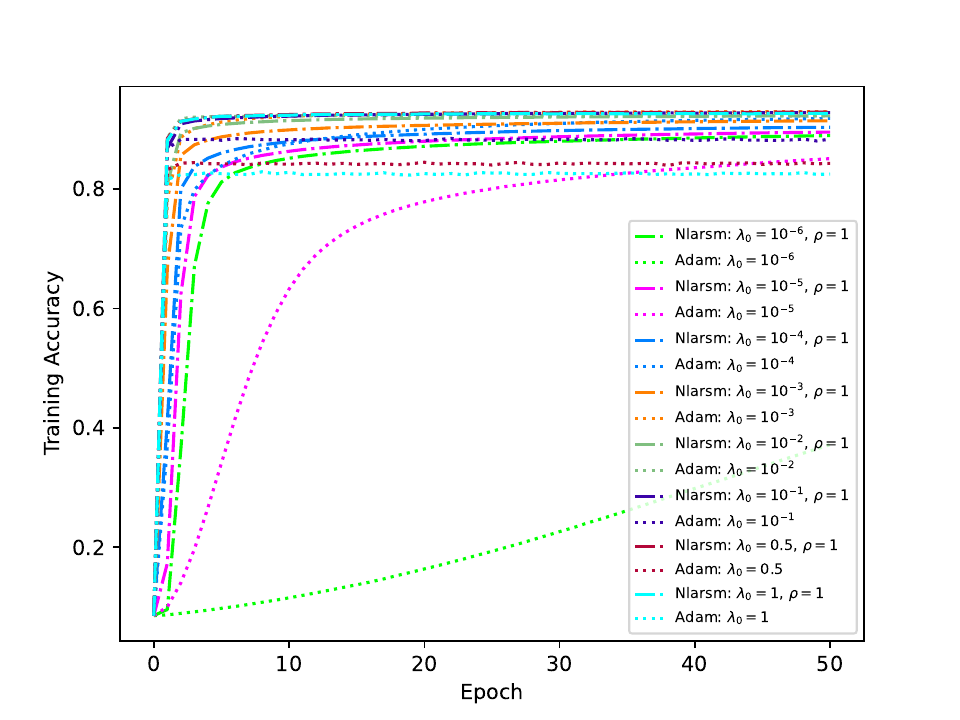}\hspace{-0.5mm}%
  \includegraphics[width=0.22\linewidth, trim=0 10 30 0]{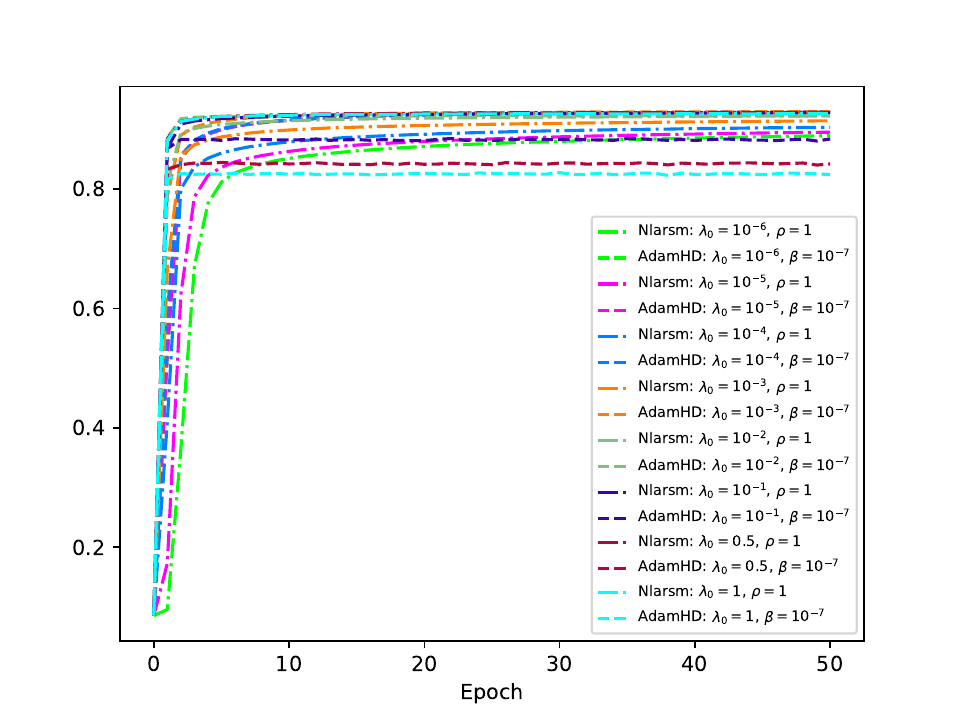}\hspace{-0.5mm}%
  \includegraphics[width=0.22\linewidth, trim=0 10 30 0]{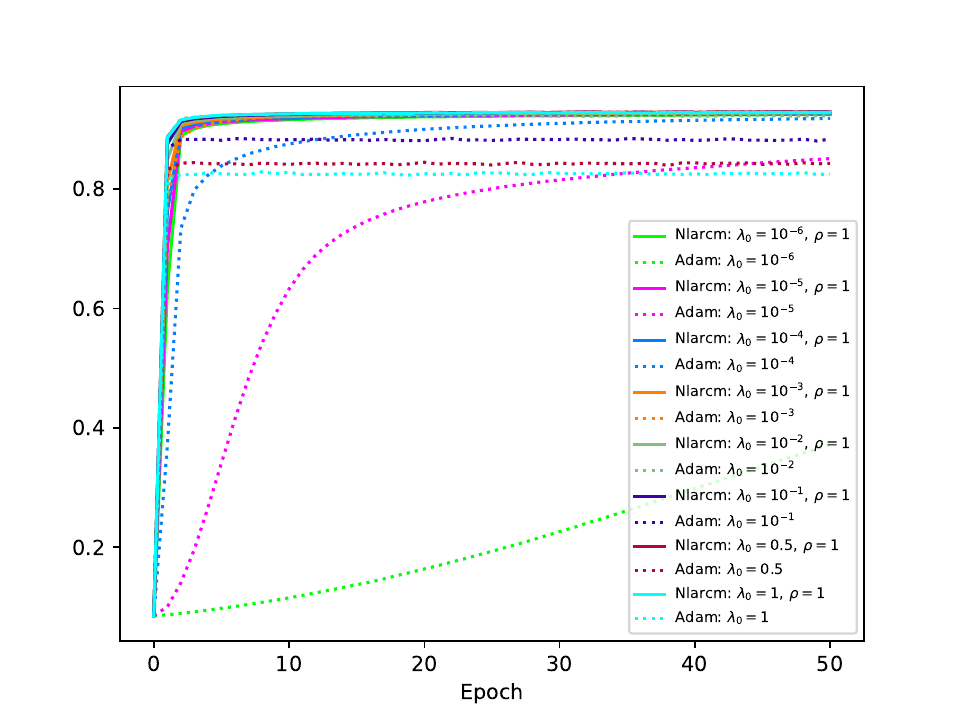}\hspace{-0.5mm}%
  \includegraphics[width=0.22\linewidth, trim=0 10 30 0]{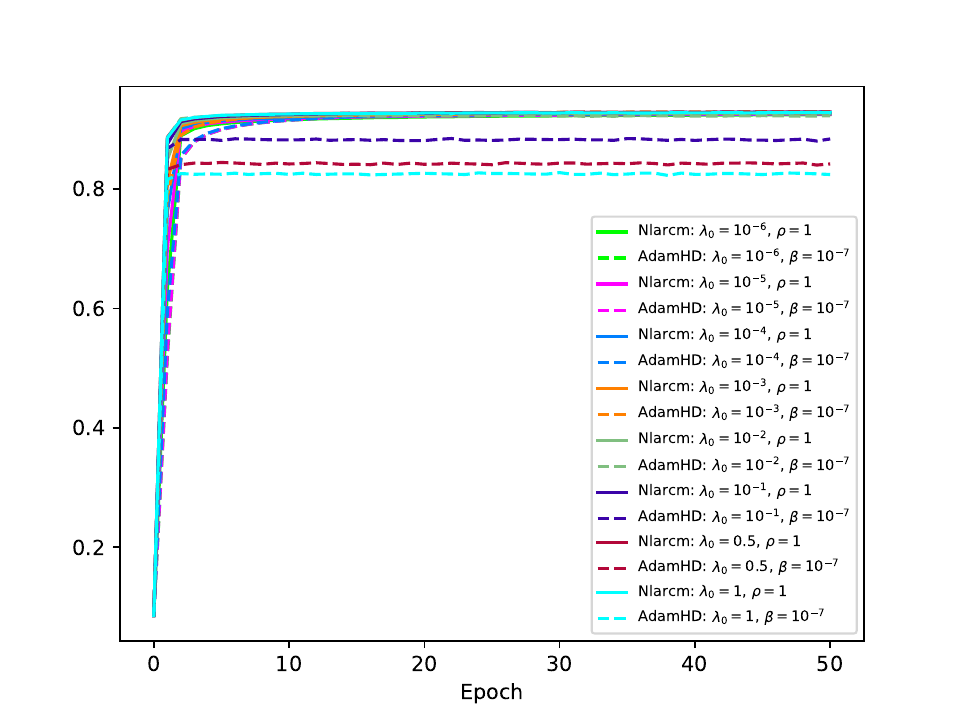}

    \vspace{1\baselineskip}
  \caption{Logistic regression model on MNIST data: Performance comparison of Nlarsm and Nlarcm versus Adam and AdamHD (with $\beta = 10^{-7}$) across varied learning rates. }
  \vspace{1\baselineskip}
  \label{fig:exp-logistic-mnist-nlar-adam-adamhd-1e-7}
\end{figure*}

\paragraph{The Logistic regression model on the MNIST dataset.} The results for comparison of Nlarsm and Nlarcm versus Adam and AdamHD are shown in Figure \ref{fig:exp-logistic-mnist-nlar-adam-adamhd-1e-7} for $\beta=10^{-7}$, for $\beta=10^{-4}$, see Figure \ref{fig:exp-logistic-mnist-nlar-adam-adamhd-1e-4} of the appendix. We observe that the performance of Nlarsm and Nlarcm improves as the initial learning rate increases. In contrast, the performance of both Adam and AdamHD is adversely affected by large initial learning rates. While AdamHD outperforms Adam, Nlarsm and Nlarcm perform the same or better across different initial learning rates. Regardless of the initial learning rate, Nlarcm consistently performs well. The main observation is the efficiency and robustness of Nlarsm and Nlarcm across various learning rates, a characteristic that appears in other experiments as well.

\begin{figure*}[!ht]
  \centering
  \includegraphics[width=0.22\linewidth, trim=0 10 30 0]{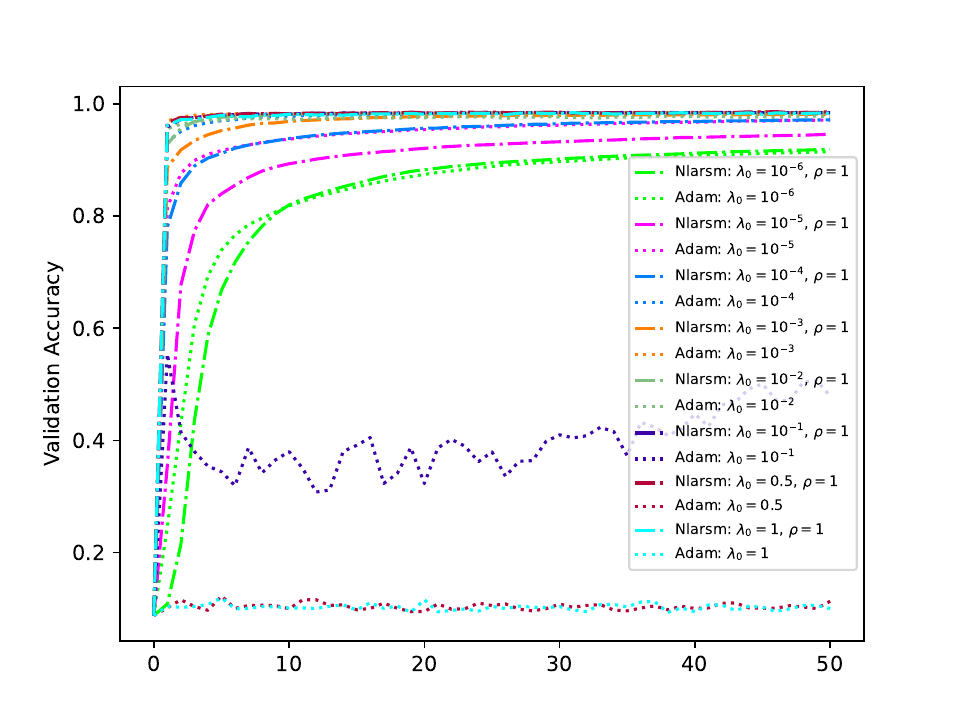}\hspace{-0.5mm}%
  \includegraphics[width=0.22\linewidth, trim=0 10 30 0]{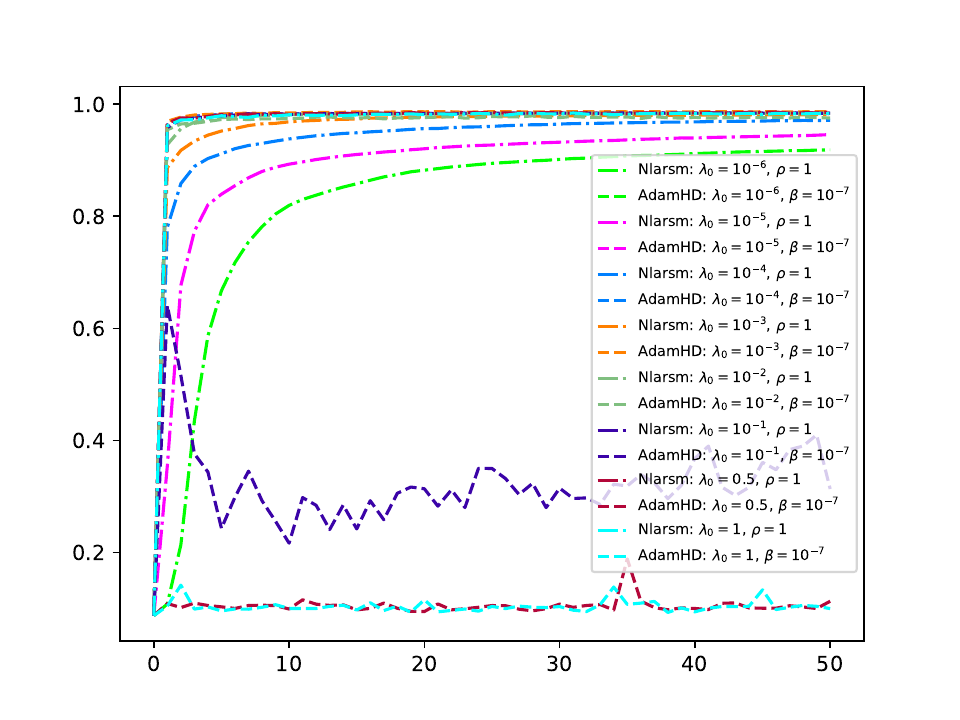}\hspace{-0.5mm}%
  \includegraphics[width=0.22\linewidth, trim=0 10 30 0]{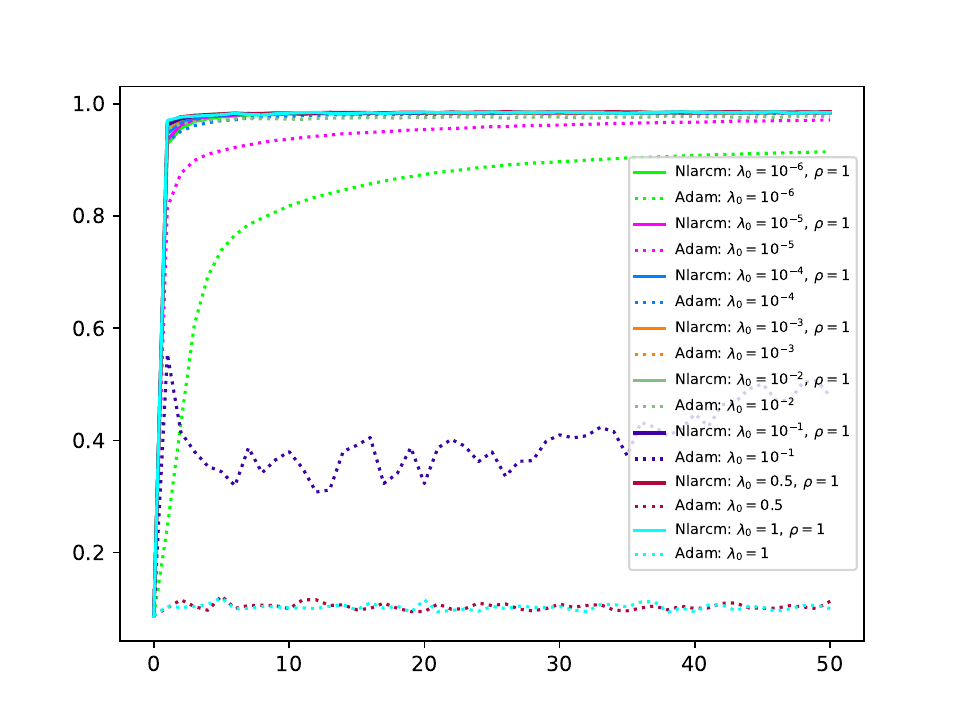}\hspace{-0.5mm}%
  \includegraphics[width=0.22\linewidth, trim=0 10 30 0]{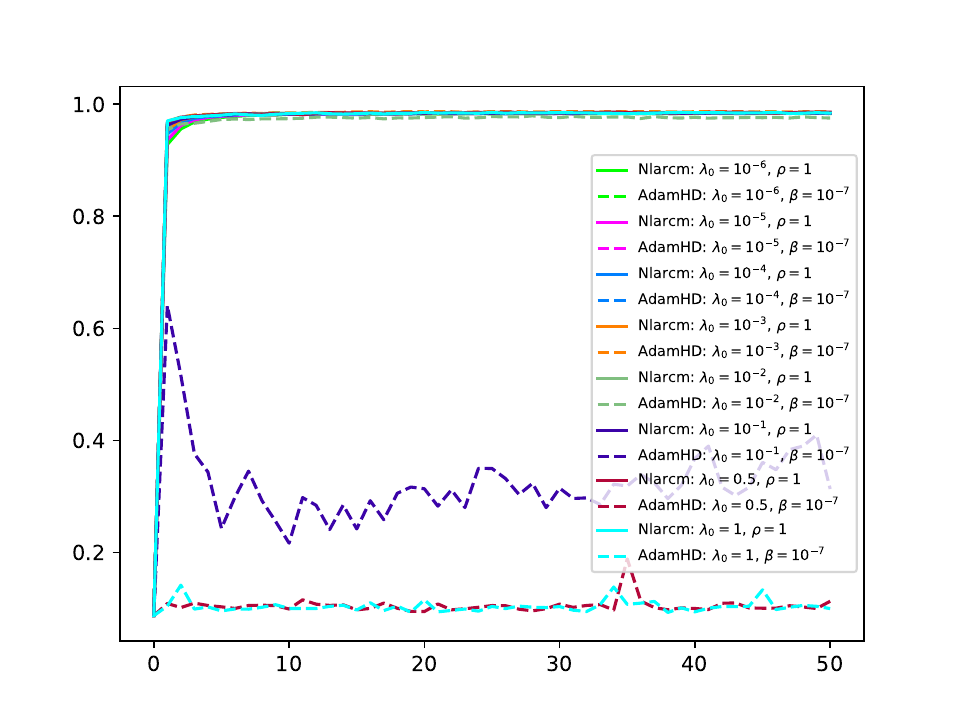}

  \includegraphics[width=0.22\linewidth, trim=0 10 30 0]{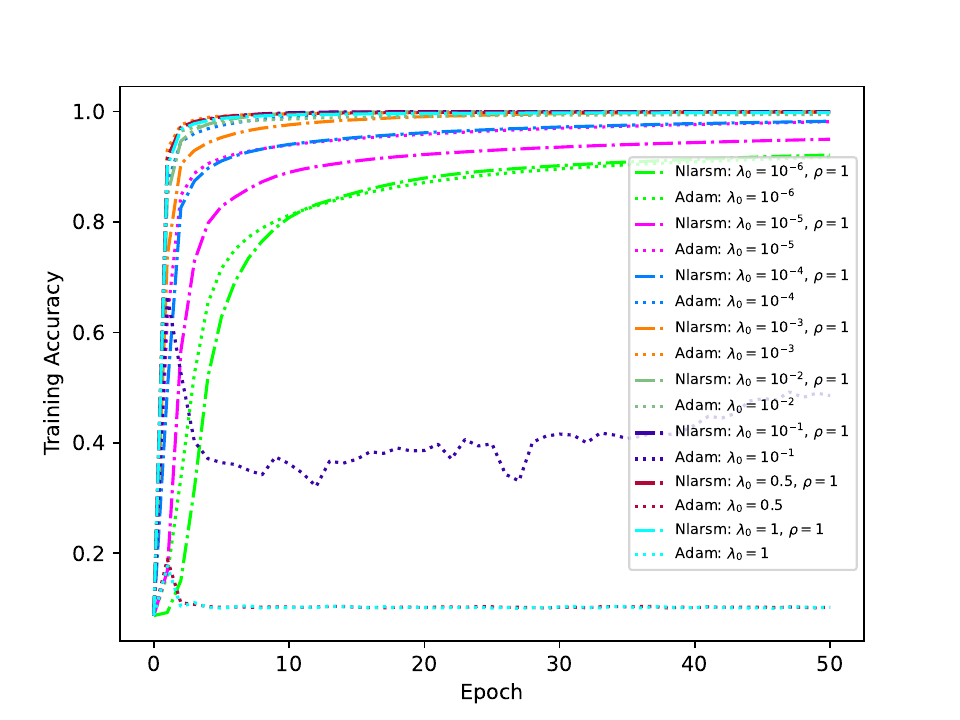}\hspace{-0.5mm}%
  \includegraphics[width=0.22\linewidth, trim=0 10 30 0]{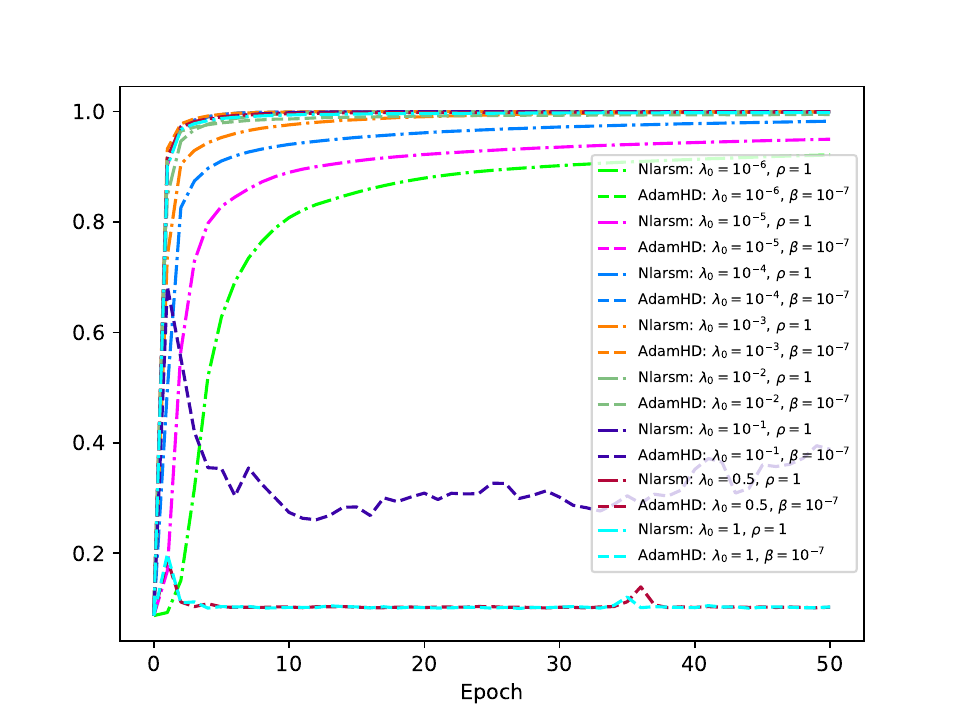}\hspace{-0.5mm}%
  \includegraphics[width=0.22\linewidth, trim=0 10 30 0]{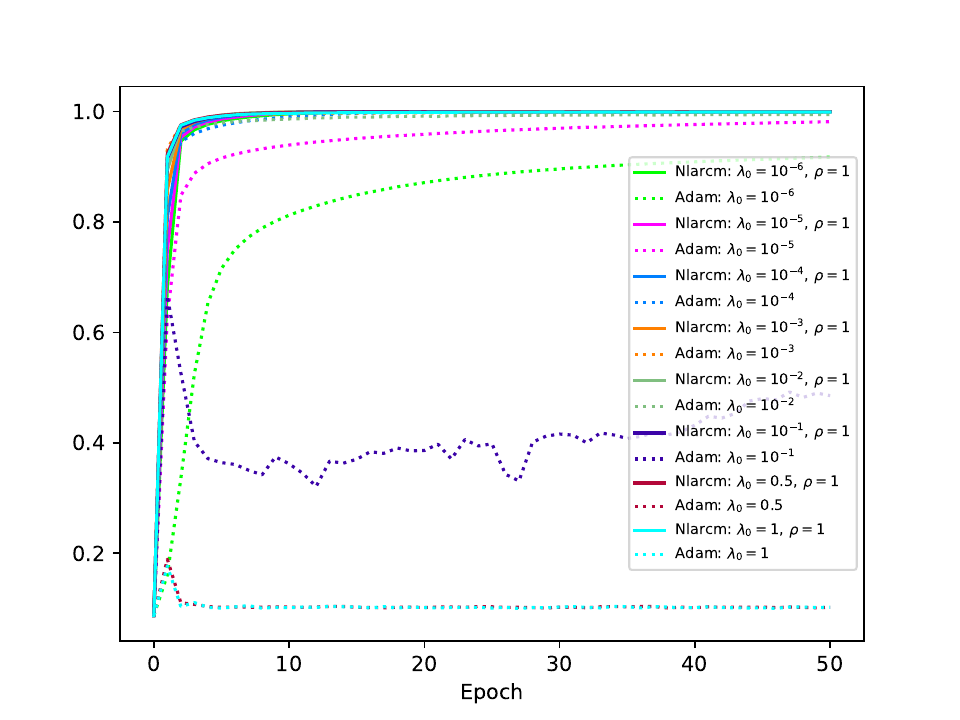}\hspace{-0.5mm}%
  \includegraphics[width=0.22\linewidth, trim=0 10 30 0]{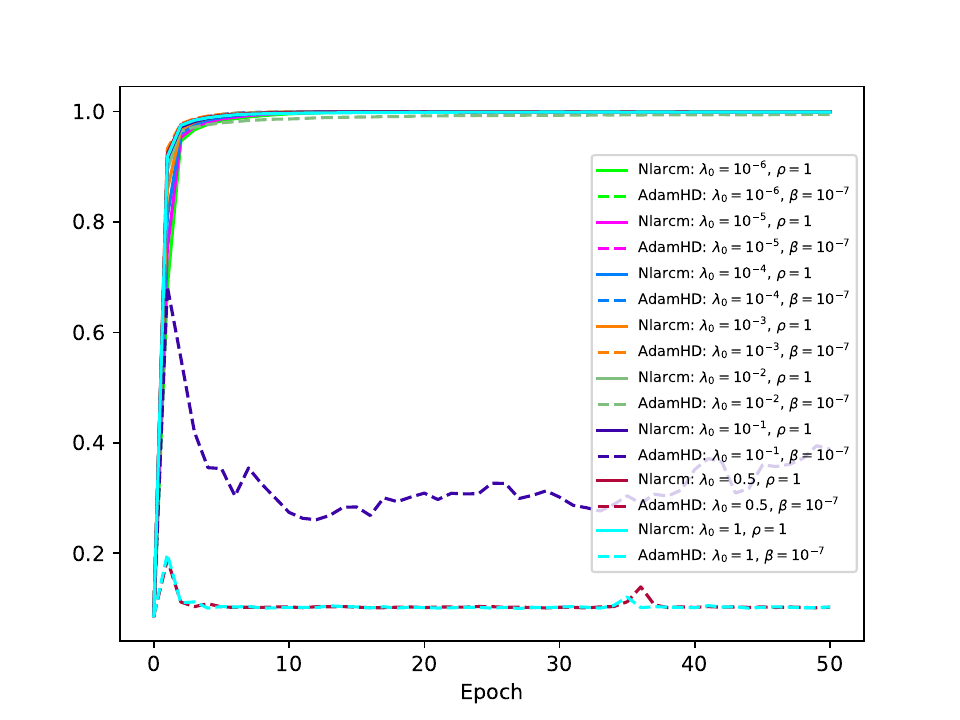}

    \vspace{1\baselineskip}
  \caption{MLP2h on the MNIST dataset: Performance comparison of Nlarsm and Nlarcm versus Adam and AdamHD (with $\beta = 10^{-7}$) across varied learning rates. }
  \vspace{1\baselineskip}
  \label{fig:exp-mlp-mnist-nlar-adam-adamhd-1e-7}
\end{figure*}

\paragraph{The MLP2h model on the MNIST dataset.} Recall that the MLP2h model features two hidden layers, each with 1,000 neurons and ReLU activation functions on each node. As it is observed in Figure \ref{fig:exp-mlp-mnist-nlar-adam-adamhd-1e-7}, while Adam and AdamHD show instability or divergence for $\lambda_0 \geq 0.1$, Nlarsm and Nlarcm demonstrate a remarkable resistance to divergence and perform well even with high values of initial learning rates. In both Nlarsm and Nlarcm, the performance improves as the initial learning rate increases.  The properties of Nlarsm and Nlarcm in this experiment are similar to those observed in the first experiment. Figure \ref{fig:exp-mlp-mnist-nlar-adam-adamhd-1e-4} of the appendix shows a similar comparison for $\beta = 10^{-4}$.  

\begin{figure*}[!ht]
  \centering
  \includegraphics[width=0.22\linewidth, trim=0 10 30 0]{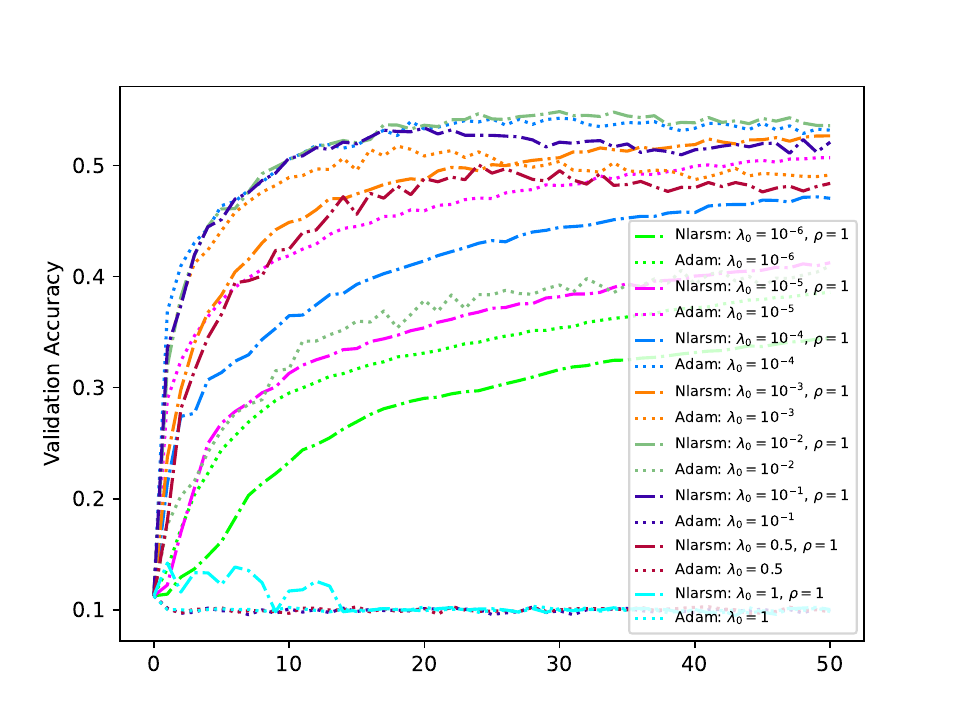}\hspace{-0.5mm}%
  \includegraphics[width=0.22\linewidth, trim=0 10 30 0]{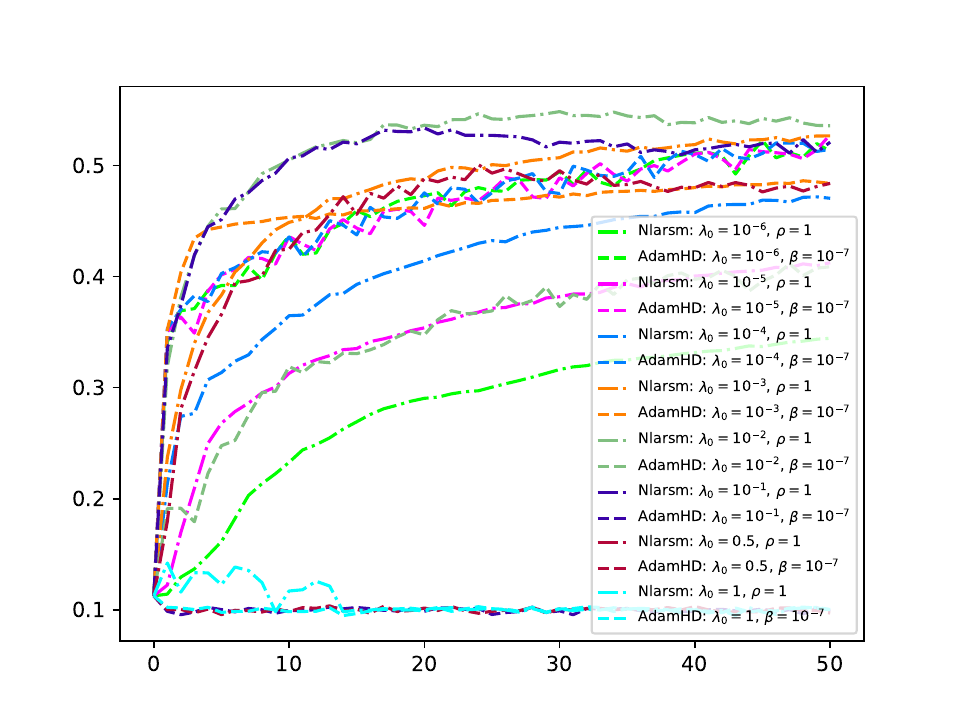}\hspace{-0.5mm}%
  \includegraphics[width=0.22\linewidth, trim=0 10 30 0]{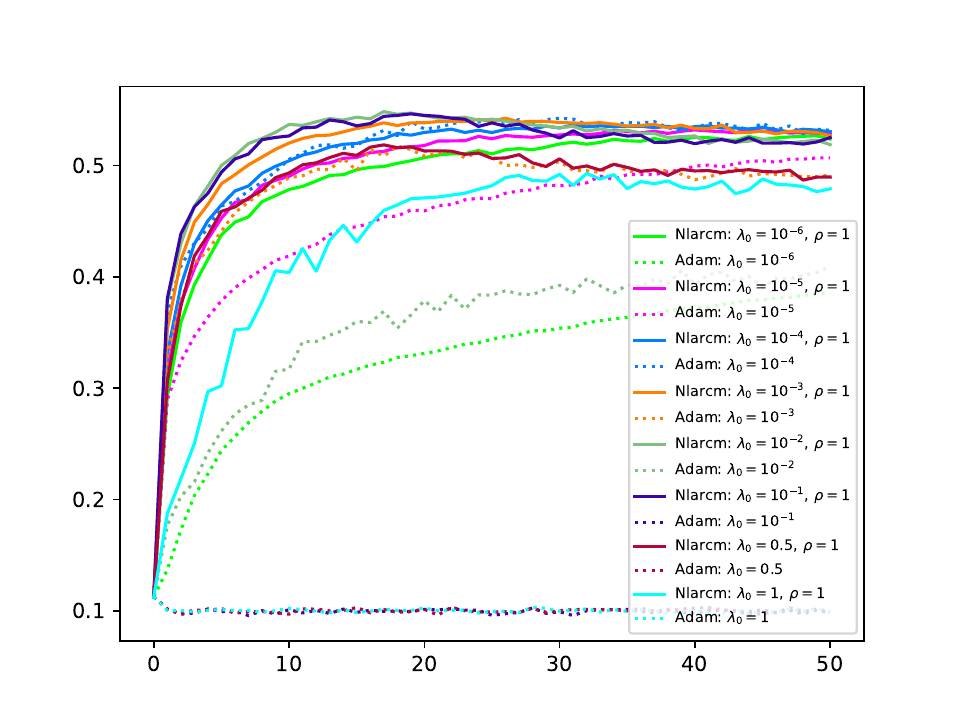}\hspace{-0.5mm}%
  \includegraphics[width=0.22\linewidth, trim=0 10 30 0]{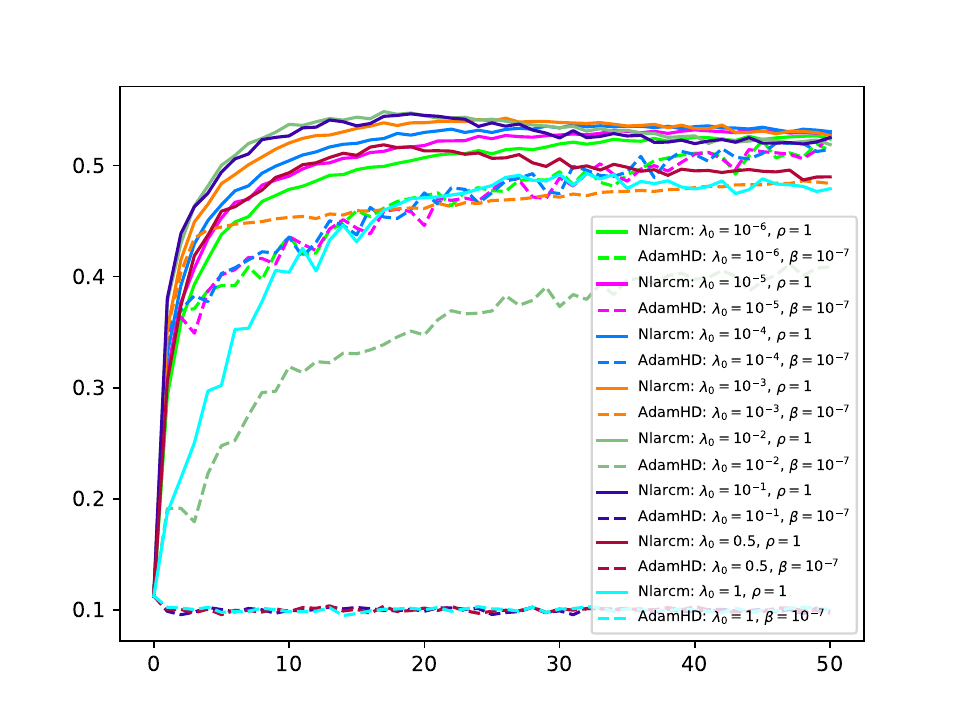}

  \includegraphics[width=0.22\linewidth, trim=0 10 30 0]{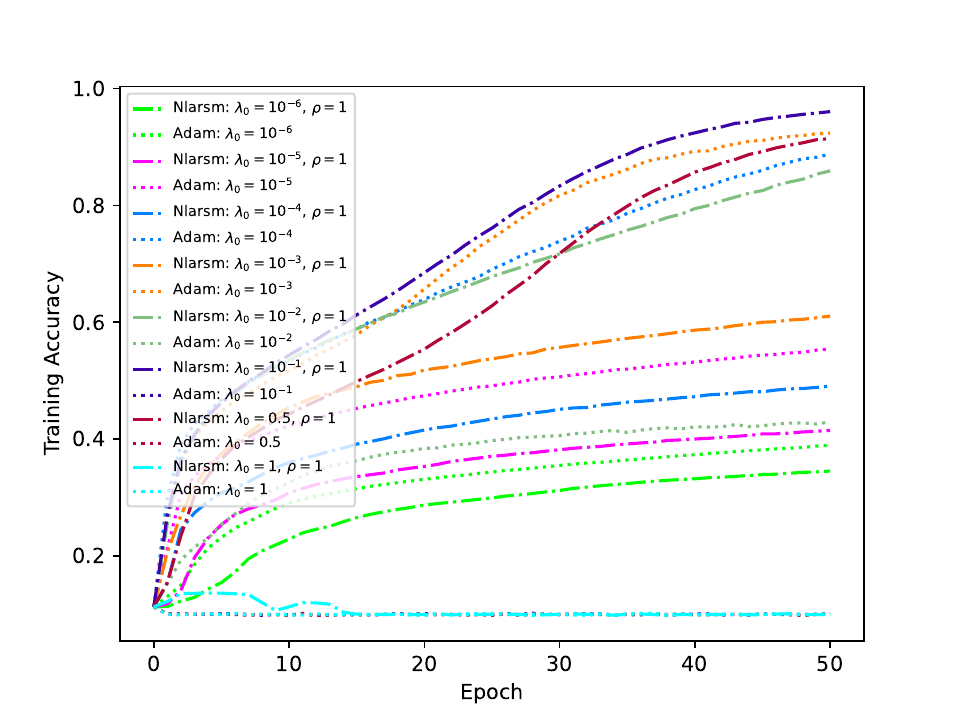}\hspace{-0.5mm}%
  \includegraphics[width=0.22\linewidth, trim=0 10 30 0]{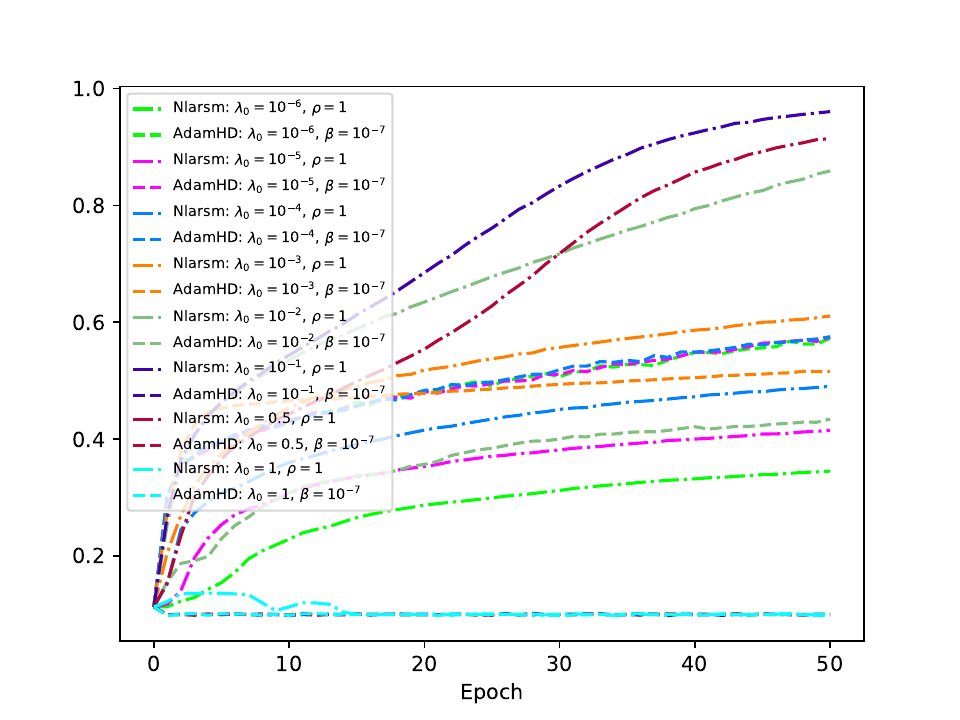}\hspace{-0.5mm}%
  \includegraphics[width=0.22\linewidth, trim=0 10 30 0]{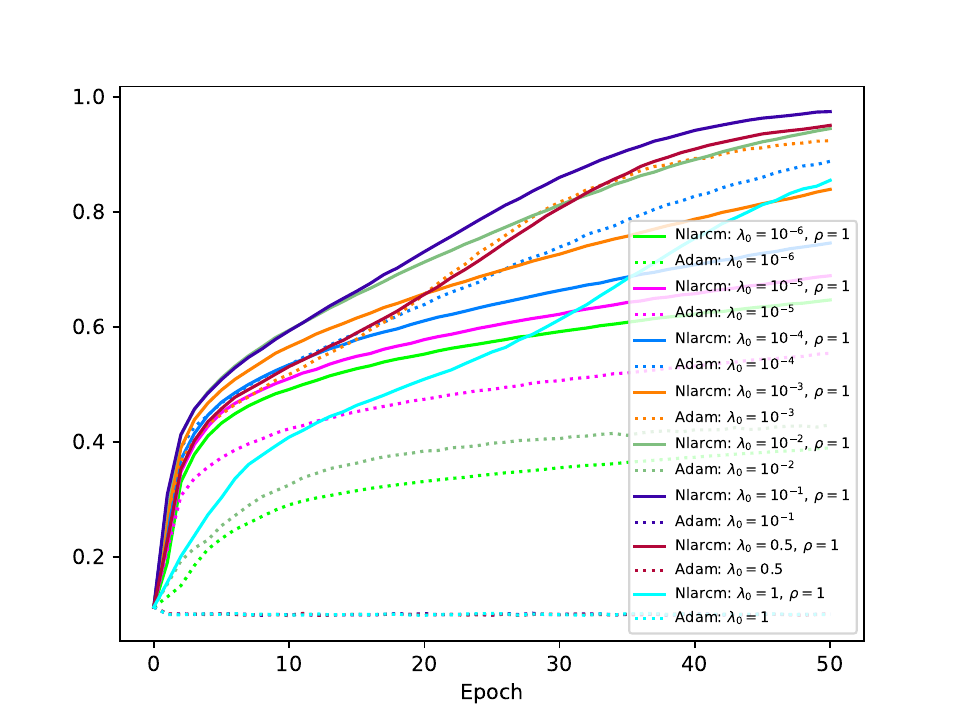}\hspace{-0.5mm}%
  \includegraphics[width=0.22\linewidth, trim=0 10 30 0]{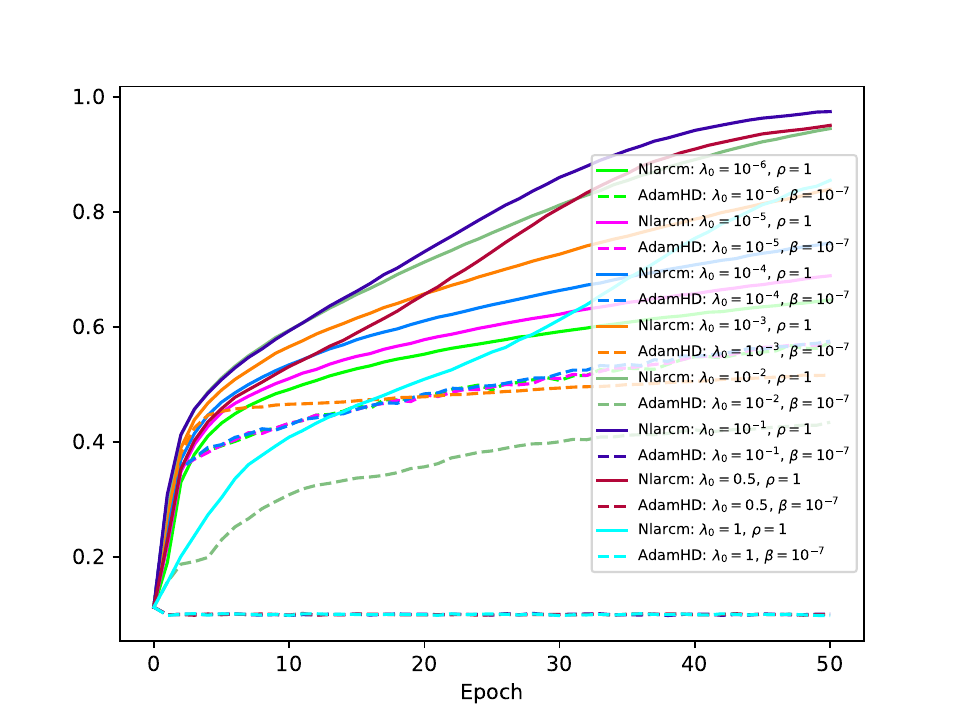}

    \vspace{1\baselineskip}
  \caption{MLP7h on the CIFAR10 dataset: Performance comparison of Nlarsm and Nlarcm versus Adam and AdamHD (with $\beta = 10^{-7}$) across varied learning rates. }
  \vspace{1\baselineskip}
  \label{fig:exp-mlp7h-CIFAR10-nlar-adam-adamhd-1e-7}
\end{figure*}

\paragraph{The MLP7h model on the CIFAR10 dataset.} Recall that MLP7h features seven hidden layers each with 512 nodes, ReLU activation functions for all nodes in the hidden layers, and softmax activation for the outer layer. Figure \ref{fig:exp-mlp7h-CIFAR10-nlar-adam-adamhd-1e-7} shows the comparison of Nlarsm and Nlarcm with Adam and AdamHD with $\beta=10^{-7}$. The same as the first two experiments, Nlarsm and Nlarcm performs quite well, compared to Adam and AdamHD, across large values of initial learning rates. The performance during the first 10 epochs is particularly remarkable, an observation that was not as apparent in the first two experiments. For most learning rates, Nlarsm and Nlarcm achieve the best performance before epoch 20 (for $\lambda_0=0.1$) after which  the validation accuracy curve begins changing direction slightly indicating overfitting. Overfitting is a common issue in adaptive learning methods and is not specific to Nlar \citep{wilson2017marginal}. The robustness feature persists here as well; both Adam and AdamHD do not show any convergence for $\lambda_0 \geq 0.1$.

\begin{figure*}[!ht]
  \centering
  \includegraphics[width=0.22\linewidth, trim=0 10 30 0]{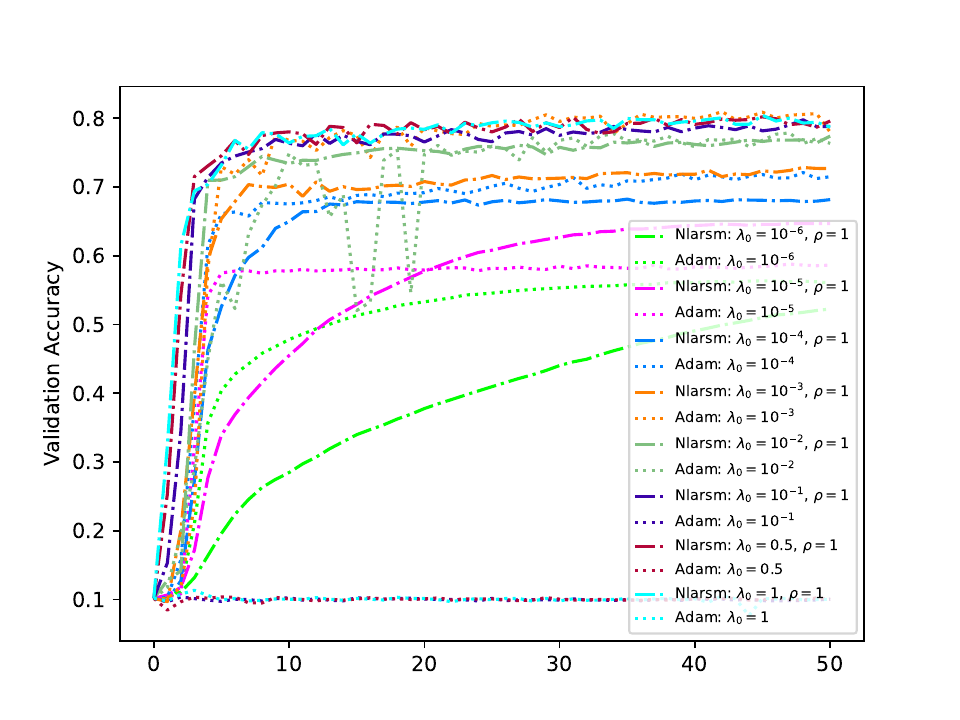}\hspace{-0.5mm}%
  \includegraphics[width=0.22\linewidth, trim=0 10 30 0]{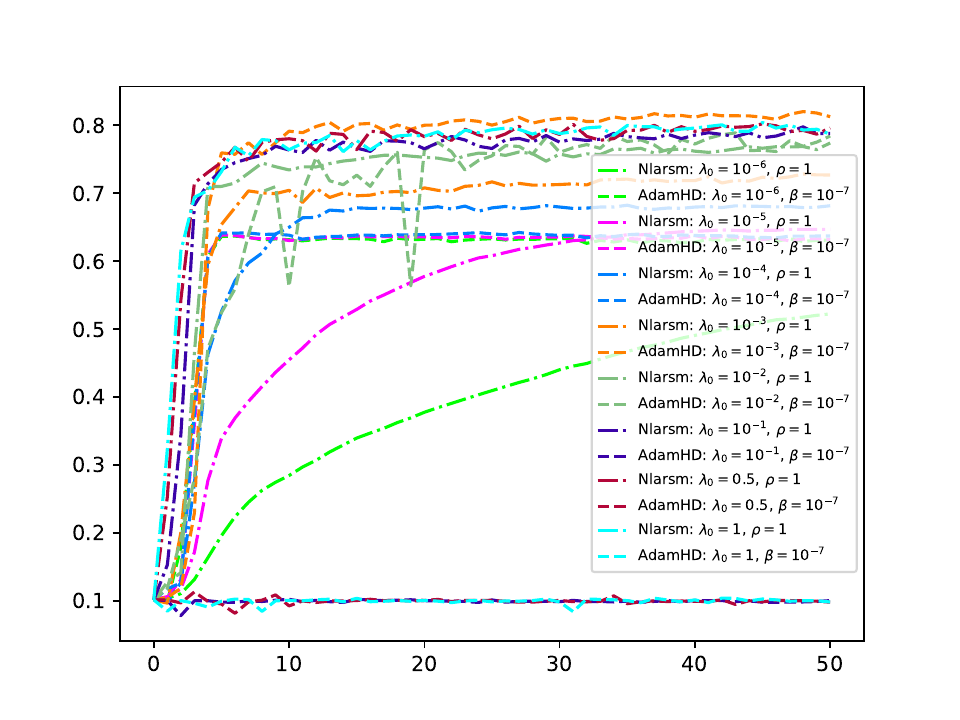}\hspace{-0.5mm}%
  \includegraphics[width=0.22\linewidth, trim=0 10 30 0]{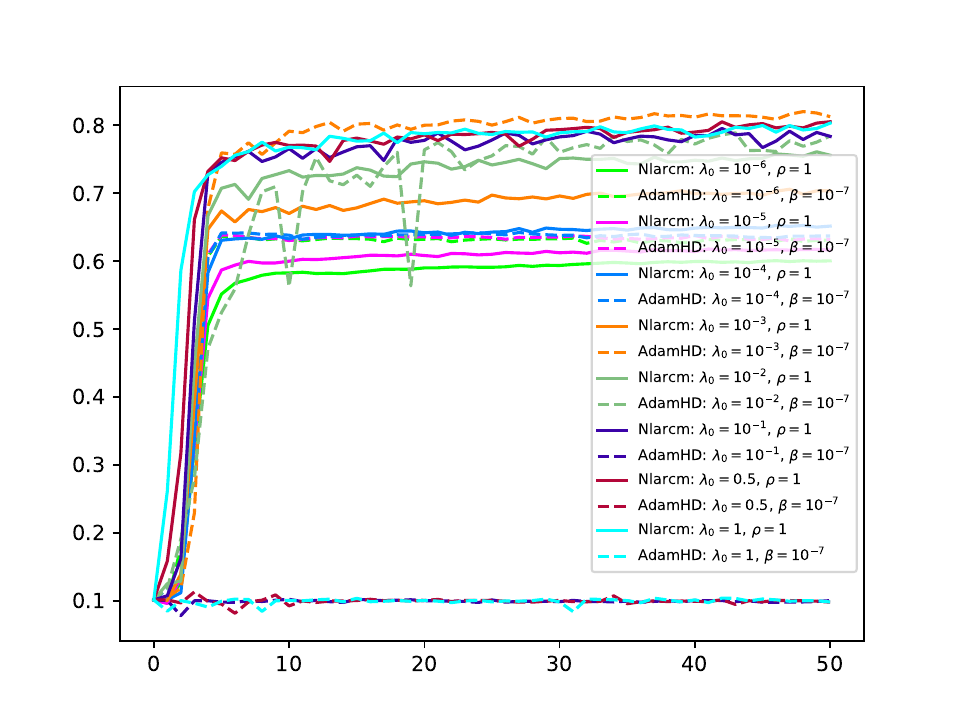}\hspace{-0.5mm}%
  \includegraphics[width=0.22\linewidth, trim=0 10 30 0]{imgs/CIFAR10_vgg11_nlarc_adamhd_mu=1_beta=1e-07_minlr=None_50_0_1_300val_accuracy.pdf}

  \includegraphics[width=0.22\linewidth, trim=0 10 30 0]{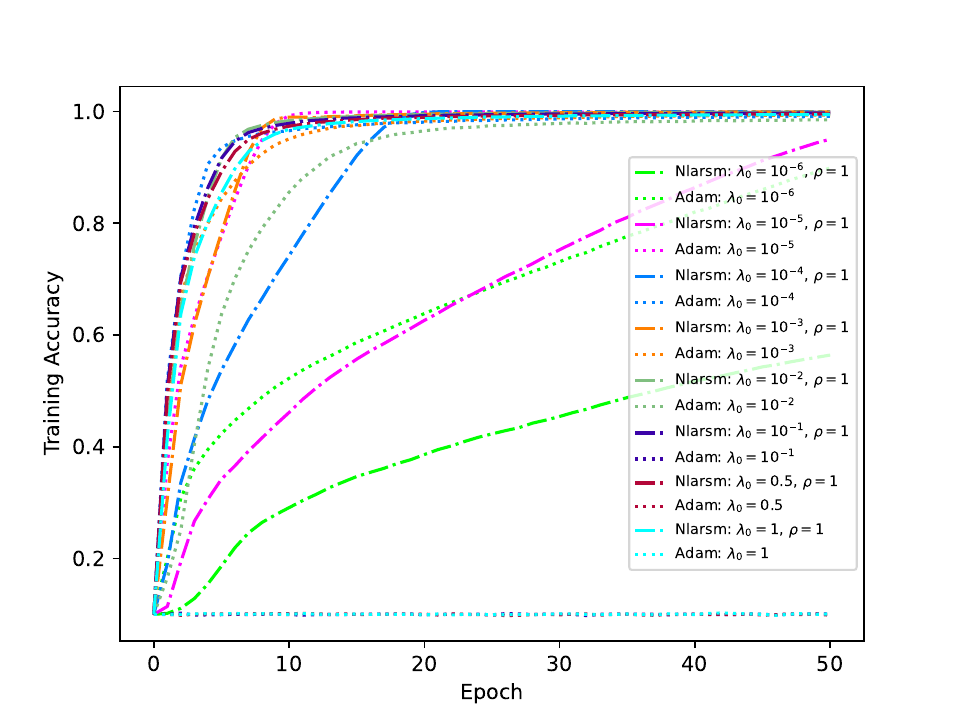}\hspace{-0.5mm}%
  \includegraphics[width=0.22\linewidth, trim=0 10 30 0]{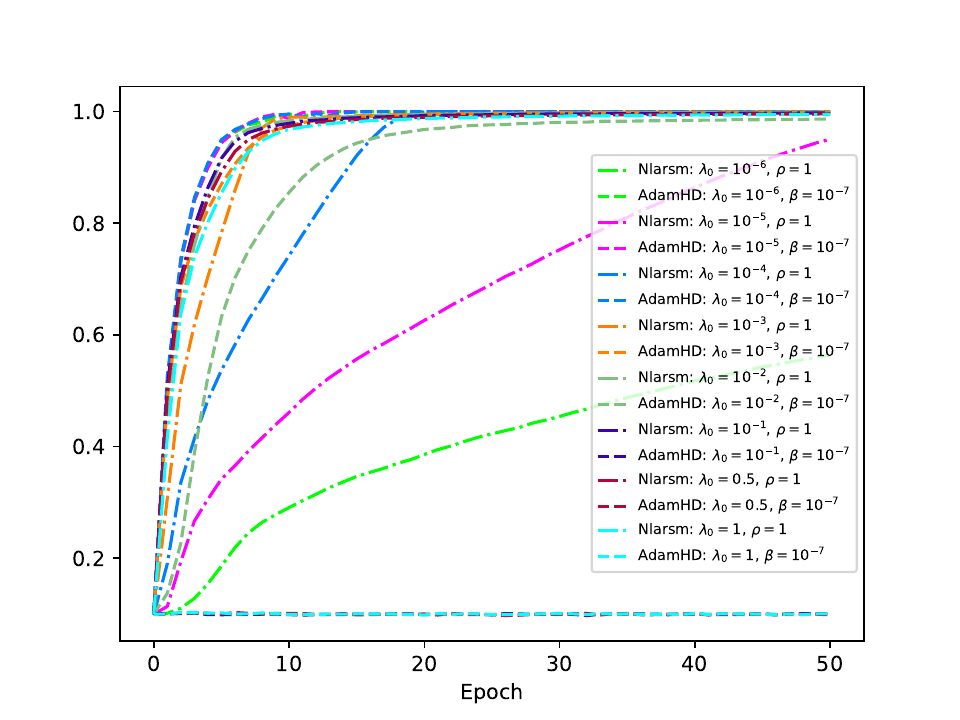}\hspace{-0.5mm}%
  \includegraphics[width=0.22\linewidth, trim=0 10 30 0]{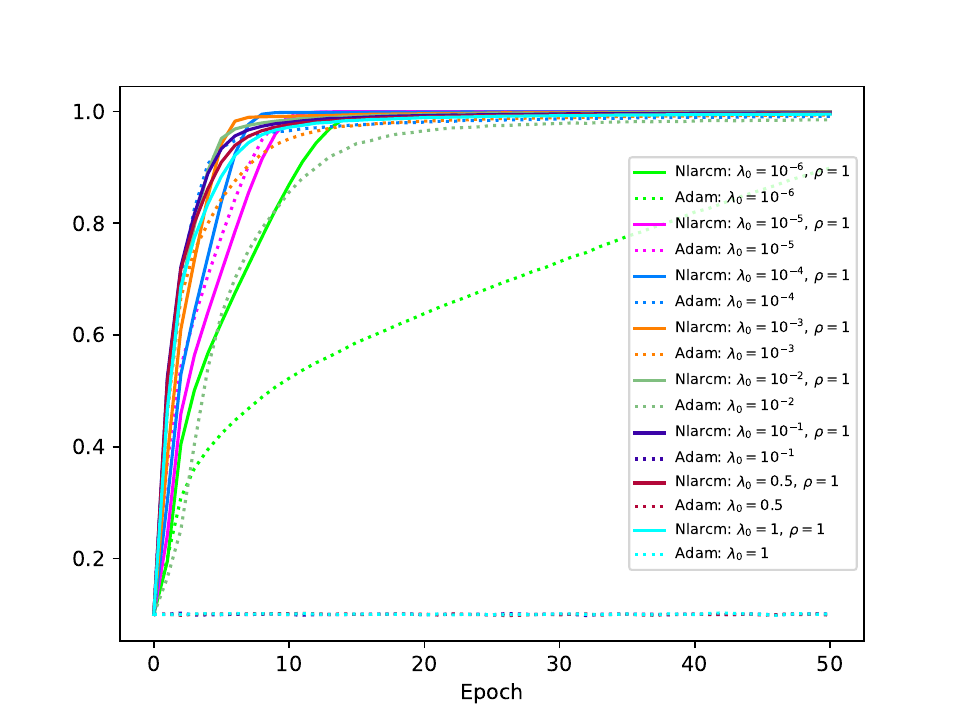}\hspace{-0.5mm}%
  \includegraphics[width=0.22\linewidth, trim=0 10 30 0]{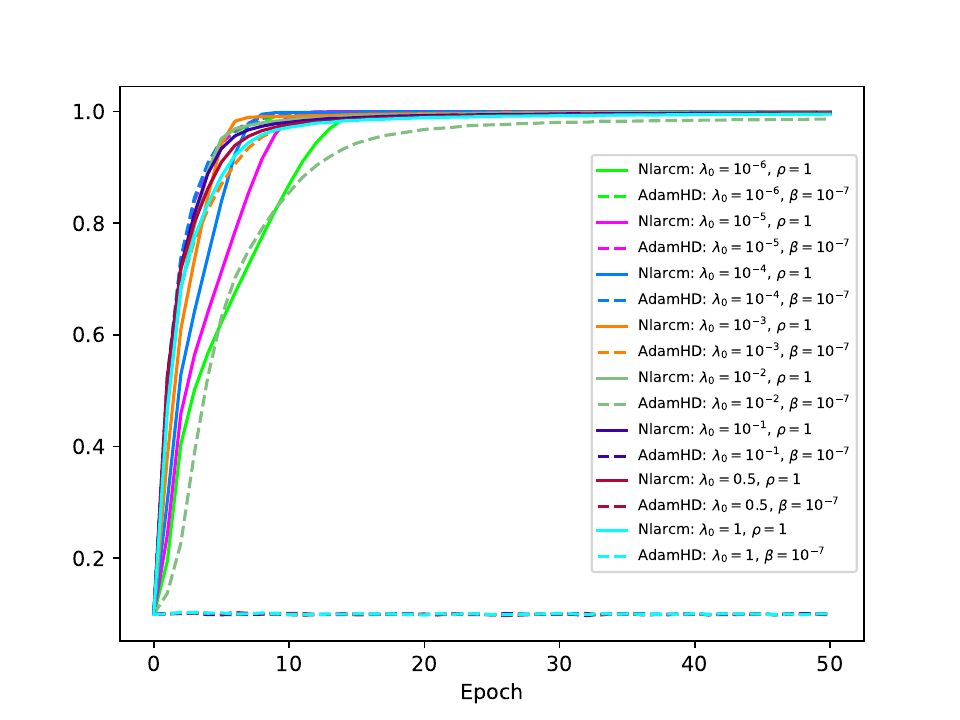}

    \vspace{1\baselineskip}
  \caption{VGG11 on the CIFAR10 dataset: Performance comparison of Nlarsm and Nlarcm versus Adam and AdamHD (with $\beta = 10^{-7}$) across varied learning rates. }
  \vspace{1\baselineskip}
  \label{fig:exp-vgg11-CIFAR10-nlar-adam-adamhd-1e-7}
\end{figure*}

\paragraph{The VGG11 on the CIFAR10 dataset.}
VGG11 is a more complex model and help us to better understand the properties of the optimizers. For AdamHD, we have tried three values for $\beta$ that are $10^{-8}, 10^{-7}, 10^{-4}$, and the best performance was achieved by $\beta=10^{-7}$ shown in Figures \ref{fig:exp-vgg11-CIFAR10-nlar-adam-adamhd-1e-7} (the results for $\beta=10^{-8}$ and $\beta=10^{-4}$ are shown respectively in Figures \ref{fig:exp-vgg11-CIFAR10-nlar-adam-adamhd-1e-8} and \ref{fig:exp-vgg11-CIFAR10-nlar-adam-adamhd-1e-4} of the appendix). Both Adam and AdamHD, they either show instability or divergence for $\lambda_0\geq 0.02$. In contrast, Nlarsm and Nlarcm have shown a stable convergence even for $\lambda_0=1$. Like the previous experiments, Nlarscm shows a remarkable speed in convergence for the first few epochs. The best performance is achieved for $\lambda_0\geq 0.1$. There is only one specific case of $\beta=10^{-8}$ and $\lambda_0=0.001$ that AdamHD slightly outperforms Nlarsm and Nlarcm.

\begin{figure*}[!ht]
  \centering
  \includegraphics[width=0.22\linewidth, trim=0 10 30 0]{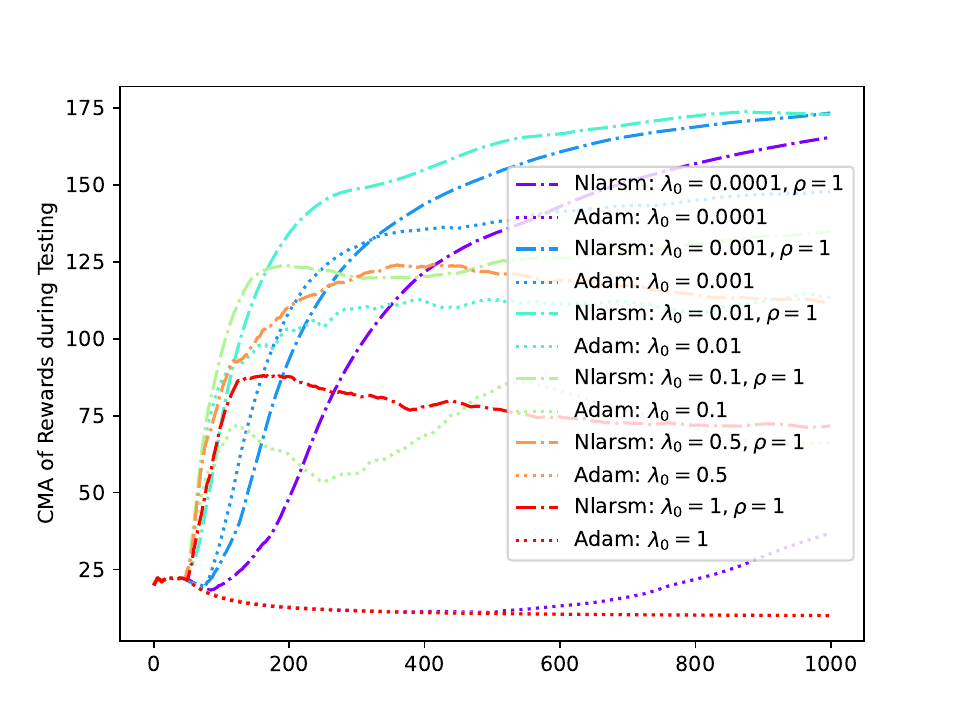}\hspace{-0.5mm}%
    \includegraphics[width=0.22\linewidth, trim=0 10 30 0]{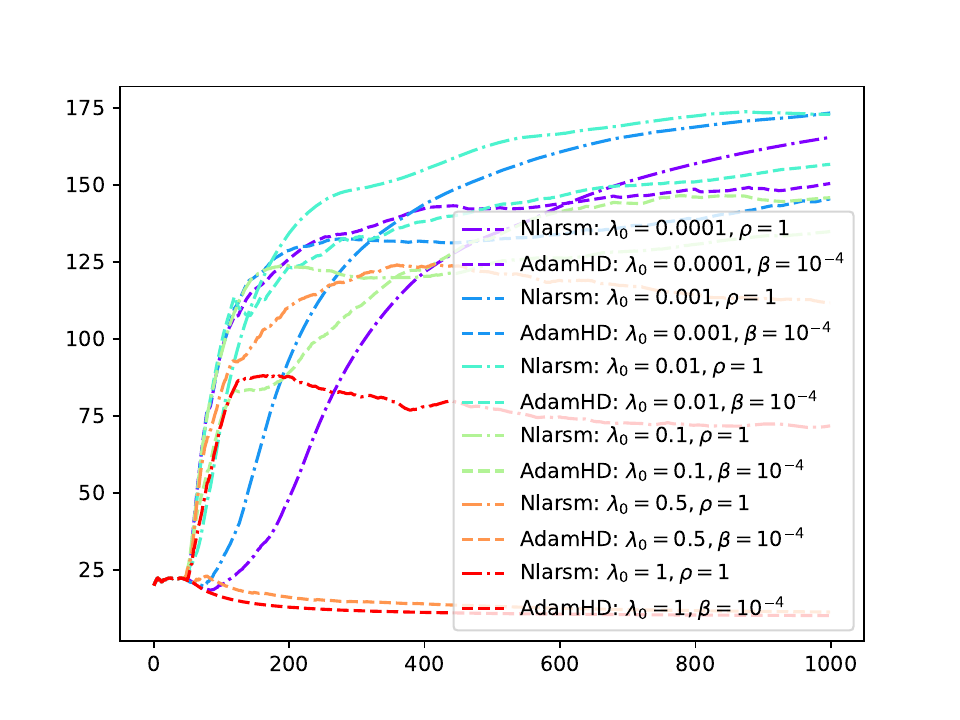}\hspace{-0.5mm}%
  \includegraphics[width=0.22\linewidth, trim=0 10 30 0]{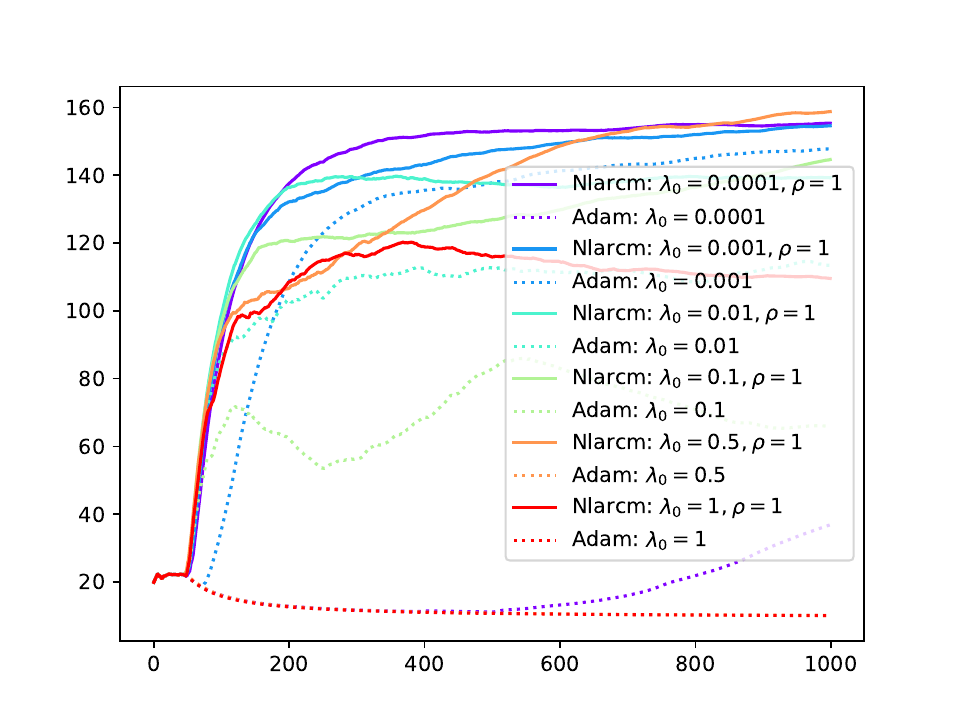}\hspace{-0.5mm}%
  \includegraphics[width=0.22\linewidth, trim=0 10 30 0]{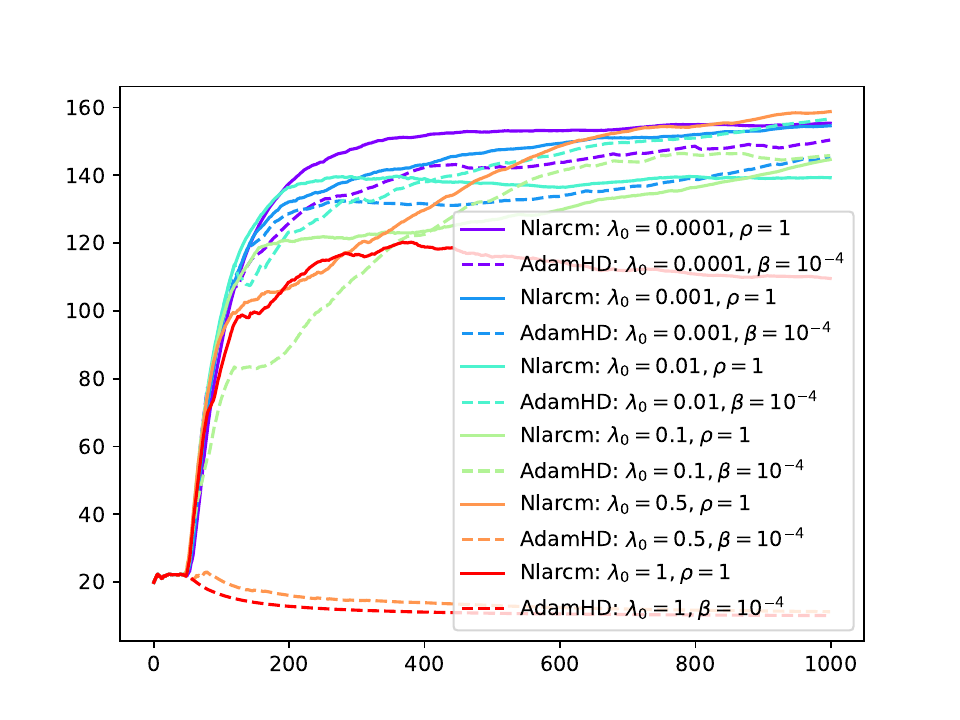}\hspace{-0.5mm}%

  \includegraphics[width=0.22\linewidth, trim=0 10 30 0]{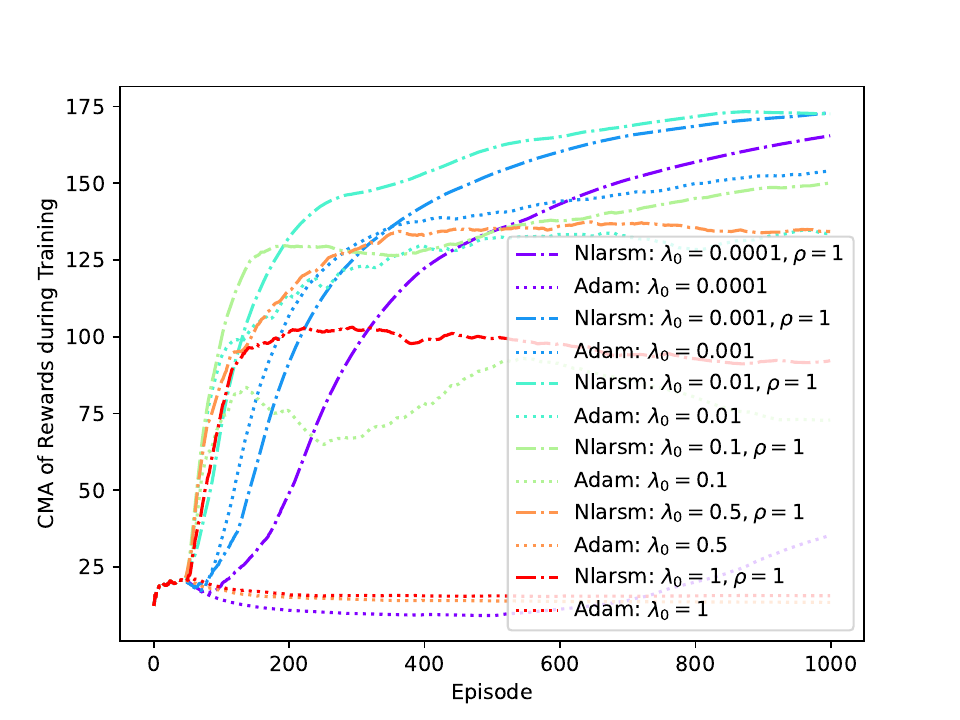}\hspace{-0.5mm}%
    \includegraphics[width=0.22\linewidth, trim=0 10 30 0]{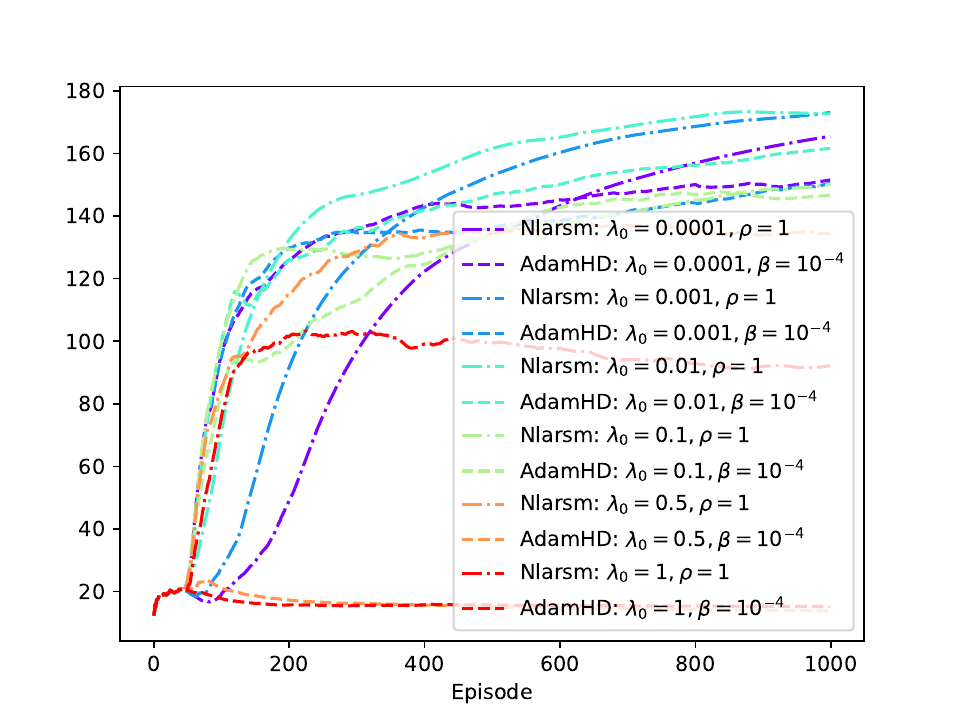}\hspace{-0.5mm}%
  \includegraphics[width=0.22\linewidth, trim=0 10 30 0]{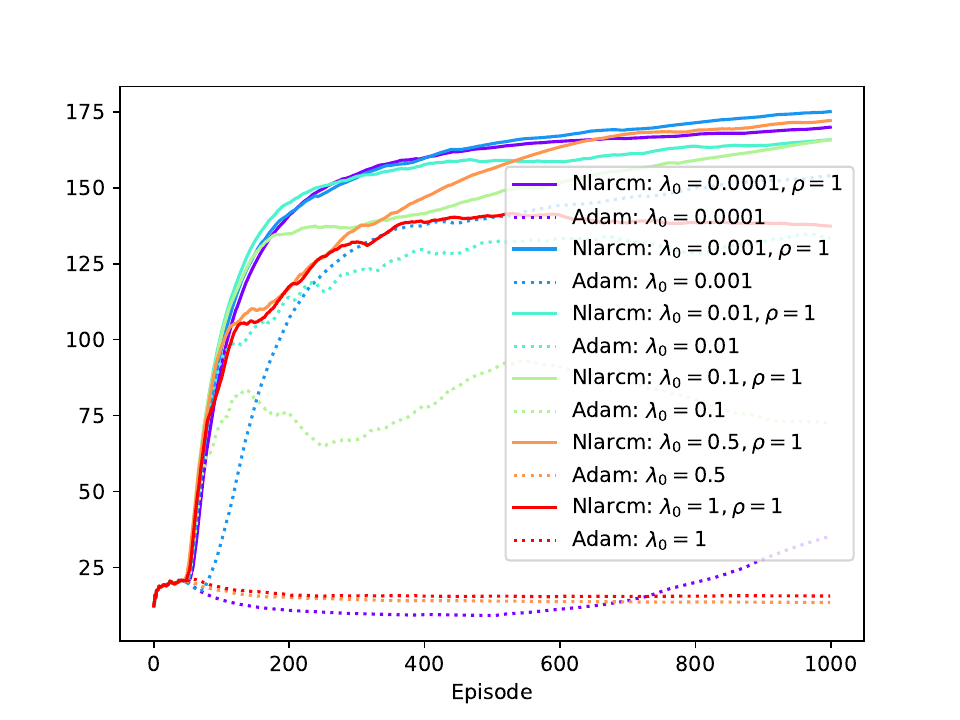}\hspace{-0.5mm}%
  \includegraphics[width=0.22\linewidth, trim=0 10 30 0]{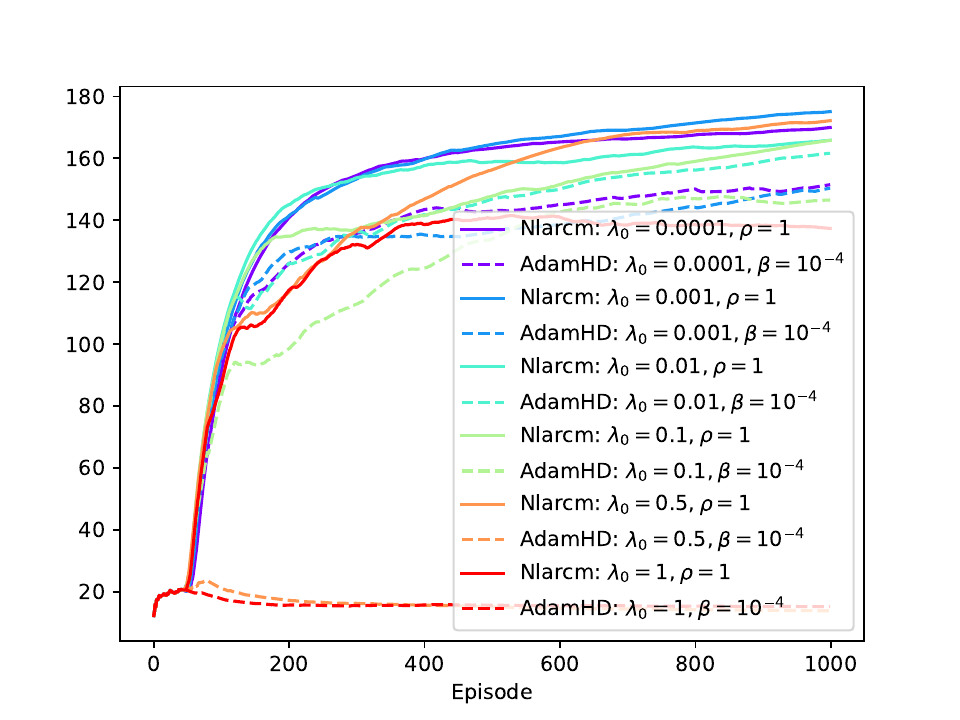}\hspace{-0.5mm}%
    \vspace{1\baselineskip}
  \caption{CartPole-v0: Performance comparison of Nlarsm and Nlarcm versus Adam and AdamHD (with $\beta = 10^{-4}$) across varied learning rates.}
  \vspace{1\baselineskip}
  \label{fig:exp-CartPole-nlar-adam-adamhd-beta=0_0001}
\end{figure*}

\paragraph{The DDQN model on the classical control problems of CartPole-v0.}   In this part of our experiments, we test and compare the effectiveness of these optimizers using the DDQN model. As it is shown in Figure \ref{fig:exp-CartPole-nlar-adam-adamhd-beta=0_0001}, overall, the agent equipped with the Nlarcm optimizer outperforms both Adam and AdamHD (for $\beta=10^{-4}$) in terms of consistency, robustness, and convergence. Figure \ref{fig:exp-CartPole-nlar-adam-adamhd-beta=1e-07} of the appendix, provides a similar result for AdamHD with $\beta=10^{-7}$. The behavior of Nlarsm eventually degrades only for very large learning rates $\lambda_0\geq 0.5$. 

For the initial episodes, unlike Adam and AdamHD, Nlarsm and Nlarcm perform well across all learning rates. However, in both Nlar algorithms, a more stable and consistent performance is eventually achieved with smaller initial learning rates. While this might seem contrary to the results from the classification experiments, we believe that this is due to the RL environments  being more stochastic and involve high variance in gradients due to the random nature of actions and rewards. In this context, small learning rates help by making cautious updates (as the agent explores the environment), leading to more stable learning and better long-term performance. In other words, small learning rates prevent the optimizer from making large, potentially destabilizing updates that could hinder learning. In contrast, in classification problems, the model receives immediate feedback from the loss function after each batch, so larger learning rates can accelerate the optimization process without the risk of destabilizing the learning as much as in RL.

\paragraph{Summary of the observations.} Across classification problems, both Nlarcm and Nlarsm demonstrate remarkable robustness and perform better in the presence of large learning rates. For small MLPs, Nlarcm's performance remains consistent across different learning rates, reducing the need to guess the best initial learning rates. However, like other adaptive learning methods, Nlarsm and Nlarcm have exhibited overfitting issues in the MLP7h model. This problem could be mitigated using strategies such as stopping rules, reducing learning rate schedules, or simply by lowering the momentum $\rho$.

Overall, the Nlarsm and Nlarcm optimizers perform well across a diverse range of initial learning rates during the first few epochs or episodes. This suggests strong initial adaptability, primarily due to leveraging adaptive momentum cautiously. This approach helps stabilize updates during the early phases of training, leading to robust performance across different learning rates.

\paragraph{Time Complexity.} The experiments were conducted using two NVIDIA GeForce RTX 3090 GPUs, each with 24 GB of VRAM, and an AMD Ryzen Threadripper PRO 3995WX processor with 64 cores and 128 threads. For small networks, the difference of execution times between different optimizers, is negligible. The longest running times per epoch were observed for the VGG11 model, with approximately 60.50, 58.25, 54.50, and 56.00 seconds per epoch for Nlarcm, Nlarsm, Adam, and AdamHD, respectively. So based on our implementations, Adam optimizer had the best running time, but this result is subject to our implementation method.  

\paragraph{Default parameters.} Following the sensitivity analysis and the previous experiments, for float64 precision accuracy, we recommend the following default parameters: $k=1$, $c=10^{-30}$, $c^{\prime}=10^{-30}$, $b=1$, and $\rho=1$. In lower precision settings, such as float32, the adjustment of $c$ and $c^\prime$ is advised; for example, we recommend $c=c^{\prime}=10^{-19}$ for float32. Regarding $\lambda_0$, for models with a static environment or datasets, we suggest $\lambda_0 = 0.1$ or $\lambda_0 = 0.01$. For models where the environment changes stochastically and one-step gradient descent is needed, such as in DDQN, while $\lambda_0 = 0.1$ has shown reasonable convergence, better performance may be achieved with smaller learning rates, such as $\lambda_0 = 0.0001$.

%% file: sections/conclusions_future_work.tex
\section{Conclusion and future work}
\label{sec:conclusion}

We have introduced a new class of adaptive learning algorithms, called Nlar, based on a time-series modeling, where the convergence of the learning rates is proved in a general framework applicable to  even non-convex problems. We have explored in detail two special cases of Nlar, called Nlarcm and Nlarsm which utilize a cautious adaptive momentum scheme. Across multiple experiments, we have shown that Nlarcm and Nlarsm exhibit strong initial adaptability and consistency (especially in the first few epochs) even for large initial learning rates of 0.1 or higher. Overall, they outperformed Adam and AdamHD across a variety of learning rates.

The experiments suggest that algorithms based on Nlar are promising and hold significant potential for extension and combination with other methods. For example, exploring more sophisticated forms of \(c_{t}(d)\), \(\sigma_t^\prime(d)\), \(b_{t}(d)\) (rather than using constants), $\rho_t(d)$, and \(f\) presents an interesting avenue for future research. Additionally, investigating the effects of noise distributions other than the uniform distribution could yield valuable insights. Regarding the magnitude of the noise, similar to RL algorithms, incorporating a decay mechanism for \(\sigma_t^\prime(d)\) and $\sigma_t(d)$ to allow for early exploration is another potential area of investigation. Moreover, it is possible to adapt Nlar to initialize all learning rates (one per trainable variable) randomly, which could address the challenge of selecting the ideal initial learning rate, although the effectiveness of this approach would need to be verified. Finally, from a theoretical perspective, our update rules can be viewed similarly to the Kalman filtering concept proposed by \citet{kalman1960new}, where the learning rate is treated as a noisy observation. This connection to Kalman filtering presents another interesting avenue for future exploration.

%% file: sections/declaration.tex

\acks{
The author received no specific funding for this work.
}

%% file: sections/appendix.tex
\appendix

\renewcommand{\thetable}{A.\arabic{table}} 
\setcounter{table}{0} 

\section{Adagrad and Adadelta update rules}
\label{sec:app-overview}

For all $t=0,1,2,\dots$, the update rule of Adagrad is as follows:
\begin{equation}\label{eq:adagrad}
\theta_{t+1}(d) = \theta_{t}(d) - \dfrac{\gamma}{\sqrt{\sum_{s=0}^t(\frac{\partial L_s}{\partial \theta_{s}(d)})^2}}\dfrac{\partial L_t}{\partial \theta_{t}(d)},\text{ for all } d\text{ in }\mathcal D.  
\end{equation}
In this formula, $\dfrac{\gamma}{\sqrt{\sum_{s=0}^t(\frac{\partial L_s}{\partial \theta_{s}(d)})^2}}$ can be interpreted as the specific learning rate along the dimension  $d$.

In Adagrad's update rule \eqref{eq:adagrad}, the accumulation of the squared partial derivatives causes the learning rate to decrease over time, eventually approaching zero asymptotically.  Adadelta was introduced as the next adaptive algorithm to address the diminishing learning rates problem of Adagrad. Setting initial values $g_{0}=0$ and $\delta_{0}=0$, the update rule of Adadelta is as follows:
\begin{equation}
\label{eq:adadelta}
    \theta_{t+1}(d) = \theta_{t}(d) - \dfrac{\sqrt{(\delta_{t}(d))^2+\epsilon}}{\sqrt{(g_{t+1}(d))^2+\epsilon}}\dfrac{\partial L_t}{\partial {\theta_{t}(d)}}, t=0,1,2,\dots,
\end{equation}
where $\epsilon>0$ is a constant, $\rho$ is the decay rate, and $g_{t}(d)$ and $\delta_{t}(d)$ are given by
\begin{equation}
    (g_{t}(d))^2 = \rho (g_{t-1}(d))^2 + (1- \rho) \frac{\partial L_t}{\partial \theta_{t}(d)},\; t=1,2,3,\dots,
\end{equation}
\begin{equation}
    (\delta_{t}(d))^2 = \rho (\delta_{t-1}(d))^2+(1-\rho)(\theta_{t}(d) - \theta_{t-1}(d))^2,\; t=1,2,3,\dots.
\end{equation}


\section{Proofs}
In this section of the appendix, we provide the proofs. We begin with the proof of Theorem \ref{th:main} which is inspired by the proof of Theorem 1 of \citet{aase1983recursive}. 


\subsection{\textbf{Proof of Theorem \ref{th:main}}}
\label{proof:th-main}
\begin{proof}
Consider a fixed dimension $d$. By the assumption $\theta_{t+1}(d) = \theta_{t}(d) - \gamma(d)  f(\dfrac{\partial L_t}{\partial\theta_{t}(d)}) + \sigma_{t}(d)\epsilon_{t+1}(d),\; t=0, 1, 2,\dots$, which can be substituted in the estimator given by Equation \eqref{eq:ktheta}. After some rearrangements,  $\hat\gamma_{t+1}(d)$ is equal to:
\begin{equation}
\label{eq:proof1}
\begin{aligned}
    &= \dfrac{k_t^{'}(d) - \sum_{s=0}^t \left( (\sigma_{s}(d))^{-2} f(\dfrac{\partial L_s}{\partial\theta_{s}(d)})(-\gamma(d)  f(\dfrac{\partial L_s}{\partial\theta_{s}(d)}) + \sigma_{s}(d)\epsilon_{s+1}(d)\right)}{k_t(d) +\sum_{s=0}^t \left((\sigma_{ s}(d))^{-1}f(\dfrac{\partial L_s}{\partial\theta_{s}(d)})\right)^2}\\
    & = \dfrac{k_t^{'}(d) + \gamma(d) \sum_{s=0}^t  \sigma_{s}(d)^{-2} (f(\dfrac{\partial L_s}{\partial\theta_{s}(d)}))^2 - \sum_{s=0}^t W_s(d)}{k_t(d) +\sum_{s=0}^t \left((\sigma_{ s}(d))^{-1}f(\dfrac{\partial L_s}{\partial\theta_{s}(d)})\right)^2}\\
    & = \dfrac{k_t^{'}(d)/t + \gamma(d) [\sum_{s=0}^t  \left(\sigma_{s}(d)^{-1} f(\dfrac{\partial L_s}{\partial\theta_{s}(d)})\right)^2]/t -\dfrac{ \sum_{s=0}^t W_s(d)}{t}}{k_t(d)/t +[\sum_{s=0}^t \left((\sigma_{ s}(d))^{-1}f(\dfrac{\partial L_s}{\partial\theta_{s}(d)})\right)^2]/t}\\
\end{aligned},
\end{equation}
where 
$W_s(d) = (\sigma_{s}(d))^{-1}\epsilon_{s+1}(d)f(\dfrac{\partial L_s}{\partial\theta_{s}(d)}))$.

Next, we show that $\lim_{t\rightarrow\infty}\dfrac{\sum_{s=0}^t W_s(d)}{t}=0$. In order to do so, note that
\begin{equation*}
    \begin{aligned}
        \E[(W_s(d))^2/s^2] &=\E [\left((\sigma_{s}(d))^{-1}\epsilon_{s+1}(d) f(\dfrac{\partial L_s}{\partial\theta_{s}(d)})\right)^2/s^2]\\
        &=\E \left [\E[\left((\sigma_{s}(d))^{-1}\epsilon_{s+1}(d) f(\dfrac{\partial L_s}{\partial\theta_{s}(d)})\right)^2/s^2|\mathfrak F_{s}]\right]\\
        &=\E \left [((\sigma_{s}(d))^{-1}f(\dfrac{\partial L_s}{\partial\theta_{s}(d)}))^2\E[(\epsilon_{s+1}(d) )^2|\mathfrak F_{s}]\right]/s^2\\
        &=\E[(\epsilon_{s+1}(d))^2]\E \left [((\sigma_{s}(d))^{-1}f(\dfrac{\partial L_s}{\partial\theta_{s}(d)}))^2\right]/s^2\\
        &=e(d) \E \left [((\sigma_{s}(d))^{-1}f(\dfrac{\partial L_s}{\partial\theta_{s}(d)}))^2\right]/s^2
    \end{aligned}
\end{equation*}
where we have used the independence of $\epsilon_{s+1}(d)$  from $\mathfrak{F}_{s}$ and $e(d)=\mathbb{E}[\epsilon_{s+1}(d)^2]<\infty$, due to Assumption \ref{assump:epsilon}. Next, we simplify $\E \left [((\sigma_{s}(d))^{-1}f(\dfrac{\partial L_s}{\partial\theta_{s}(d)}))^2\right]/s^2$. Because of part (c) of Assumption \ref{assump:epsilon}, we suppose that the partial derivatives are non-zero, and from Assumption \ref{assump:grad}, we can suppose that $\sigma_{s}(d)=\min(c_{s}(d), |f(\dfrac{\partial L_s}{\partial\theta_{s}(d)})|)$. Then
\begin{equation*}
\begin{aligned}
    (\sigma_{s}(d))^{-1}|f(\dfrac{\partial L_s}{\partial\theta_{s}(d)})|=&\; (\sigma_{s}(d))^{-1}|f(\dfrac{\partial L_s}{\partial\theta_{s}(d)})|1_{\{|f(\dfrac{\partial L_s}{\partial\theta_{s}(d)})|\leq c_{s}(d)\}} +\\&\; (c_{s}(d))^{-1}|f(\dfrac{\partial L_s}{\partial\theta_{s}(d)})|1_{\{|f(\dfrac{\partial L_s}{\partial\theta_{s}(d)})| > c_{s}(d)\}}   \\
    \leq &\; 1_{\{|f(\dfrac{\partial L_s}{\partial\theta_{s}(d)})|\leq c_{s}(d)\}}  
    + (c_{s}(d))^{-1}|b_{s}(d)|1_{\{|f(\dfrac{\partial L_s}{\partial\theta_{s}(d)})| > c_{s}(d)\}}
\end{aligned},
\end{equation*}
where we have used $|f(\dfrac{\partial L_s}{\partial\theta_{s}(d)})|\leq |b_s(d)|$ and the fact that partial derivatives are non-zero. The last two terms of the above equation are mutually exclusive, and so we obtain:
\begin{equation*}
    \begin{aligned}
        ((\sigma_{s}(d))^{-1}f(\dfrac{\partial L_s}{\partial\theta_{s}(d)}))^2 
        \leq& 1+((c_{s}(d))^{-1}b_{s}(d))^2
    \end{aligned}
\end{equation*}

Hence, by Assumptions \ref{assump:epsilon} and \ref{assump:grad}, we obtain $$ \sum_{s=1}^t \E [\left((\sigma_{s}(d))^{-1}\epsilon_{s+1}(d) f(\dfrac{\partial L_s}{\partial\theta_{s}(d)})\right)^2/s^2]\leq e(d) \sum_{s=1}^\infty (1+((c_{s}(d))^{-1}b_{s}(d))^2)/s^2<\infty.$$
Therefore, $\sum_{s\geq 1}\E[(W_s(d))^2]/s^2<\infty$, also, $\E[W_s(d)]=0$,  and so, by Proposition IV.6.1 of \citet{neveu1965mathematical}, $\dfrac{\sum_{s=0}^t W_s(d)}{t} = \dfrac{ \sum_{s=1}^t (\sigma_{s}(d))^{-1}\epsilon_{s+1}(d)f(\dfrac{\partial L_s}{\partial\theta_{s}(d)})}{t}$ converges to zero as $t\rightarrow\infty$.

On the other hand, if $\sigma_{s}(d)=\min(c_{s}(d), |f(\dfrac{\partial L_s}{\partial\theta_{s}(d)})|)$, then $\dfrac{\sum_{s=0}^t \left((\sigma_{ s}(d))^{-1}f(\dfrac{\partial L_s}{\partial\theta_{s}(d)})\right)^2}{t}$ is equal to
\begin{equation}\label{eq:sumnonzero}
\begin{aligned}
    &=\dfrac{\sum_{s=0}^t 1_{|f(\dfrac{\partial L_s}{\partial\theta_{s}(d)})|\leq c_{s}(d)} + \sum_{s=0}^t \left((c_{ s}(d))^{-1} f(\dfrac{\partial L_s}{\partial\theta_{s}(d)})\right)^21_{|f(\dfrac{\partial L_s}{\partial\theta_{s}(d)})| > c_{s}(d)}}{t}\\
    &\geq \dfrac{\sum_{s=0}^t 1_{|f(\dfrac{\partial L_s}{\partial\theta_{s}(d)})|\leq c_{s}(d)}+\sum_{s=0}^t 1_{|f(\dfrac{\partial L_s}{\partial\theta_{s}(d)})| >c_{s}(d)}}{t}\\
    &=\dfrac{\sum_{s=0}^t 1}{t}=1,
\end{aligned}
\end{equation}
where we have used $(c_{s}(d))^{-1} f(\dfrac{\partial L_s}{\partial\theta_{s}(d)}) \geq 1$, on $\{|f(\dfrac{\partial L_s}{\partial\theta_{s}(d)})| > c_{s}(d)\}$, and the fact that by part (c) of Assumption \ref{assump:epsilon}, other than a finite number of steps, the partial derivatives are non-zero. This leads to \[\liminf_{t\rightarrow\infty}\sum_{s=0}^t \left((\sigma_{ s}(d))^{-1}f(\dfrac{\partial L_s}{\partial\theta_{s}(d)})\right)^2/t>0,\] and so from Equation \eqref{eq:proof1}, we have $\hat\gamma_{t+1}(d) \rightarrow\gamma(d) $, as $t\rightarrow\infty$ which concludes the proof.

\end{proof}


\subsection{\textbf{Proof of Theorem \ref{th:momentum}}}
\label{proof:th-momentum}
\begin{proof}
The proof of this theorem, follows the same lines as in the proof of Theorem \ref{th:main}, and so we only provide an outline. Our update rule for the corresponding parameter is $\theta_{t+1}(d) = \theta_{t}(d) + \rho_t(d)v_{t}(d) - \gamma(d)  f(\dfrac{\partial L_t}{\partial\theta_{t}(d)}) + \sigma_{t}(d)\epsilon_{t+1}(d),\; t=0, 1, 2,\dots$. This can be substituted in the estimator given by \eqref{eq:ktheta3} which after some rearrangements,  $\hat{\zeta}_{t+1}(d)$ is equal to:
\begin{equation*}
\begin{aligned}
    \hat{\zeta}_{t+1}(d) = \dfrac{k\hat{\zeta}_0(d)/t + \gamma(d) [\sum_{s=0}^t  \left((\sigma_{s}(d))^{-1} f(\dfrac{\partial L_s}{\partial\theta_{s}(d)})\right)^2]/t -\dfrac{ \sum_{s=0}^t W_s}{t}  -S_t}{k/t +[\sum_{s=0}^t \left((\sigma_{ s}(d))^{-1}f(\dfrac{\partial L_s}{\partial\theta_{s}(d)})\right)^2]/t}.\\
\end{aligned}
\end{equation*}
where $$W_s=(\sigma_{s}(d))^{-1}\epsilon_{s+1}(d)f(\dfrac{\partial L_s}{\partial\theta_{s}(d)})$$ and $$S_t = \dfrac{ \sum_{s=0}^t (\sigma_{s}(d))^{-2}\rho_s(d)v_s(d)f(\dfrac{\partial L_s}{\partial\theta_{s}(d)})}{t}.$$
Since Assumptions \ref{assump:epsilon}, \ref{assump:grad}, and \ref{assump:kprime} hold, from the same lines of proof as in Theorem \ref{th:main}, we have that $\lim_{t\rightarrow\infty}\dfrac{ \sum_{s=0}^t W_s}{t}=\lim_{t\rightarrow\infty}\dfrac{ \sum_{s=0}^t (\sigma_{s}(d))^{-1}\epsilon_{s+1}(d)f(\dfrac{\partial L_s}{\partial\theta_{s}(d)})}{t}=0$. 

Also, note that $|S_t|\leq  \dfrac{ \sum_{s=0}^t (\sigma_{s}(d))^{-2} |\rho_s(d)||v_s(d)||f(\dfrac{\partial L_s}{\partial\theta_{s}(d)})|}{t}$, where due to Kronecker's test $\dfrac{ \sum_{s=0}^t (\sigma_{s}(d))^{-2} |\rho_s(d)||v_s(d)||f(\dfrac{\partial L_s}{\partial\theta_{s}(d)})|}{t}$, converges to zero (as $t\rightarrow\infty$) if \[\sum_{s=1}^\infty\dfrac{  (\sigma_{s}(d))^{-2} |\rho_s(d)||v_s(d)||f(\dfrac{\partial L_s}{\partial\theta_{s}(d)})|}{s}\leq\sum_{s=1}^\infty\dfrac{  (\sigma_{s}(d))^{-2} |\rho_s(d)||v_s(d)||b_s(d)|}{s}<\infty.\] 
The last inequality holds by the assumption, and this shows that  $\lim_{t\rightarrow\infty}S_t=0$.

On the other hand, similar to the proof of Theorem \ref{th:main}, \[\liminf_{t\rightarrow\infty}\sum_{s=0}^t \left((\sigma_{ s}(d))^{-1}f(\dfrac{\partial L_s}{\partial\theta_{s}(d)})\right)^2/t>0,\] therefore, we have $\hat{\zeta}_{t+1}(d) \rightarrow\gamma(d) $, as $t\rightarrow\infty$ which concludes the proof.
\end{proof}


\subsection{\textbf{Proof of Proposition \ref{prop:Nlarsm}}}
\label{proof:prop-Nlarsm}
\begin{proof}
In the case of Nlarsm, the update rule along the dimension $d$ is $\theta_{t+1}(d) = \theta_{t}(d) + \rho_t(d)v_{t}(d) - \gamma(d)  f(\dfrac{\partial L_t}{\partial\theta_{t}(d)}) + \sigma_{t}^{'}(d)\epsilon_{t+1}(d),\; t=0, 1, 2,\dots$. This can be substituted in the estimator given by Equation \eqref{eq:zetastar} which after some rearrangements,  $\hat{\zeta^\star  }_{t+1}(d)$ becomes:
\begin{equation*}
\begin{aligned}
    \hat{\zeta^\star}_{t+1}(d) = \dfrac{k \hat{\zeta}_{0}^\star(d) /t + \gamma(d) [\sum_{s=0}^t  \left( f(\dfrac{\partial L_s}{\partial\theta_{s}(d)})\right)^2]/t -\dfrac{ \sum_{s=0}^t W_s}{t}  -S_t}{k/t +[\sum_{s=0}^t \left(f(\dfrac{\partial L_s}{\partial\theta_{s}(d)})\right)^2]/t},\\
\end{aligned}
\end{equation*}
where $$W_s=\sigma_{s}^{'}(d)\epsilon_{s+1}(d)f(\dfrac{\partial L_s}{\partial\theta_{s}(d)})$$ and $$S_t = \dfrac{ \sum_{s=0}^t \rho_s(d)v_s(d)f(\dfrac{\partial L_s}{\partial\theta_{s}(d)})}{t}.$$
Since Assumption \ref{assump:epsilon} holds, from the same lines of proof as in Theorem \ref{th:main}, we have that $\lim_{t\rightarrow\infty}\dfrac{ \sum_{s=0}^t W_s}{t}=\lim_{t\rightarrow\infty}\dfrac{ \sum_{s=0}^t \sigma_{s}^{'}(d)\epsilon_{s+1}(d)f(\dfrac{\partial L_s}{\partial\theta_{s}(d)})}{t}=0$, because by the assumption we have $\sum_{s=1}^\infty\E[(\sigma_s^{'}(d))^{2}]/s^2<\infty$. 

Also, note that $|S_t|\leq  \dfrac{ \sum_{s=0}^t (|\rho_s(d)||v_s(d)||f(\dfrac{\partial L_s}{\partial\theta_{s}(d)})|)}{t}$, where due to Kronecker's test $\dfrac{ \sum_{s=0}^t ( |\rho_s(d)||v_s(d)||f(\dfrac{\partial L_s}{\partial\theta_{s}(d)})|)}{t}$, converges to zero (as $t\rightarrow\infty$) if \[\sum_{s=1}^\infty\dfrac{   |\rho_s(d)||v_s(d)||f(\dfrac{\partial L_s}{\partial\theta_{s}(d)})|}{s}<B\sum_{s=1}^\infty\dfrac{  |\rho_s(d)||v_s(d)|}{s}<\infty.\] 
The last inequality holds by the assumption, and this shows that  $\lim_{t\rightarrow\infty}S_t=0$.

On the other hand, from the assumption, we have that $0<B^{'}<|f(\dfrac{\partial L_t}{\partial\theta_{t}(d)})|$. This leads to $\liminf_{t\rightarrow\infty}\sum_{s=0}^t \left(f(\dfrac{\partial L_s}{\partial\theta_{s}(d)})\right)^2/t>(B^{'})^2\liminf_{t\rightarrow\infty}\sum_{s=0}^t 1/t=(B^{'})^2>0$, therefore, we have $\hat{\zeta}_{t+1}(d) \rightarrow\gamma(d) $, as $t\rightarrow\infty$ which concludes the proof.
\end{proof}


\section{Details of datasets and the reinforcement learning environment}
\label{sec:app-data}


\subsection{Details of datasets and environments}
\paragraph{Modified NIST (MNIST).} 
This is a large database of handwritten digits that is commonly used for training and testing ML models \citep{lecun1998gradient}. Each sample is a black and white image of a handwritten digit. Furthermore, the black and white images are normalized to fit into a $28\times 28$ pixel bounding box. The entire MNIST dataset contains 70,000 images. 

\paragraph{Canadian Institute for Advanced Research, 10 classes (CIFAR10).} This is a dataset consisting of 60,000 color images normalized to fit into a $32\times 32$ pixel bounding box,  \citep{krizhevsky2009learning}. There are  10 classes in total with 6,000 images in each class.

\paragraph{CartPole-v0.}
The CartPole-v0 game is a classical control problem. In our experiments, we have used the environment of OpenAI Gym\footnote{\url{https://gymnasium.farama.org/environments/classic_control/cart_pole/}} that simulates a pole balanced on a cart. The goal is to prevent the pole from falling over by moving the cart either left or right. Hence, the action space is discrete, with two possible actions. The observation space consists of a 4-dimensional array representing the cart's position and velocity, along with the pole's angle and velocity. A reward of +1 is given for every time step that the pole remains upright. The episode starts with the pole slightly tilted, ending if either the pole angle is greater than $\pm12$ degrees, the cart position is greater than $\pm2.4$ (center of the cart reaches the edge of the display), or the episode length is greater than 200 (500 for v1).


\section{Learning curves, complimentary results, and discussions}
\label{sec:app-plots}
Our experiments are conducted using values of $\beta$ in $\{10^{-7}, 10^{-4}\}$ (for the RL experiment, the value of $\beta=10^{-8}$ is also tried). In Section \ref{sec:discussion}, the presented results are for the value of $\beta$ at which AdamHD exhibited the best performance among these values of $\beta$. In this section, we provide the comparison of Nlarsm and Nlarcm  with AdamHD for values of  $\beta$ where AdamHD does not achieve the best performance. This comparison is made across various learning rates, using Logistic regression, MLP2h, MLP7h, VGG11 models on the MNIST and CIFAR10 datasets.


\subsection{Logistic Regression, MLP2h, and MLP7h}

\begin{figure*}[!ht]
  \centering
  \includegraphics[width=0.22\linewidth, trim=0 10 30 0]{imgs/mnist_logistic_nlars_adam_mu=1_beta=None_minlr=None_50_0_1_300val_accuracy.pdf}\hspace{-0.5mm}%
  \includegraphics[width=0.22\linewidth, trim=0 10 30 0]{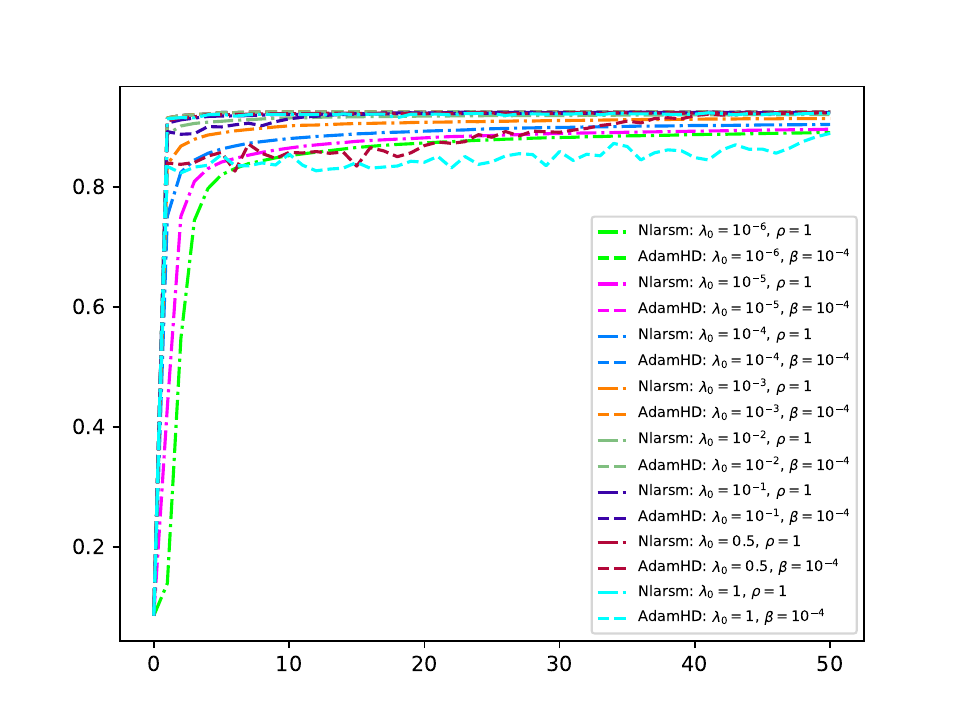}\hspace{-0.5mm}%
  \includegraphics[width=0.22\linewidth, trim=0 10 30 0]{imgs/mnist_logistic_nlarc_adam_mu=1_beta=None_minlr=None_50_0_1_300val_accuracy.pdf}\hspace{-0.5mm}%
  \includegraphics[width=0.22\linewidth, trim=0 10 30 0]{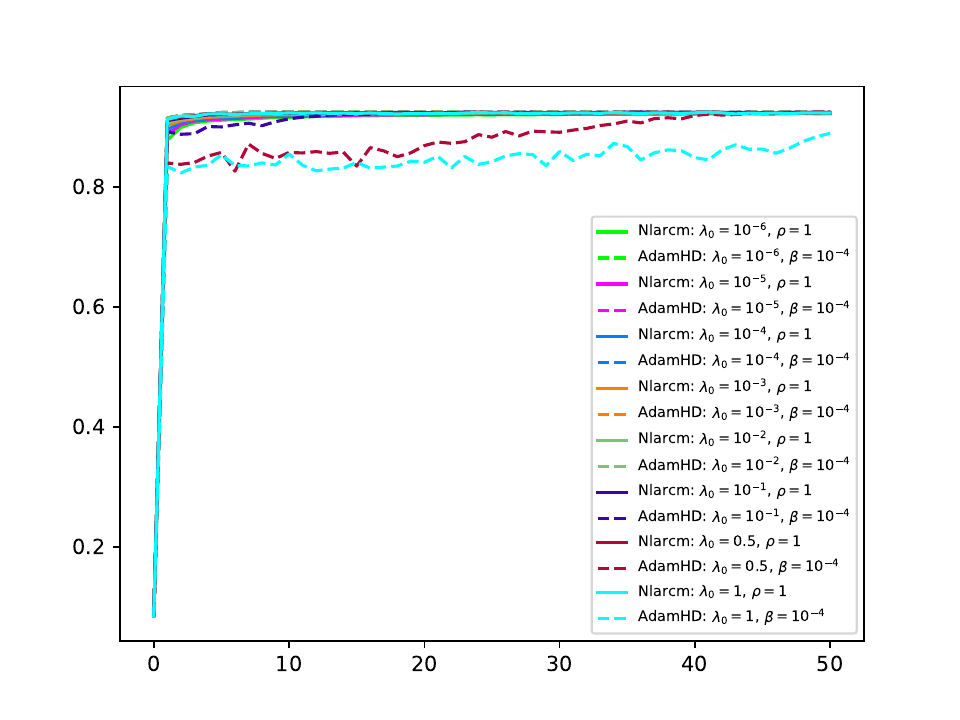}

  \includegraphics[width=0.22\linewidth, trim=0 10 30 0]{imgs/mnist_logistic_nlars_adam_mu=1_beta=None_minlr=None_50_0_1_300accuracy.pdf}\hspace{-0.5mm}%
  \includegraphics[width=0.22\linewidth, trim=0 10 30 0]{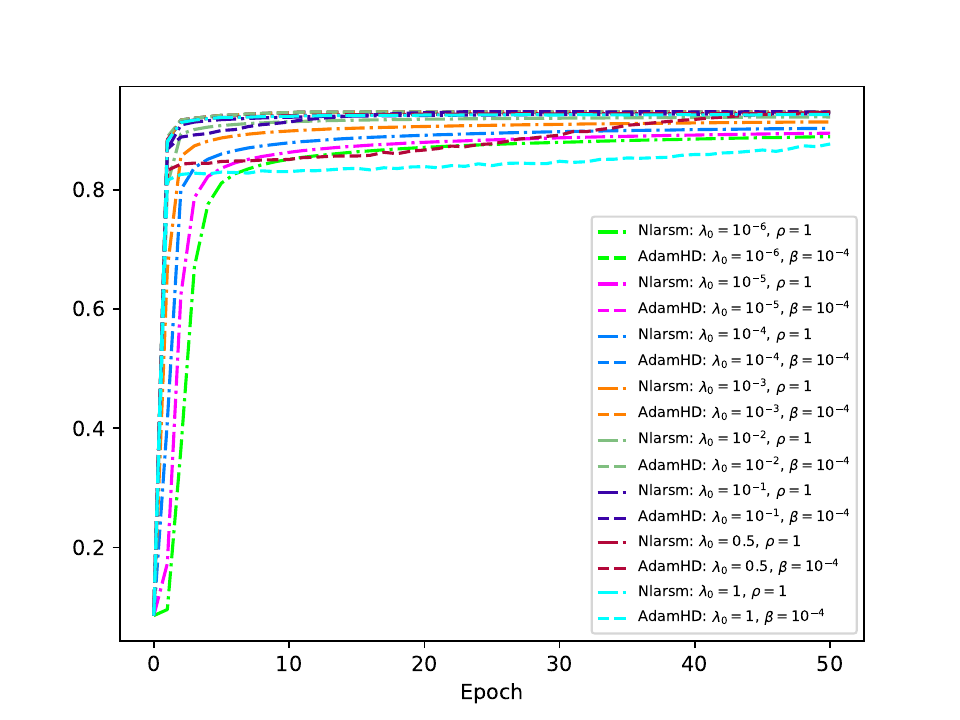}\hspace{-0.5mm}%
  \includegraphics[width=0.22\linewidth, trim=0 10 30 0]{imgs/mnist_logistic_nlarc_adam_mu=1_beta=None_minlr=None_50_0_1_300accuracy.pdf}\hspace{-0.5mm}%
  \includegraphics[width=0.22\linewidth, trim=0 10 30 0]{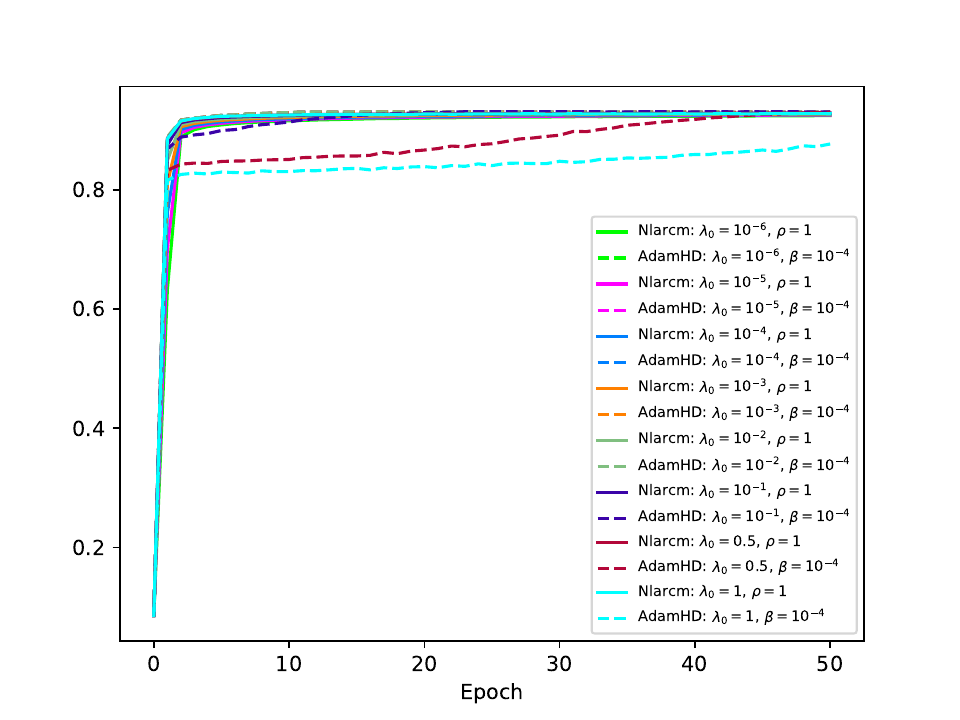}

    \vspace{1\baselineskip}
  \caption{Logistic regression model on the MNIST dataset: Performance comparison of Nlarsm and Nlarcm versus Adam and AdamHD (with $\beta = 10^{-4}$) across varied learning rates.}
  \vspace{1\baselineskip}

  \vspace{1\baselineskip}
  \label{fig:exp-logistic-mnist-nlar-adam-adamhd-1e-4}
\end{figure*}

\begin{figure*}[!ht]
  \centering
  \includegraphics[width=0.22\linewidth, trim=0 10 30 0]{imgs/mnist_mlp_nlars_adam_mu=1_beta=None_minlr=None_50_0_1_300val_accuracy.pdf}\hspace{-0.5mm}%
  \includegraphics[width=0.22\linewidth, trim=0 10 30 0]{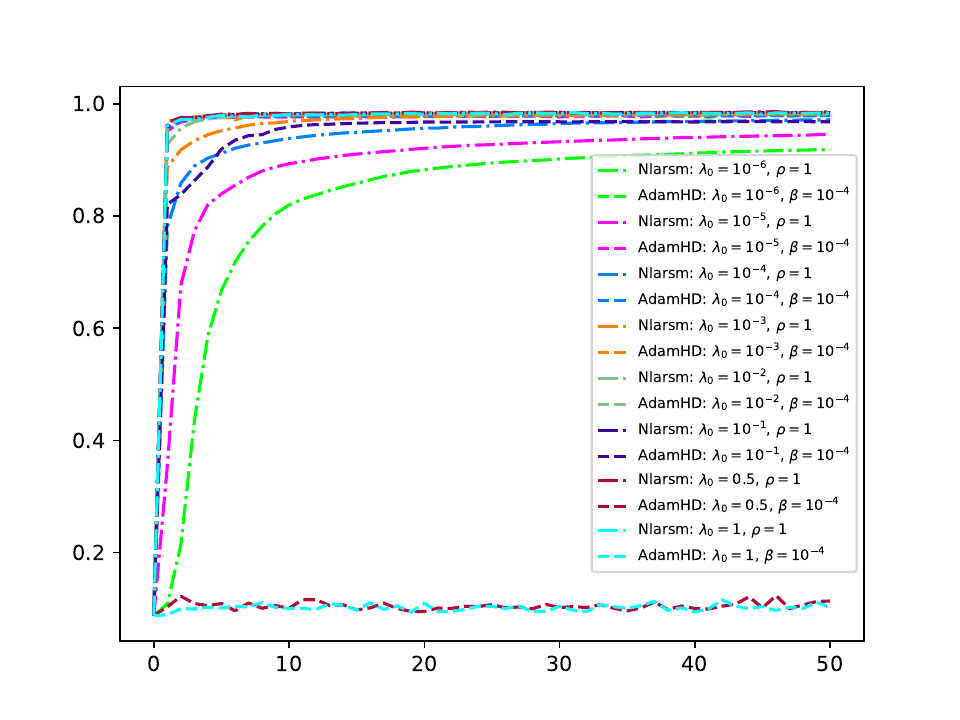}\hspace{-0.5mm}%
  \includegraphics[width=0.22\linewidth, trim=0 10 30 0]{imgs/mnist_mlp_nlarc_adam_mu=1_beta=None_minlr=None_50_0_1_300val_accuracy.pdf}\hspace{-0.5mm}%
  \includegraphics[width=0.22\linewidth, trim=0 10 30 0]{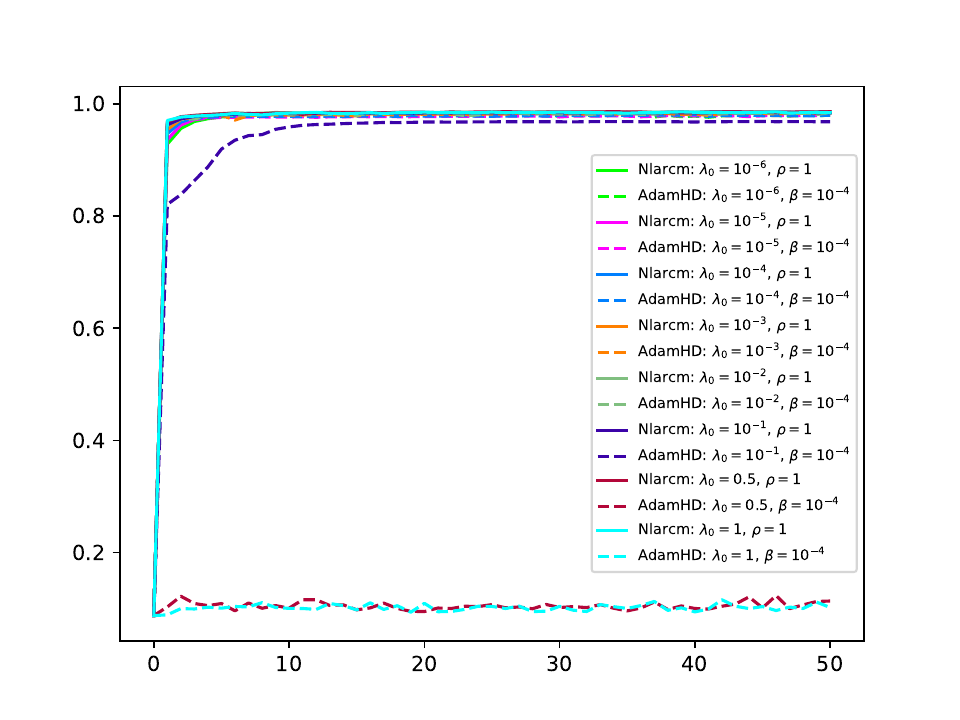}

  \includegraphics[width=0.22\linewidth, trim=0 10 30 0]{imgs/mnist_mlp_nlars_adam_mu=1_beta=None_minlr=None_50_0_1_300accuracy.pdf}\hspace{-0.5mm}%
  \includegraphics[width=0.22\linewidth, trim=0 10 30 0]{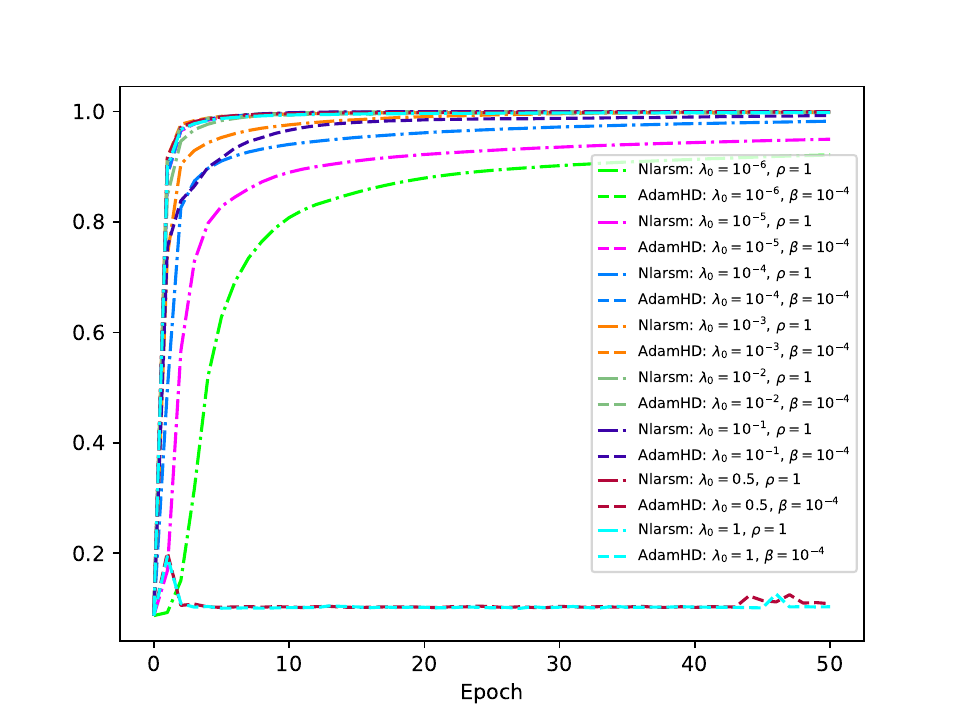}\hspace{-0.5mm}%
  \includegraphics[width=0.22\linewidth, trim=0 10 30 0]{imgs/mnist_mlp_nlarc_adam_mu=1_beta=None_minlr=None_50_0_1_300accuracy.pdf}\hspace{-0.5mm}%
  \includegraphics[width=0.22\linewidth, trim=0 10 30 0]{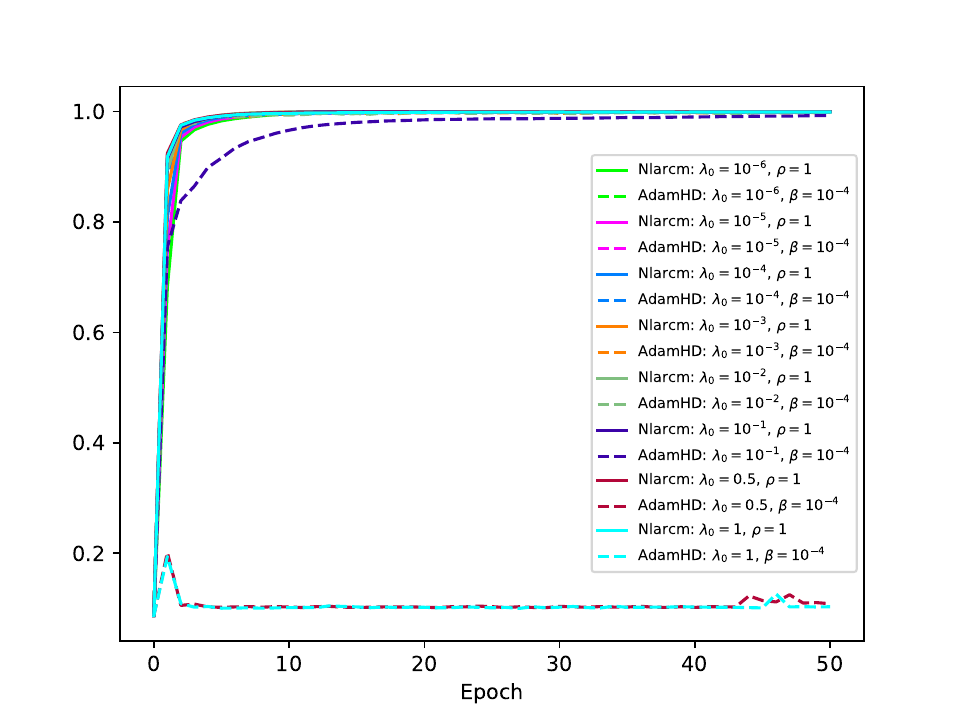}

    \vspace{1\baselineskip}
  \caption{MLP2h on the MNIST dataset: Performance comparison of Nlarsm and Nlarcm versus Adam and AdamHD (with $\beta = 10^{-4}$) across varied learning rates. }
  \vspace{1\baselineskip}
  \label{fig:exp-mlp-mnist-nlar-adam-adamhd-1e-4}
\end{figure*}

\begin{figure*}[!ht]
  \centering
  \includegraphics[width=0.22\linewidth, trim=0 10 30 0]{imgs/CIFAR10_mlp7h_nlars_adam_mu=1_beta=None_minlr=None_50_0_1_300val_accuracy.pdf}\hspace{-0.5mm}%
  \includegraphics[width=0.22\linewidth, trim=0 10 30 0]{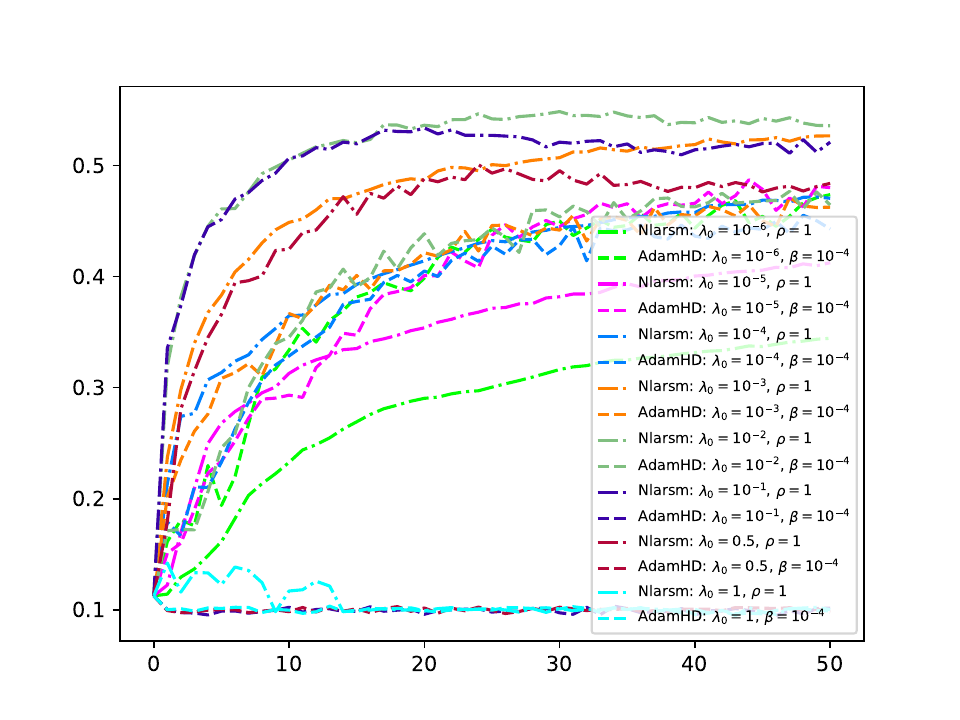}\hspace{-0.5mm}%
  \includegraphics[width=0.22\linewidth, trim=0 10 30 0]{imgs/CIFAR10_mlp7h_nlarc_adam_mu=1_beta=None_minlr=None_50_0_1_300val_accuracy.pdf}\hspace{-0.5mm}%
  \includegraphics[width=0.22\linewidth, trim=0 10 30 0]{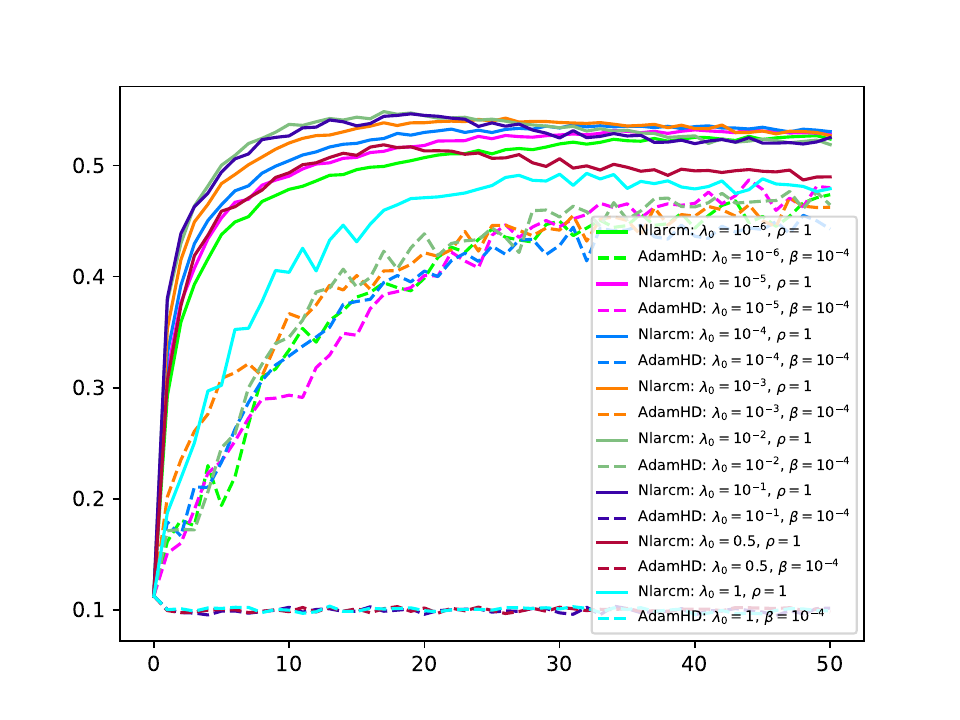}

  \includegraphics[width=0.22\linewidth, trim=0 10 30 0]{imgs/CIFAR10_mlp7h_nlars_adam_mu=1_beta=None_minlr=None_50_0_1_300accuracy.pdf}\hspace{-0.5mm}%
  \includegraphics[width=0.22\linewidth, trim=0 10 30 0]{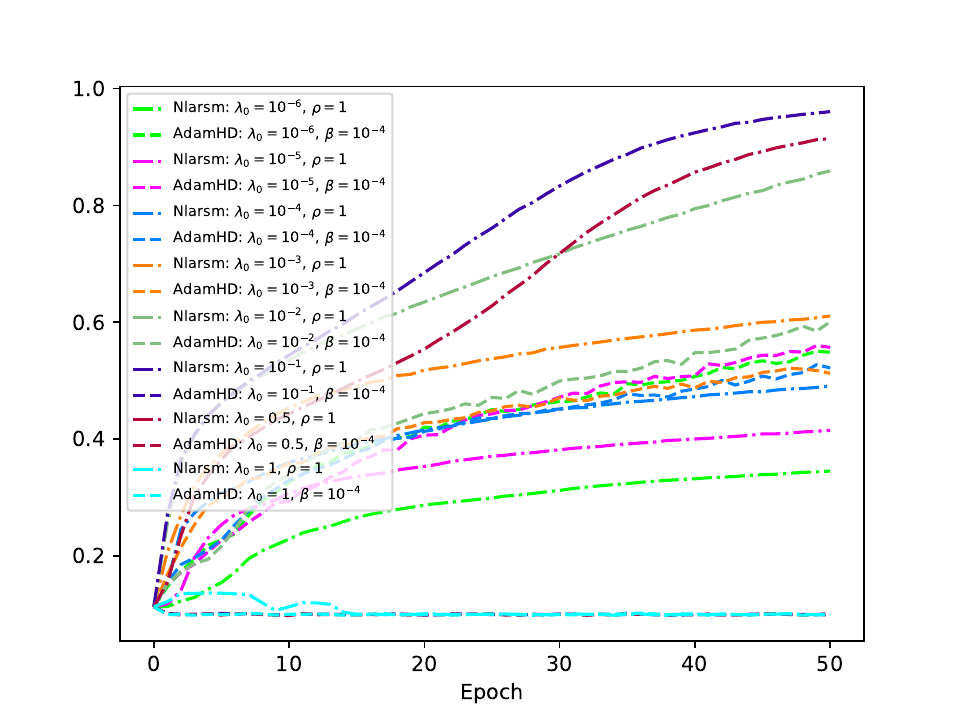}\hspace{-0.5mm}%
  \includegraphics[width=0.22\linewidth, trim=0 10 30 0]{imgs/CIFAR10_mlp7h_nlarc_adam_mu=1_beta=None_minlr=None_50_0_1_300accuracy.pdf}\hspace{-0.5mm}%
  \includegraphics[width=0.22\linewidth, trim=0 10 30 0]{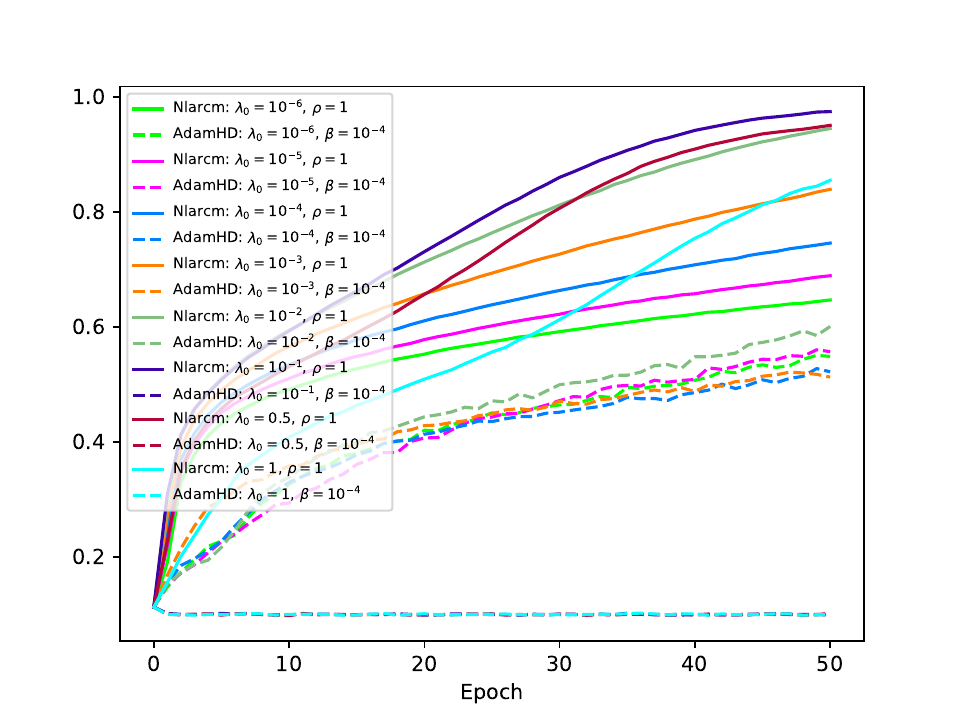}

    \vspace{1\baselineskip}
  \caption{MLP7h on the CIFAR10 dataset: Performance comparison of Nlarsm and Nlarcm versus Adam and AdamHD (with $\beta = 10^{-4}$) across varied learning rates. }
  \vspace{1\baselineskip}
  \label{fig:exp-mlp7h-CIFAR10-nlar-adam-adamhd-1e-4}
\end{figure*}

Figure \ref{fig:exp-logistic-mnist-nlar-adam-adamhd-1e-4} compares the performance of Nlarsm and Nlarcm versus Adam and AdamHD (with $\beta = 10^{-4}$) for Logistic regression model on the MNIST dataset across varied learning rates. Figures \ref{fig:exp-mlp-mnist-nlar-adam-adamhd-1e-4} and \ref{fig:exp-mlp7h-CIFAR10-nlar-adam-adamhd-1e-4} provide a similar comparison for MLP2h and MLP7h on the MNIST dataset.

\subsection{The VGG11 model on the CIFAR10 dataset}
\label{sec:vgg11-CIFAR10}

\paragraph{The description of the VGG11 model.}  The VGG11 model is structured as a sequential configuration consisting of 5 blocks, each made of a specific number of convolutional layers followed by max-pooling and batch normalization layers (making a total of 8 convolutional layers). The 5 blocks are then concluded with 3 fully connected layers. So in total, there are 8 convolutional layers plus 3 fully connected ones, hence the name VGG11 is applied. The details are as follows:

\begin{itemize}
    \item Convolutional Blocks:
\begin{itemize}
    \item Block 1: Starts with a convolutional layer of 64 filters $(3\times3)$, activated by the ReLU function, with padding set to ``same'' and $L_2$ regularization. This layer processes input images of size $32\times32$ with 3 channels. It is followed by a $2\times 2$ max-pooling layer and a batch normalization layer.
    \item Block 2: Features a single convolutional layer with 128 filters, similar padding, activation, and regularization settings, followed by the same sequence of max-pooling and batch normalization.
    \item Block 3: Comprises two convolutional layers each with 256 filters, followed by max-pooling and batch normalization.
    \item Blocks 4 and 5: Each block contains two convolutional layers with 512 filters, maintaining the same activation, padding, and regularization. Each block ends with a max-pooling and batch normalization layer.
\end{itemize}
\item Flattening and Dense Layers: The flattened output is processed through three dense layers. The first two dense layers each containing 4096 units with ReLU activation and $L_2$ regularization, designed to capture the high-level patterns in the data.
The final dense layer consists of 10 units with a softmax activation function, designated for classifying the inputs into one of ten possible categories.

\end{itemize}

\begin{figure*}[!ht]
  \centering
  \includegraphics[width=0.22\linewidth, trim=0 10 30 0]{imgs/CIFAR10_vgg11_nlars_adam_mu=1_beta=None_minlr=None_50_0_1_300val_accuracy.pdf}\hspace{-0.5mm}%
  \includegraphics[width=0.22\linewidth, trim=0 10 30 0]{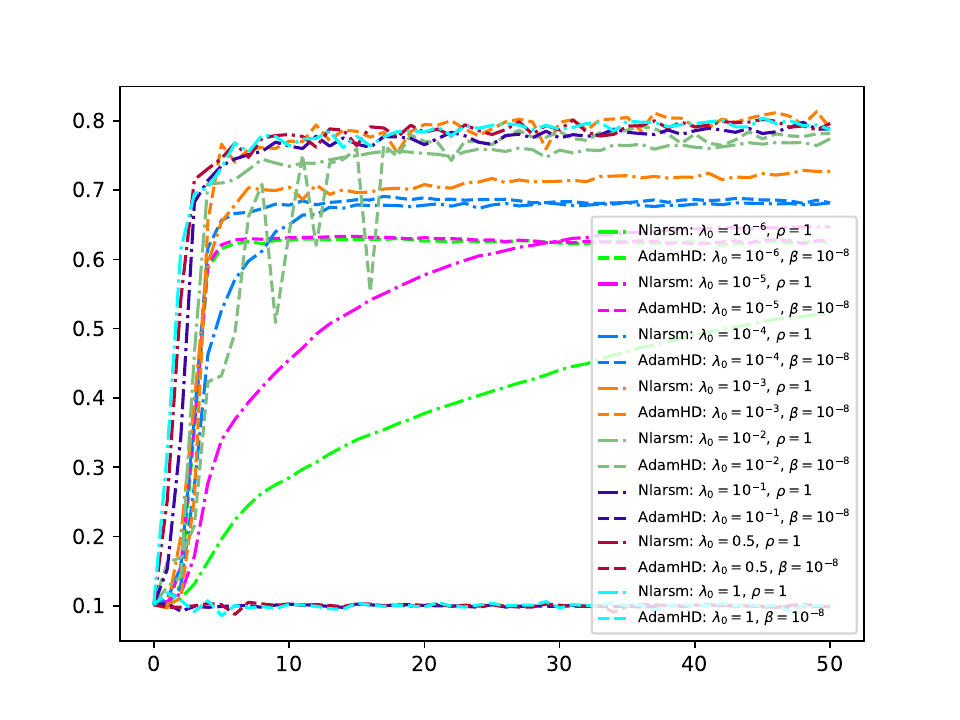}\hspace{-0.5mm}%
  \includegraphics[width=0.22\linewidth, trim=0 10 30 0]{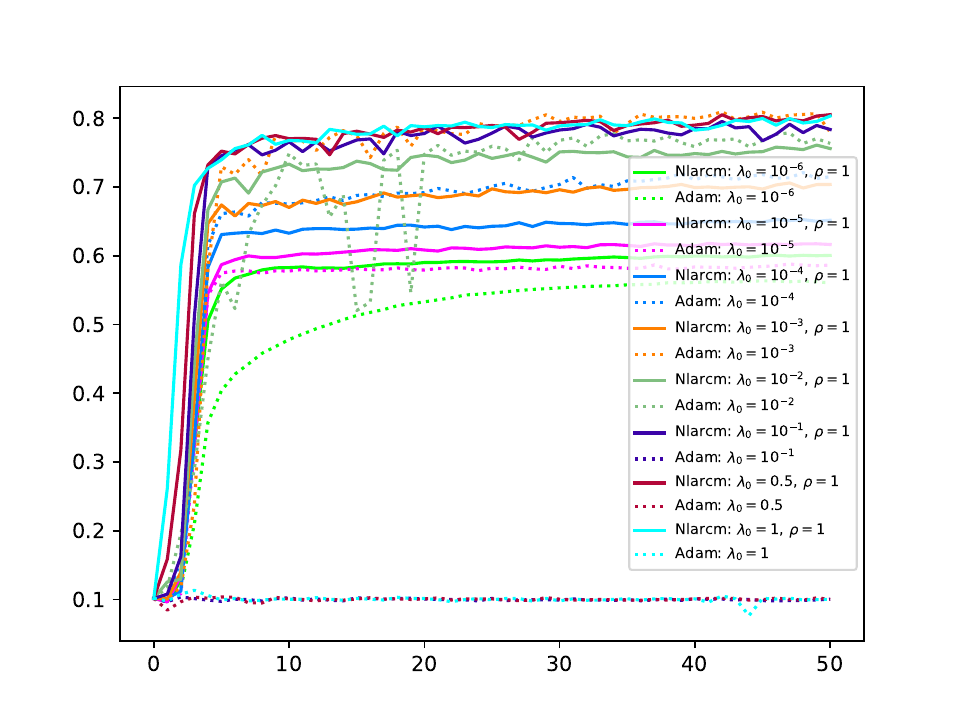}\hspace{-0.5mm}%
  \includegraphics[width=0.22\linewidth, trim=0 10 30 0]{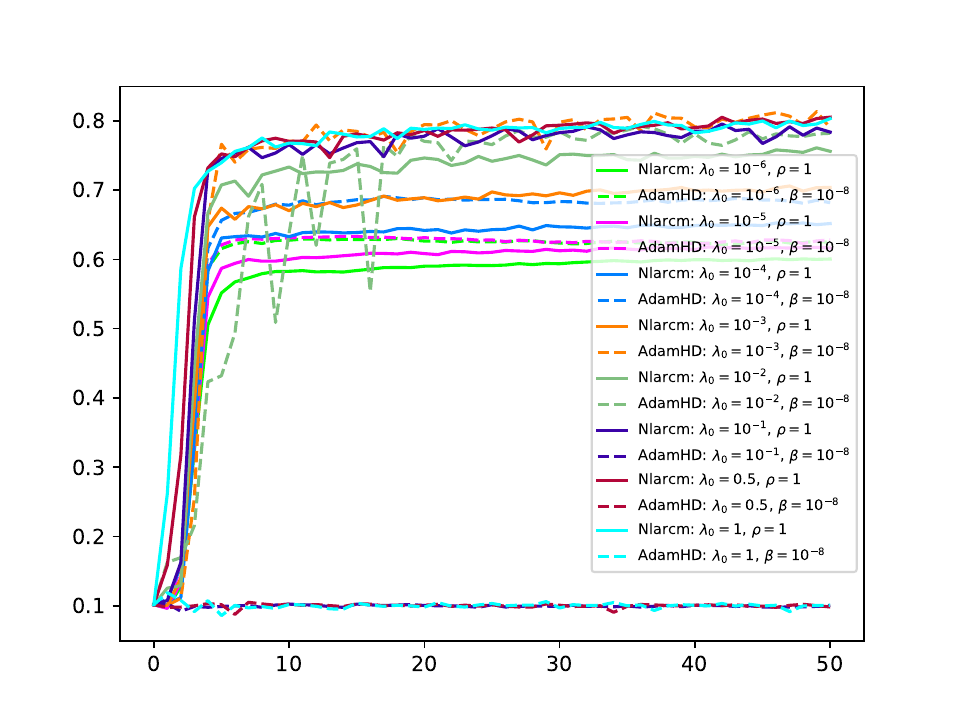}

  \includegraphics[width=0.22\linewidth, trim=0 10 30 0]{imgs/CIFAR10_vgg11_nlars_adam_mu=1_beta=None_minlr=None_50_0_1_300accuracy.pdf}\hspace{-0.5mm}%
  \includegraphics[width=0.22\linewidth, trim=0 10 30 0]{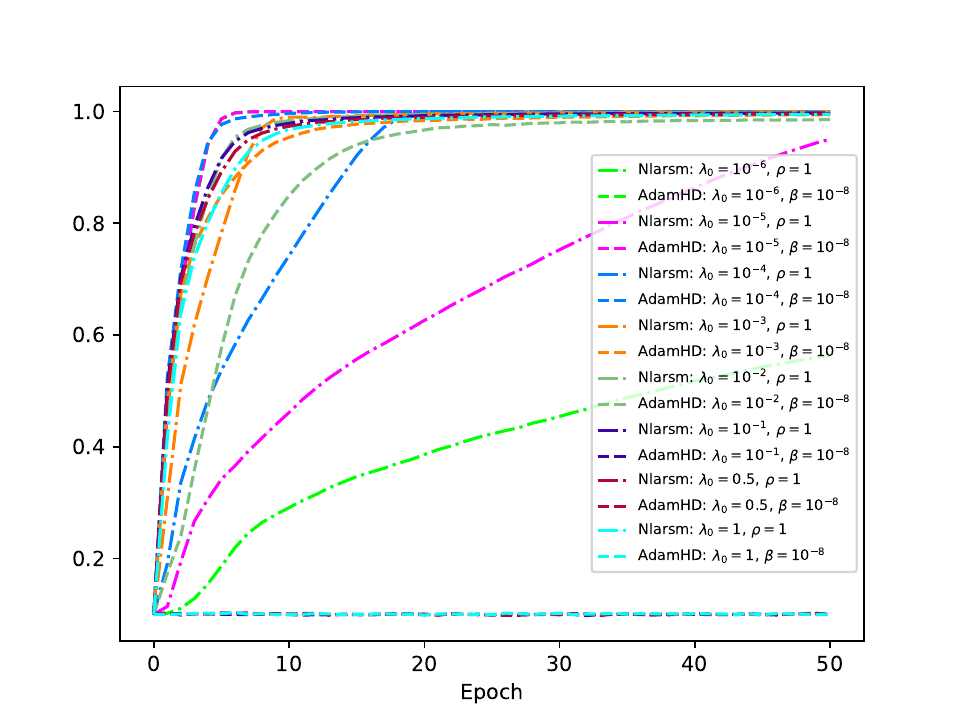}\hspace{-0.5mm}%
  \includegraphics[width=0.22\linewidth, trim=0 10 30 0]{imgs/CIFAR10_vgg11_nlarc_adam_mu=1_beta=None_minlr=None_50_0_1_300accuracy.pdf}\hspace{-0.5mm}%
  \includegraphics[width=0.22\linewidth, trim=0 10 30 0]{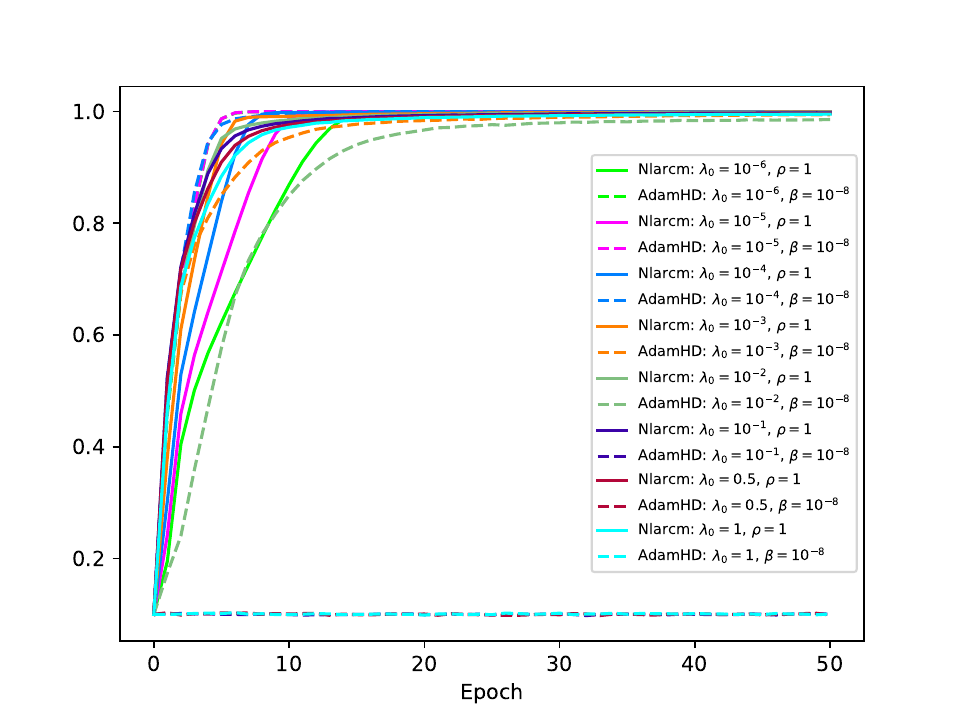}

    \vspace{1\baselineskip}
  \caption{VGG11 on the CIFAR10 dataset: Performance comparison of Nlarsm and Nlarcm versus Adam and AdamHD (with $\beta = 10^{-8}$) across varied learning rates. }
  \vspace{1\baselineskip}
  \label{fig:exp-vgg11-CIFAR10-nlar-adam-adamhd-1e-8}
\end{figure*}

\begin{figure*}[!ht]
  \centering
  \includegraphics[width=0.22\linewidth, trim=0 10 30 0]{imgs/CIFAR10_vgg11_nlars_adam_mu=1_beta=None_minlr=None_50_0_1_300val_accuracy.pdf}\hspace{-0.5mm}%
  \includegraphics[width=0.22\linewidth, trim=0 10 30 0]{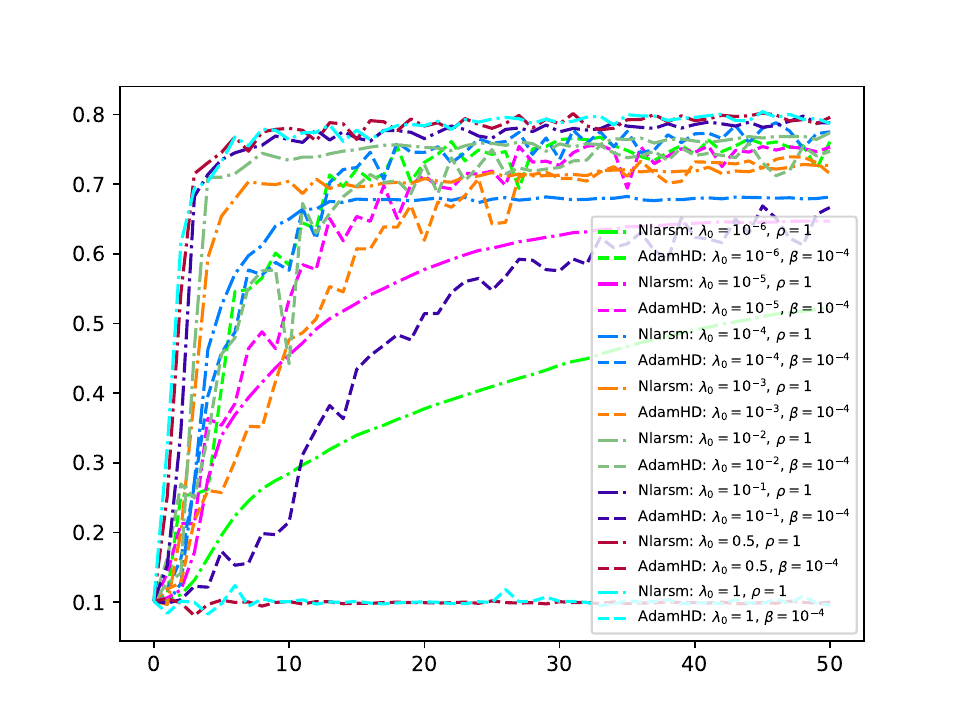}\hspace{-0.5mm}%
  \includegraphics[width=0.22\linewidth, trim=0 10 30 0]{imgs/CIFAR10_vgg11_nlarc_adam_mu=1_beta=None_minlr=None_50_0_1_300val_accuracy.pdf}\hspace{-0.5mm}%
  \includegraphics[width=0.22\linewidth, trim=0 10 30 0]{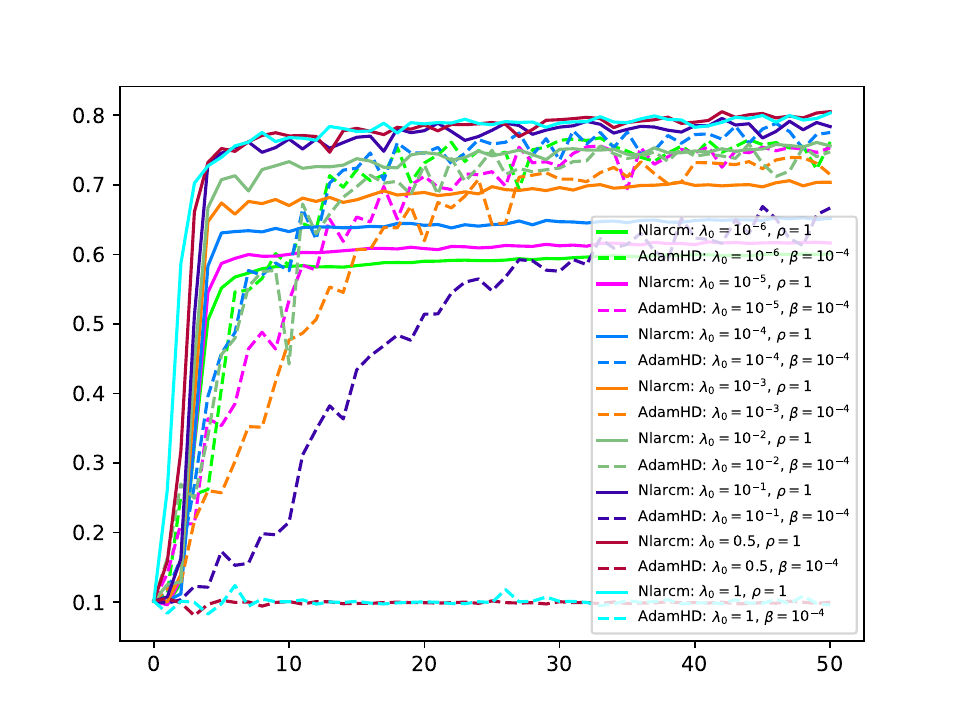}

  \includegraphics[width=0.22\linewidth, trim=0 10 30 0]{imgs/CIFAR10_vgg11_nlars_adam_mu=1_beta=None_minlr=None_50_0_1_300accuracy.pdf}\hspace{-0.5mm}%
  \includegraphics[width=0.22\linewidth, trim=0 10 30 0]{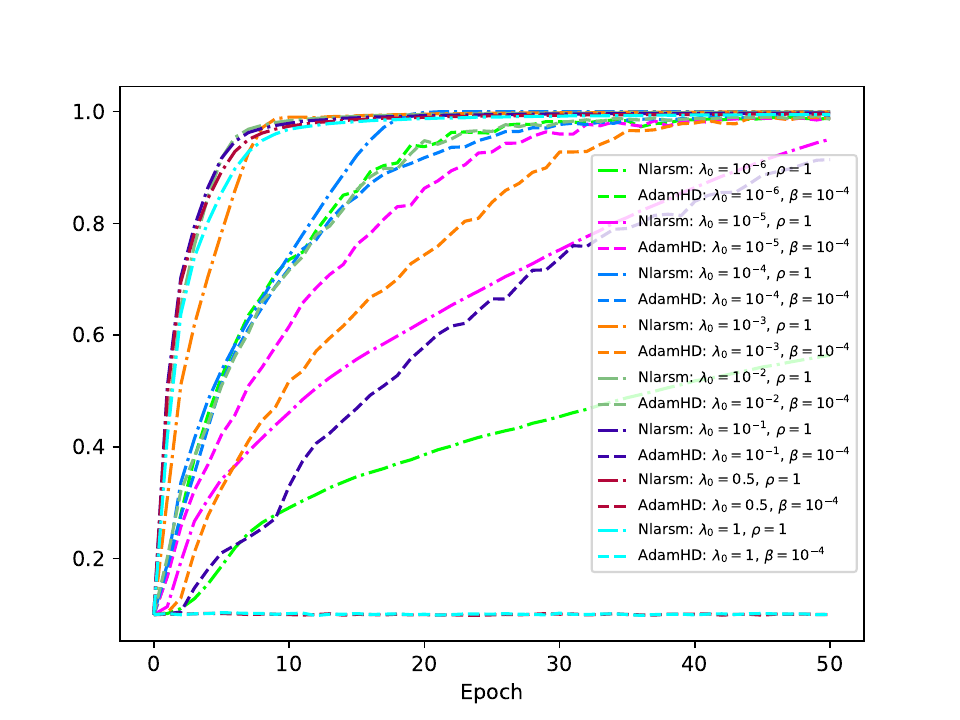}\hspace{-0.5mm}%
  \includegraphics[width=0.22\linewidth, trim=0 10 30 0]{imgs/CIFAR10_vgg11_nlarc_adam_mu=1_beta=None_minlr=None_50_0_1_300accuracy.pdf}\hspace{-0.5mm}%
  \includegraphics[width=0.22\linewidth, trim=0 10 30 0]{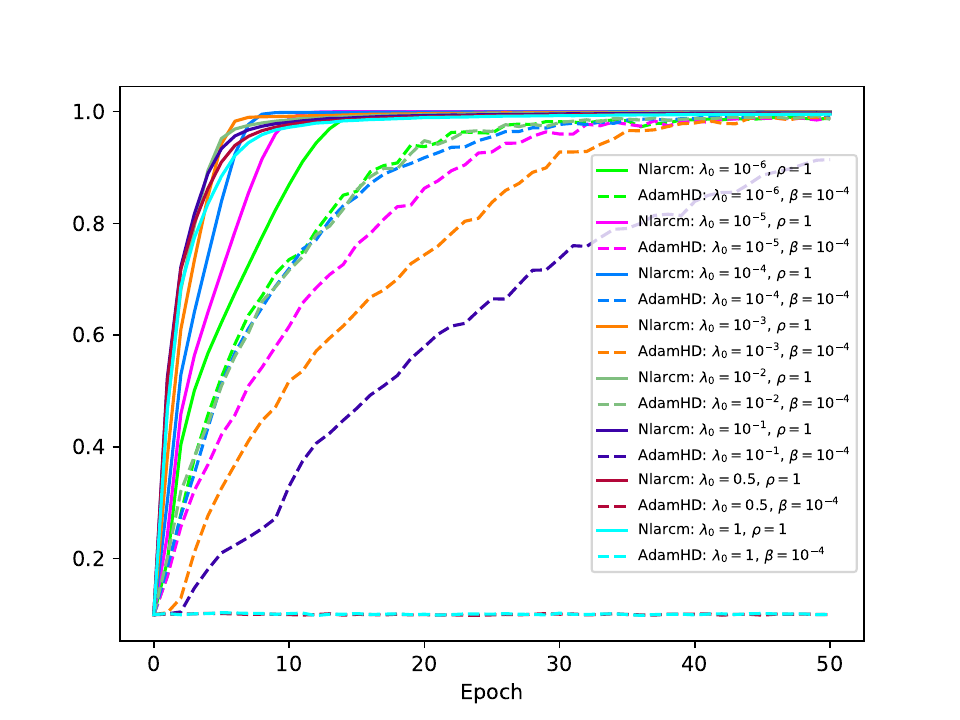}

    \vspace{1\baselineskip}
  \caption{VGG11 on the CIFAR10 dataset: Performance comparison of Nlarsm and Nlarcm versus Adam and AdamHD (with $\beta = 10^{-4}$) across varied learning rates. }
  \vspace{1\baselineskip}
  \label{fig:exp-vgg11-CIFAR10-nlar-adam-adamhd-1e-4}
\end{figure*}

Figures \ref{fig:exp-vgg11-CIFAR10-nlar-adam-adamhd-1e-8} and  \ref{fig:exp-vgg11-CIFAR10-nlar-adam-adamhd-1e-4} show the comparison of Nlarsm and Nlarcm with Adam and AdamHD (with $\beta$ in $\{10^{-8}, 10^{-4}\}$). Similar to the previous experiments, Nlarsm and Nlarcm performs quite well compared to Adam and AdamHD across large initial learning rates, especially for the first 10 epochs. The robustness feature carries on here as well; both Adam and AdamHD do not show any convergence for $\lambda_0 \geq 0.1$.

\subsection{The DDQN model on the classical control problems of CartPole-v0}   

\begin{figure*}[!ht]
  \centering
  \includegraphics[width=0.22\linewidth, trim=0 10 30 0]{imgs/dqn_CartPole_nlars_adam_mu=1_beta=None_minlr=None_CartPole-v0_cumulative_reward_testing.pdf}\hspace{-0.5mm}%
    \includegraphics[width=0.22\linewidth, trim=0 10 30 0]{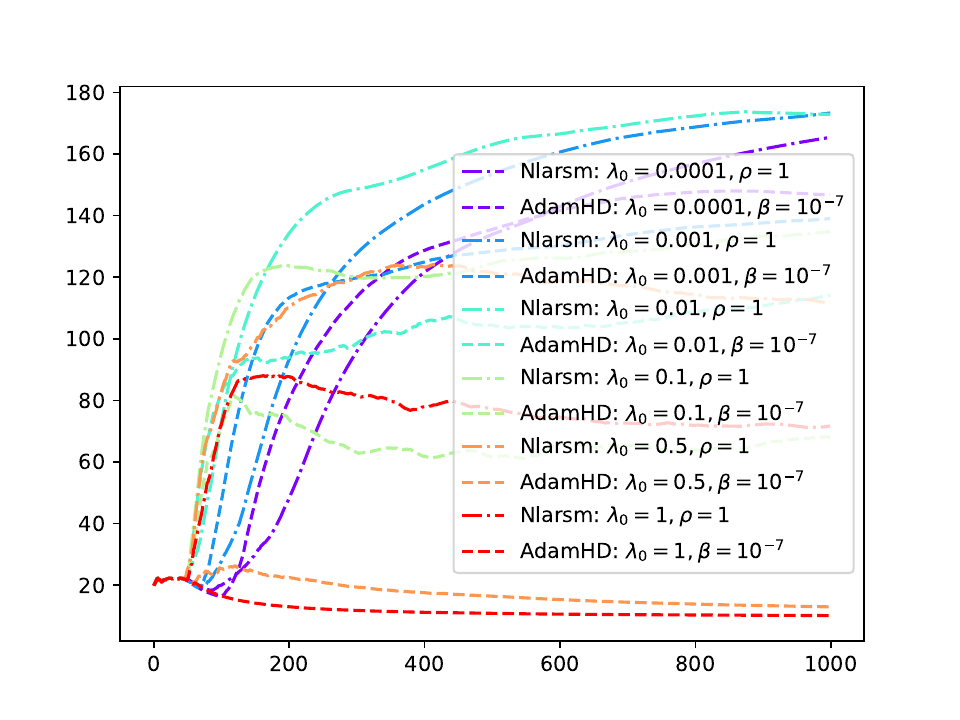}\hspace{-0.5mm}%
  \includegraphics[width=0.22\linewidth, trim=0 10 30 0]{imgs/dqn_CartPole_nlarc_adam_mu=1_beta=None_minlr=None_CartPole-v0_cumulative_reward_testing.pdf}\hspace{-0.5mm}%
  \includegraphics[width=0.22\linewidth, trim=0 10 30 0]{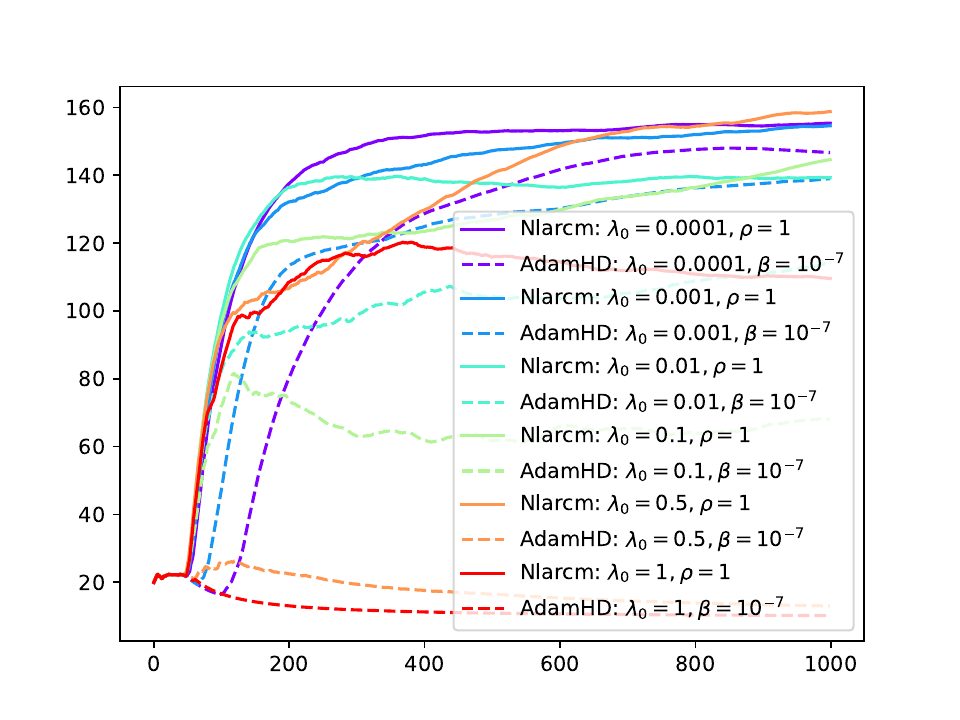}\hspace{-0.5mm}%

  \includegraphics[width=0.22\linewidth, trim=0 10 30 0]{imgs/dqn_CartPole_nlars_adam_mu=1_beta=None_minlr=None_CartPole-v0_cumulative_reward_training.pdf}\hspace{-0.5mm}%
    \includegraphics[width=0.22\linewidth, trim=0 10 30 0]{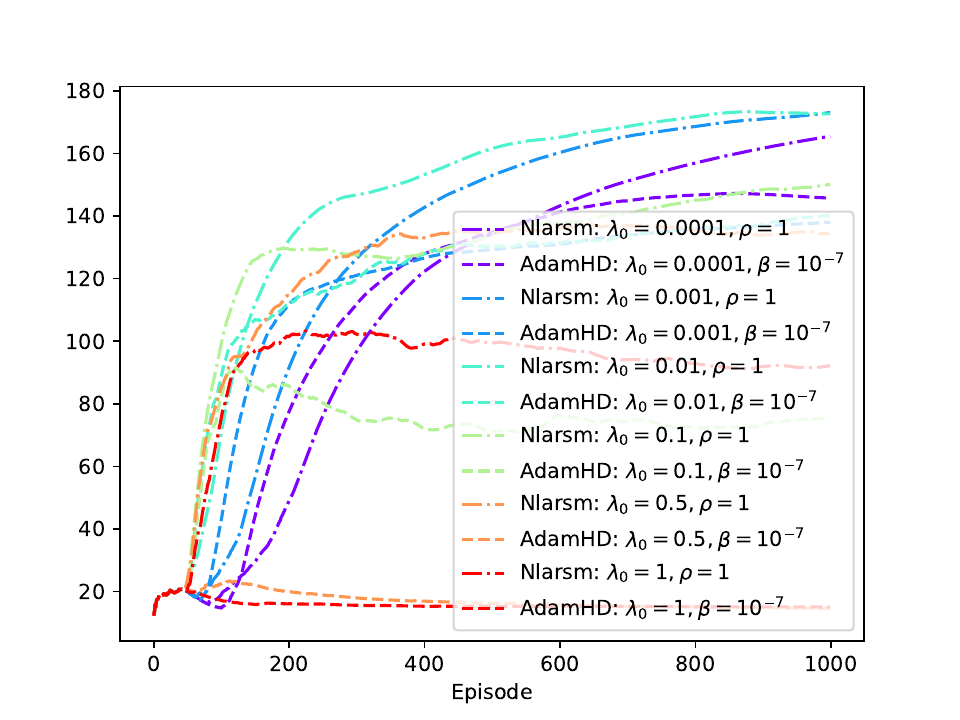}\hspace{-0.5mm}%
  \includegraphics[width=0.22\linewidth, trim=0 10 30 0]{imgs/dqn_CartPole_nlarc_adam_mu=1_beta=None_minlr=None_CartPole-v0_cumulative_reward_training.pdf}\hspace{-0.5mm}%
  \includegraphics[width=0.22\linewidth, trim=0 10 30 0]{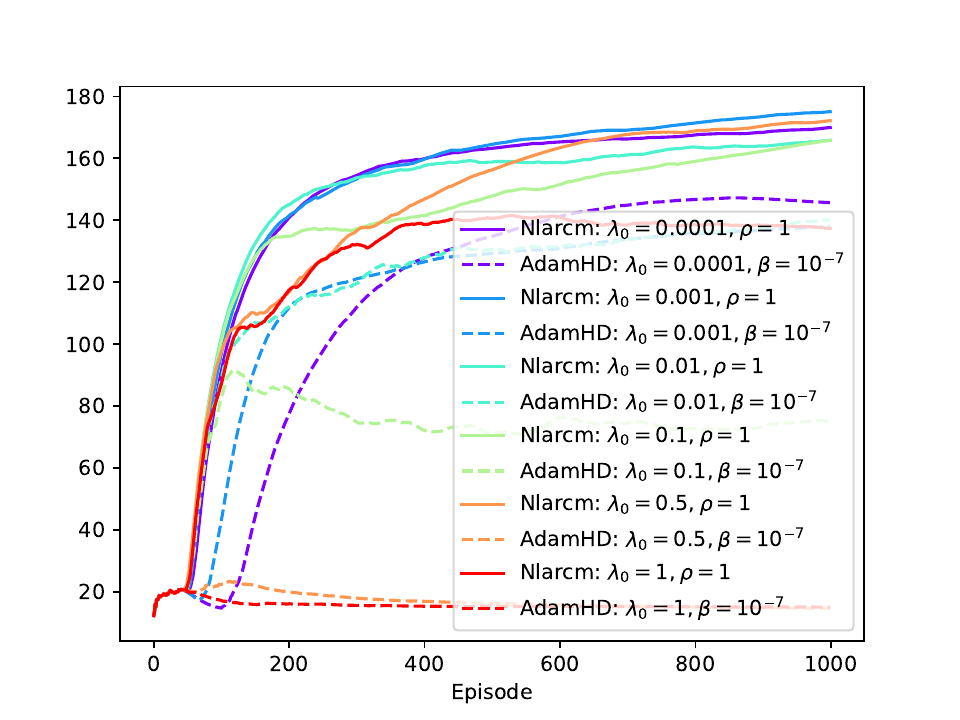}\hspace{-0.5mm}%
    \vspace{1\baselineskip}
  \caption{CartPole-v0: Performance comparison of Nlarsm and Nlarcm versus Adam and AdamHD (with $\beta = 10^{-7}$) across varied learning rates.}
  \vspace{1\baselineskip}
  \label{fig:exp-CartPole-nlar-adam-adamhd-beta=1e-07}
\end{figure*}

Figure \ref{fig:exp-CartPole-nlar-adam-adamhd-beta=1e-07} compares the performance of Nlarsm and Nlarcm versus Adam and AdamHD for $\beta=10^{-7}$ on CartPole-v0 environment.

\subsection{Nlars and Nlarc}

\begin{figure*}[!ht]
  \centering
  \includegraphics[width=0.22\linewidth, trim=0 10 30 0]{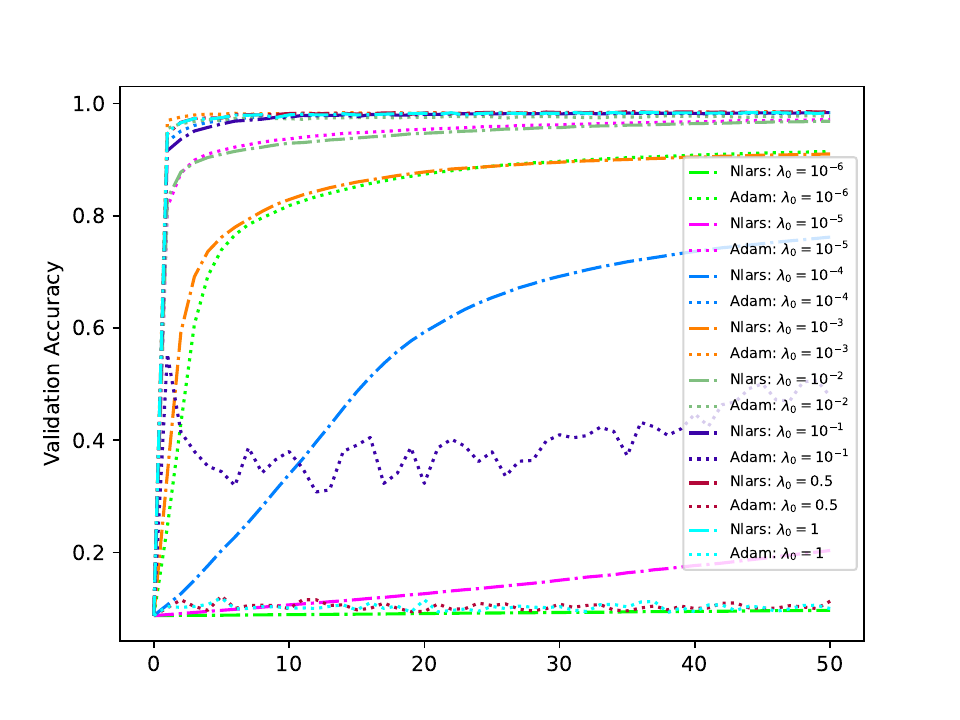}\hspace{-0.5mm}%
  \includegraphics[width=0.22\linewidth, trim=0 10 30 0]{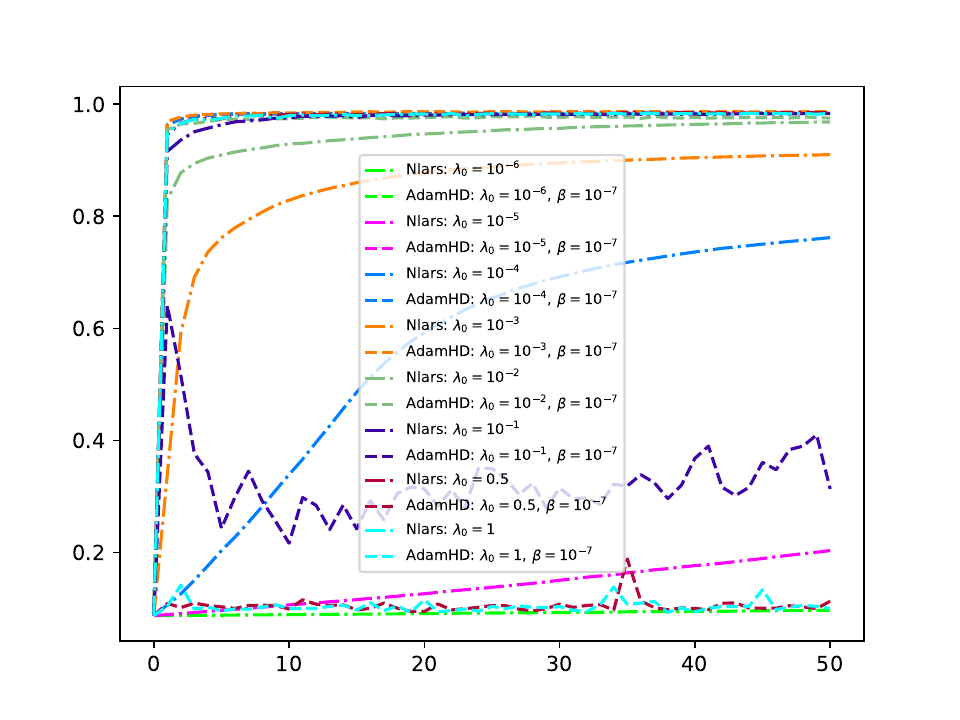}\hspace{-0.5mm}%
  \includegraphics[width=0.22\linewidth, trim=0 10 30 0]{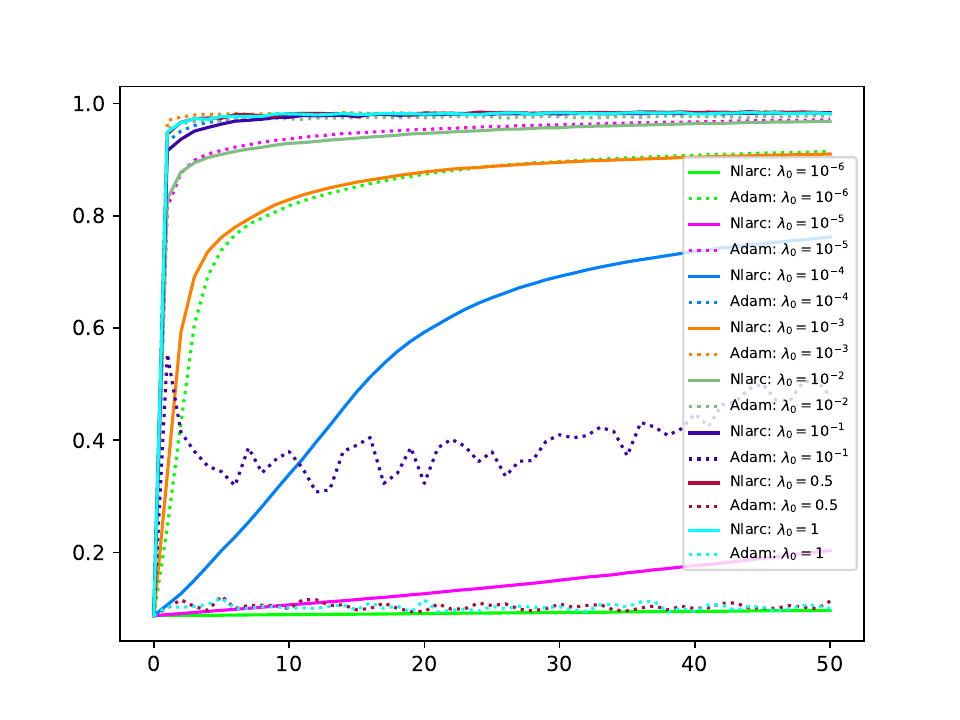}\hspace{-0.5mm}%
  \includegraphics[width=0.22\linewidth, trim=0 10 30 0]{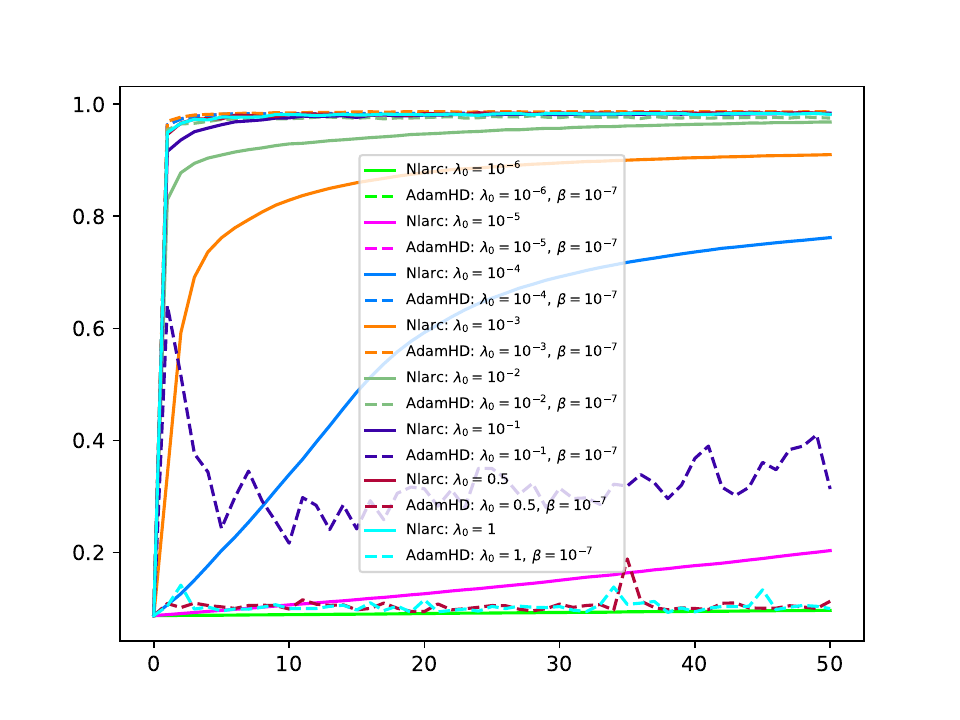}

  \includegraphics[width=0.22\linewidth, trim=0 10 30 0]{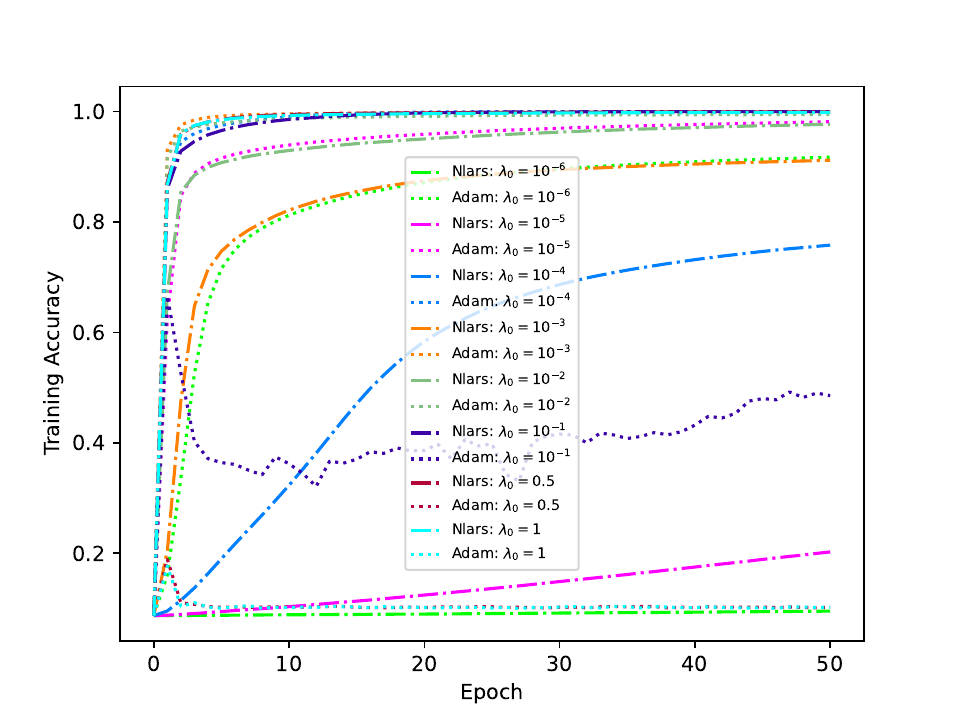}\hspace{-0.5mm}%
  \includegraphics[width=0.22\linewidth, trim=0 10 30 0]{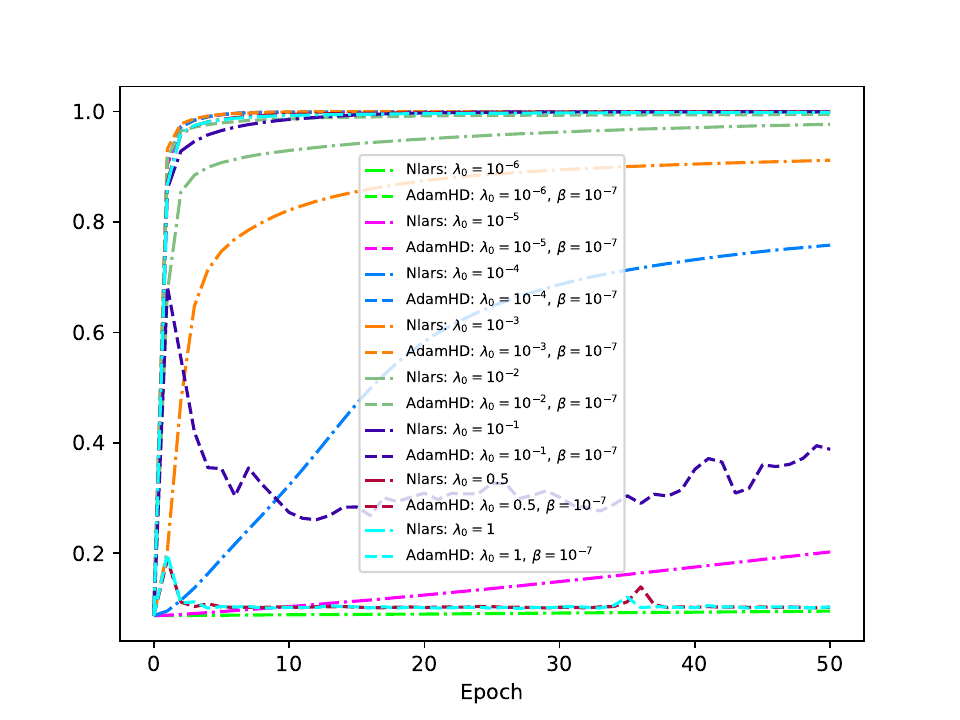}\hspace{-0.5mm}%
  \includegraphics[width=0.22\linewidth, trim=0 10 30 0]{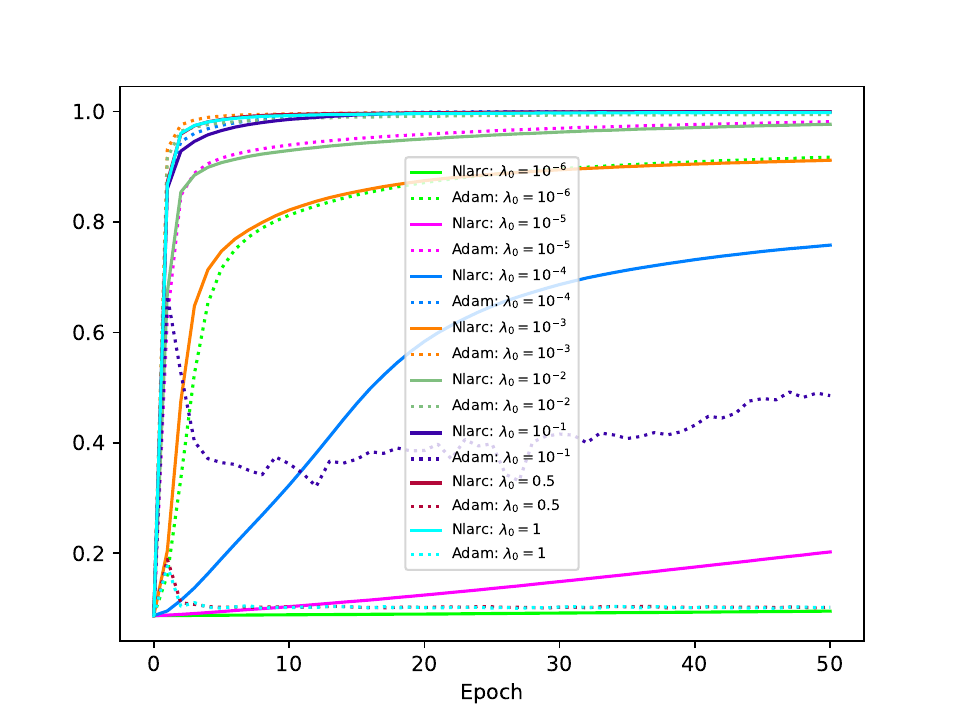}\hspace{-0.5mm}%
  \includegraphics[width=0.22\linewidth, trim=0 10 30 0]{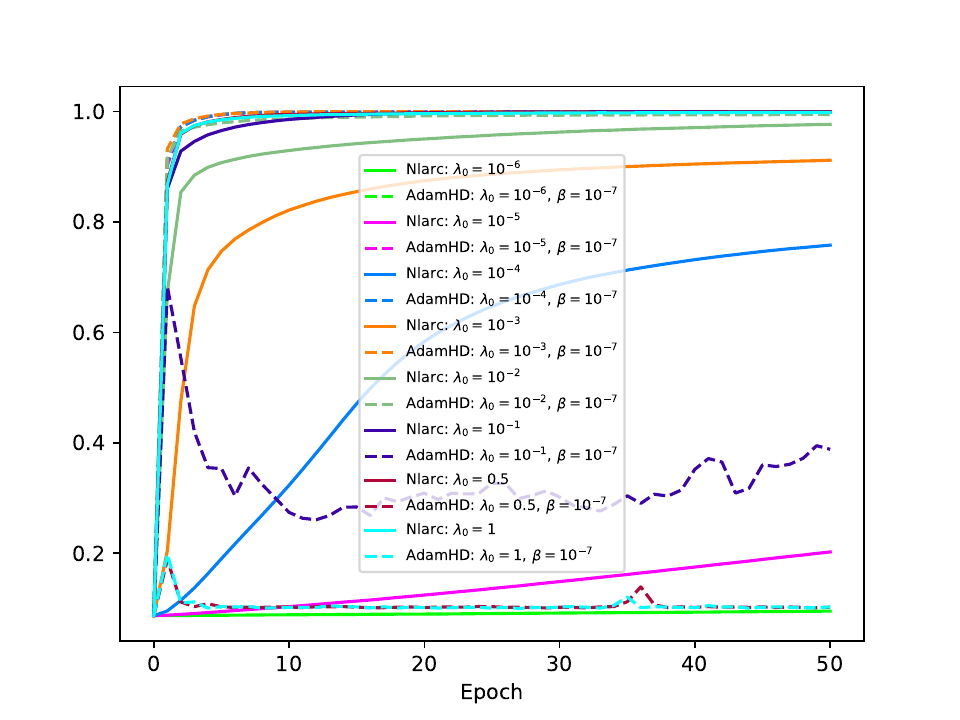}

    \vspace{1\baselineskip}
  \caption{MLP2h on the MNIST dataset: Performance comparison of Nlars and Nlarc versus Adam and AdamHD (with $\beta = 10^{-7}$) across varied learning rates.}
  \vspace{1\baselineskip}
  \label{fig:exp-mlp-mnist-nlar-mu=none-adam-adamhd-1e-7}
\end{figure*}

\begin{figure*}[!ht]
  \centering
  \includegraphics[width=0.22\linewidth, trim=0 10 30 0]{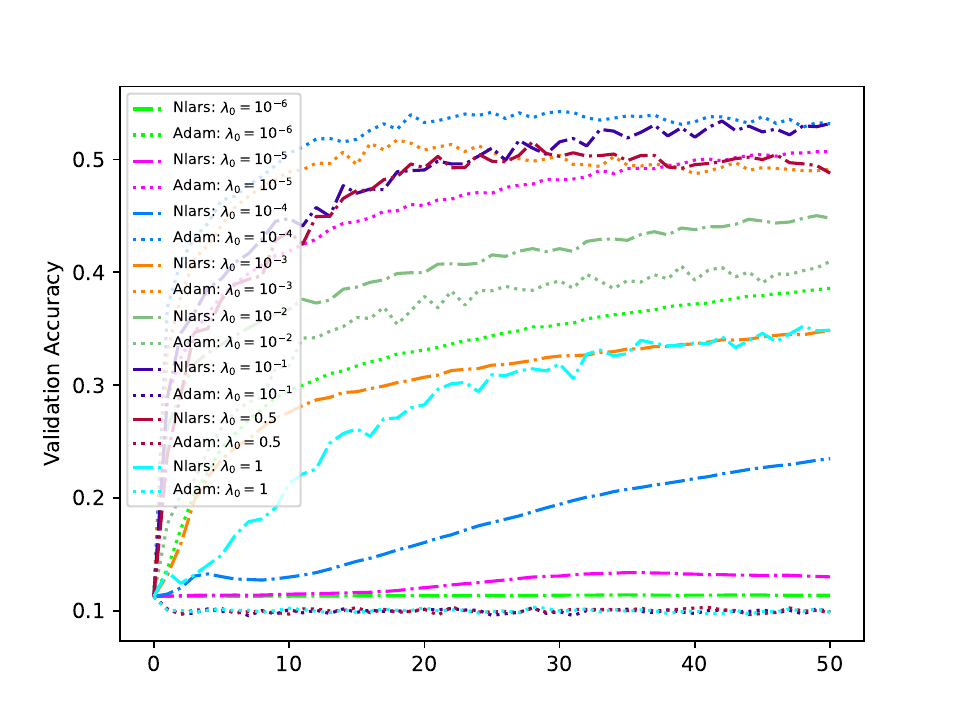}\hspace{-0.5mm}%
  \includegraphics[width=0.22\linewidth, trim=0 10 30 0]{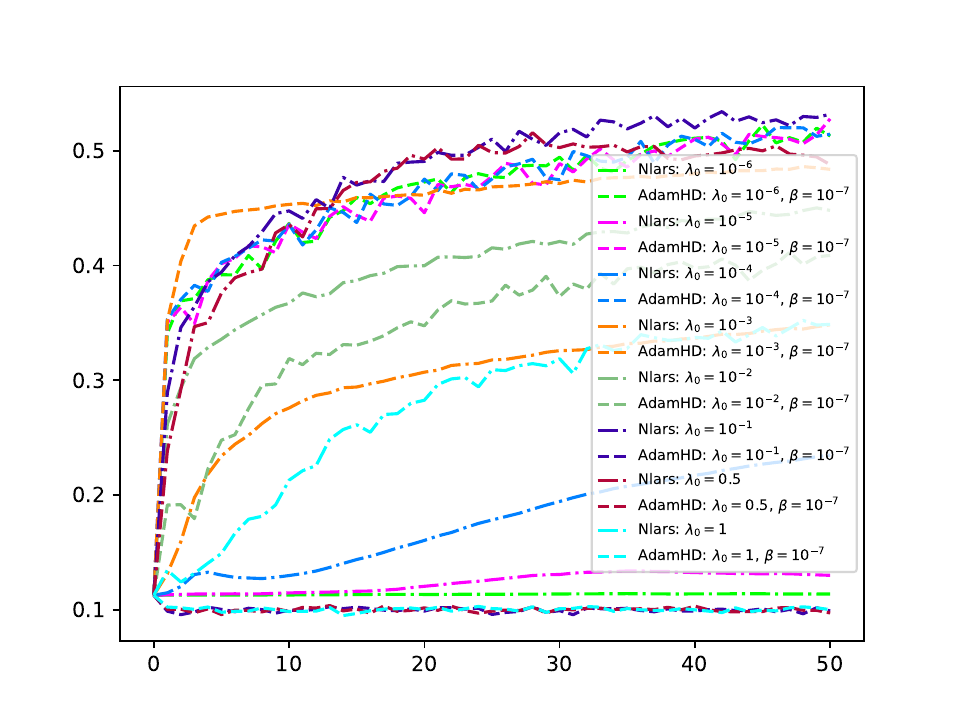}\hspace{-0.5mm}%
  \includegraphics[width=0.22\linewidth, trim=0 10 30 0]{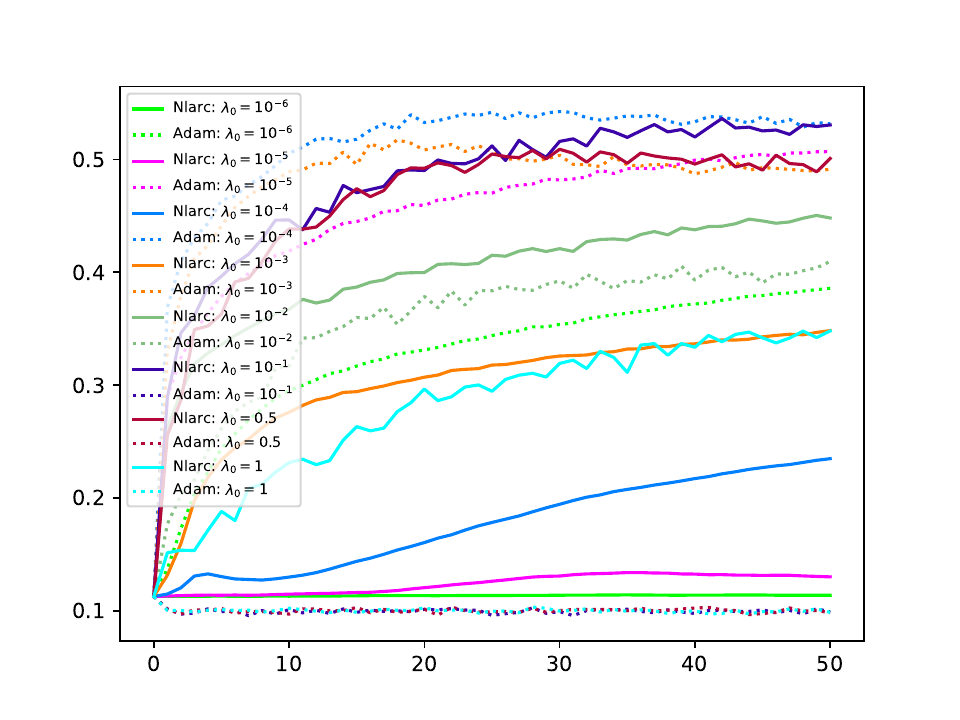}\hspace{-0.5mm}%
  \includegraphics[width=0.22\linewidth, trim=0 10 30 0]{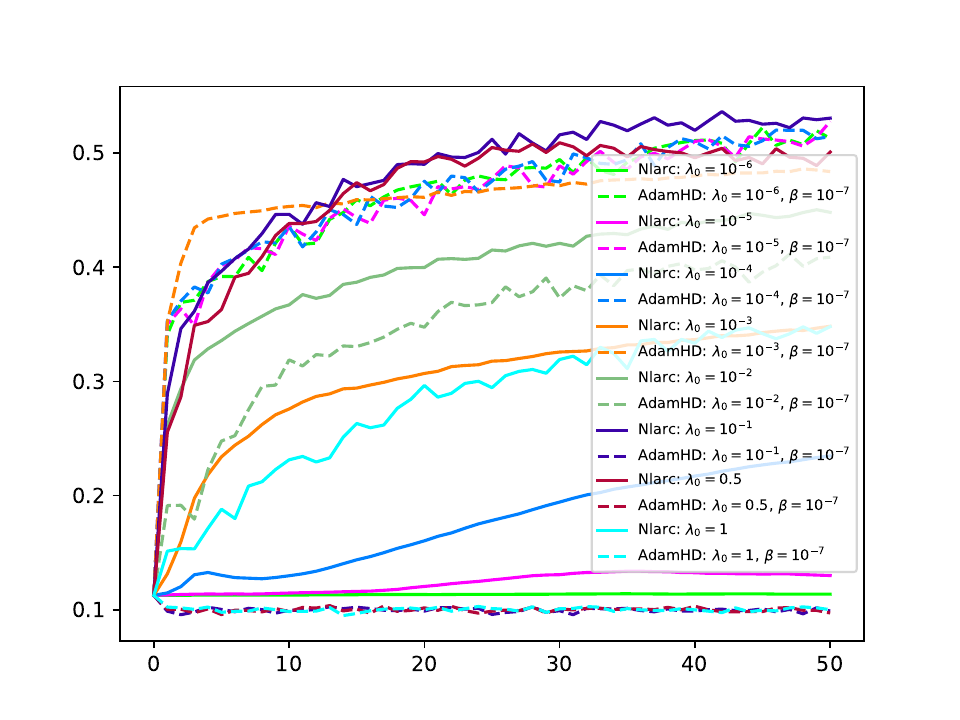}

  \includegraphics[width=0.22\linewidth, trim=0 10 30 0]{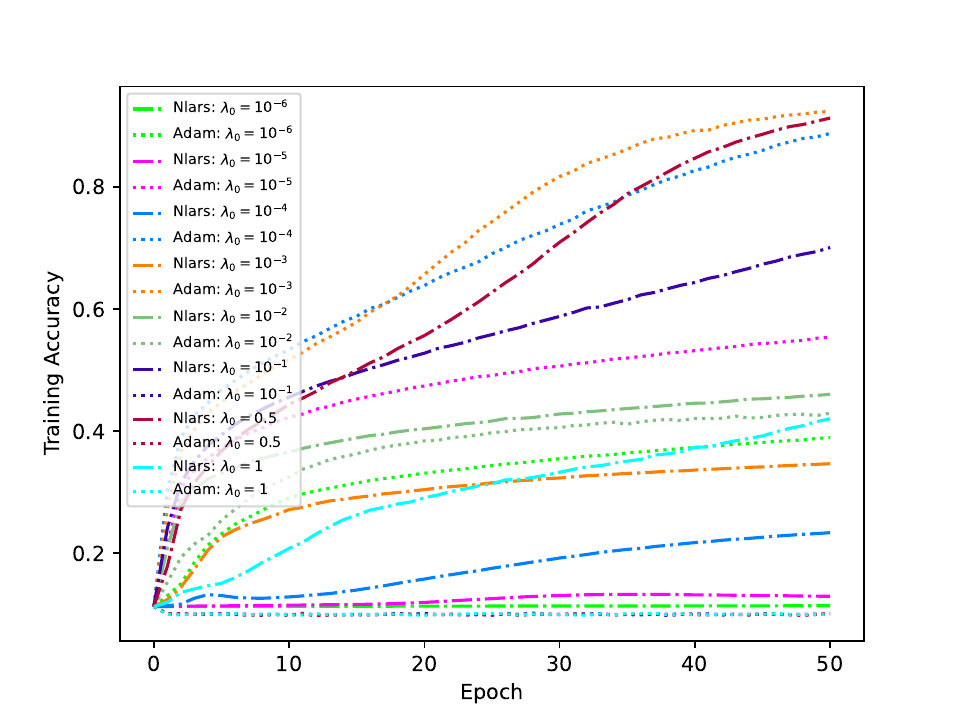}\hspace{-0.5mm}%
  \includegraphics[width=0.22\linewidth, trim=0 10 30 0]{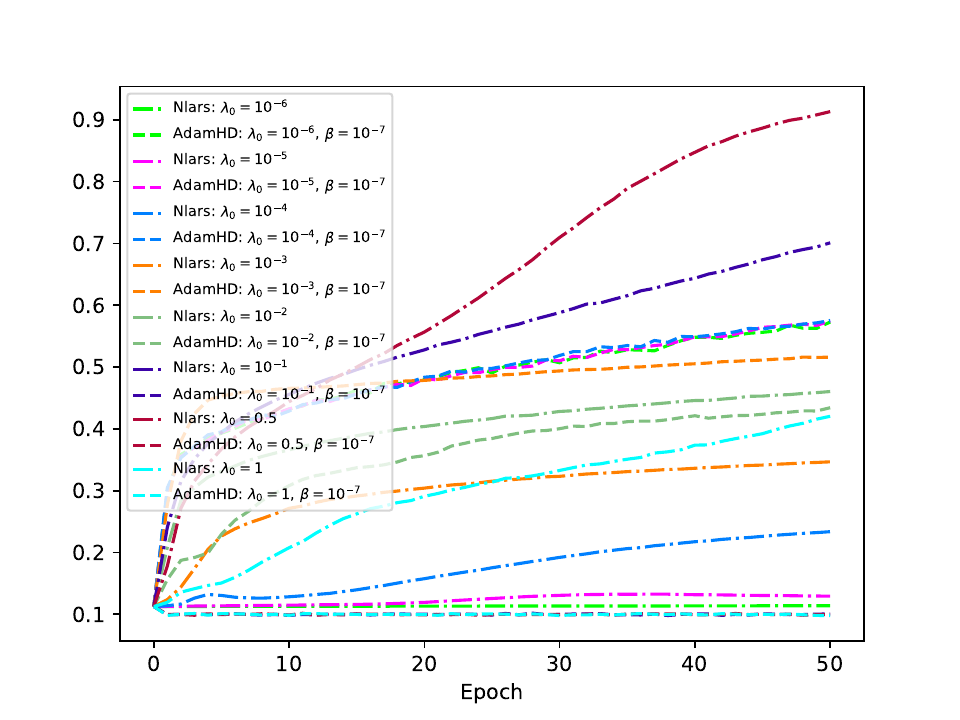}\hspace{-0.5mm}%
  \includegraphics[width=0.22\linewidth, trim=0 10 30 0]{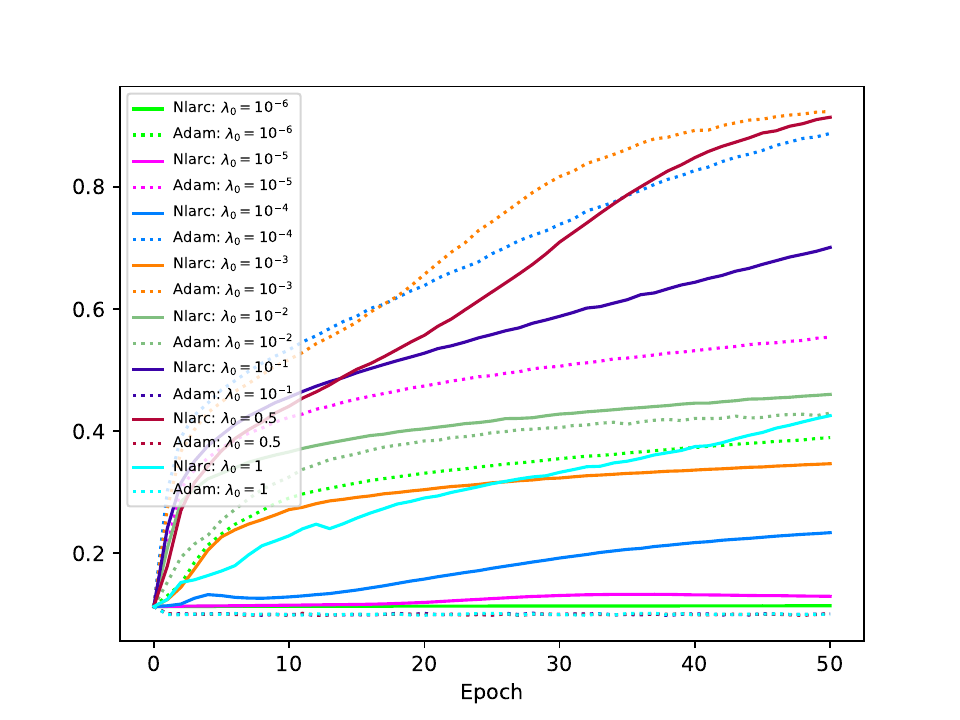}\hspace{-0.5mm}%
  \includegraphics[width=0.22\linewidth, trim=0 10 30 0]{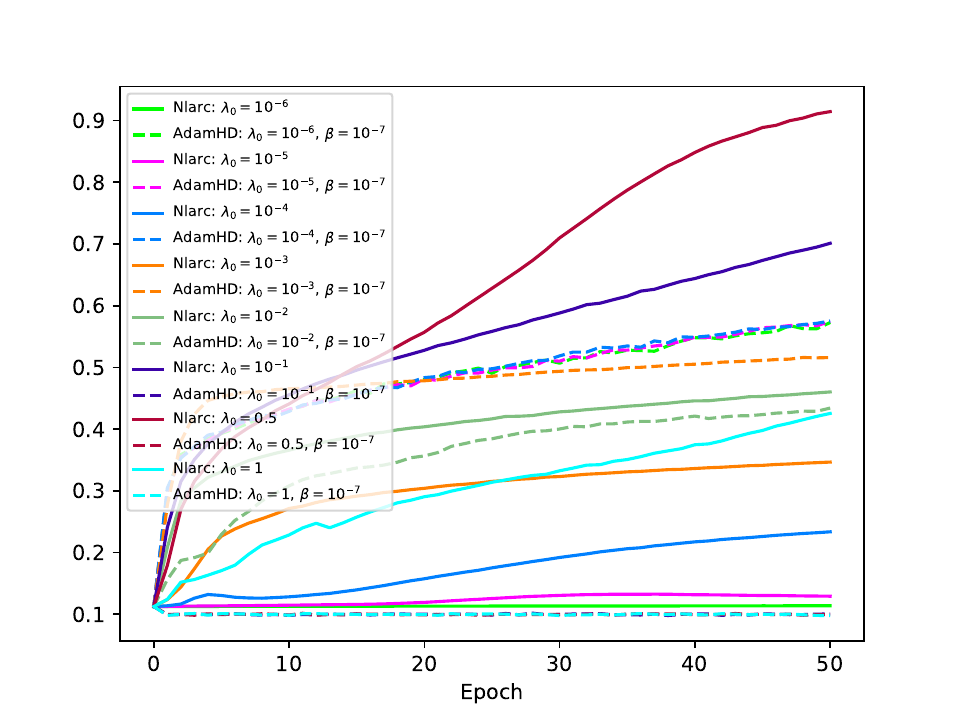}

    \vspace{1\baselineskip}
  \caption{MLP7h on the CIFAR10 dataset: Performance comparison of Nlars and Nlarc versus Adam and AdamHD (with $\beta = 10^{-7}$) across varied learning rates.}
  \vspace{1\baselineskip}
  \label{fig:exp-mlp7h-CIFAR10-nlar-mu=none-adam-adamhd-1e-7}
\end{figure*}

Recall that Nlars and Nlarc optimizers are indeed Nlarsm and Nlarcm when there is no momentum, i.e. $\rho_t(d)=0$ which is the case if $\rho=0$. Figures \ref{fig:exp-mlp-mnist-nlar-mu=none-adam-adamhd-1e-7} and \ref{fig:exp-mlp7h-CIFAR10-nlar-mu=none-adam-adamhd-1e-7} show the performance of these these optimizers using MLP2h and MLP7h on the MNIST and CIFAR10 datasets. Nlars and Nlarc demonstrate stable convergence, and to achieve good performance, these two optimizers should be applied with large initial learning rates. Overall, Nlarsm and Nlarcm achieve significantly better performance across different initial learning rates.

\section{Sensitivity analysis}
\label{sec:app-sensitivity}
In this section, through some experiments, we provide insights into the sensitivity of the Nlar algorithms with respect to the underlying parameters.
\subsection*{Sensitivity of Nlarsm and Nlarcm with respect to $k$, $c^\prime$, $c$, and $B^\prime$.}

\begin{figure*}[!ht]
  \centering
  \includegraphics[width=0.22\linewidth, trim=0 10 30 0]{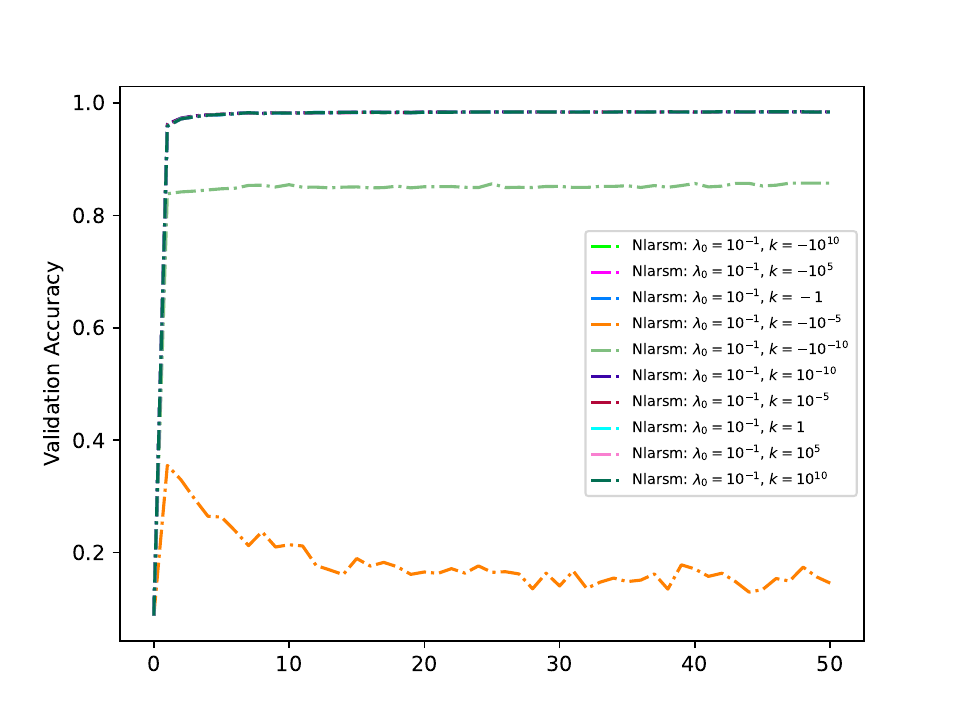}\hspace{-0.5mm}%
    \includegraphics[width=0.22\linewidth, trim=0 10 30 0]{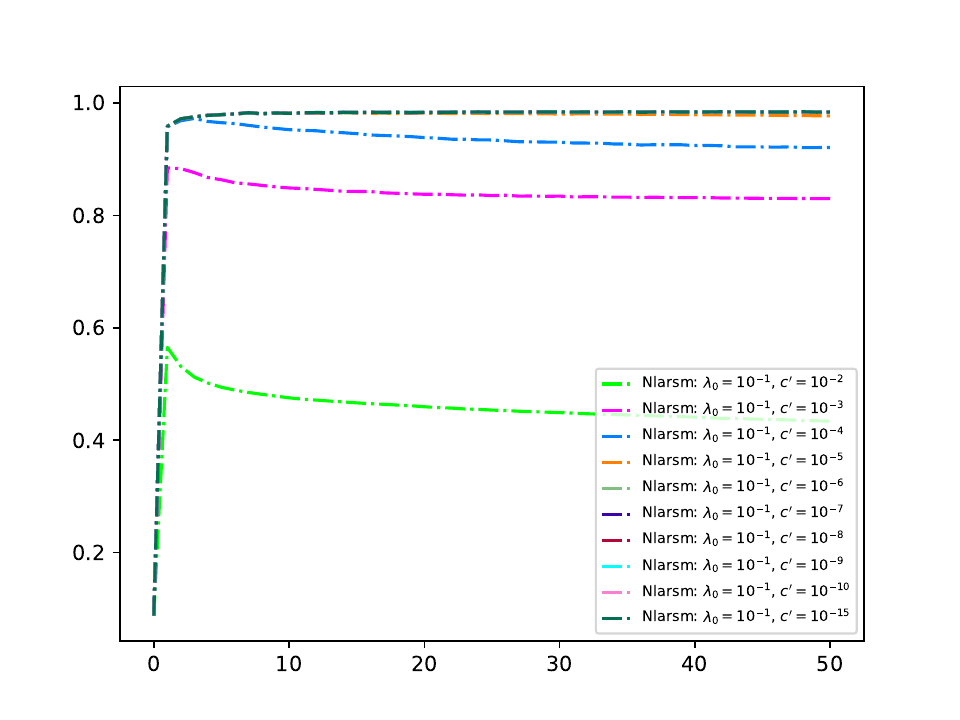}\hspace{-0.5mm}%
  \includegraphics[width=0.22\linewidth, trim=0 10 30 0]{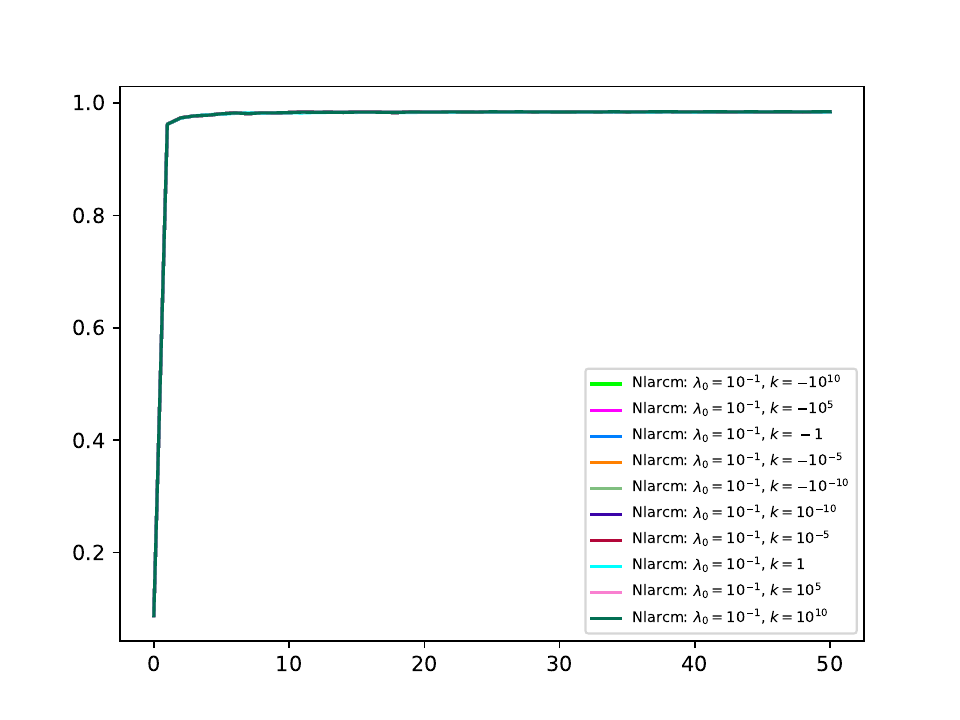}\hspace{-0.5mm}%
  \includegraphics[width=0.22\linewidth, trim=0 10 30 0]{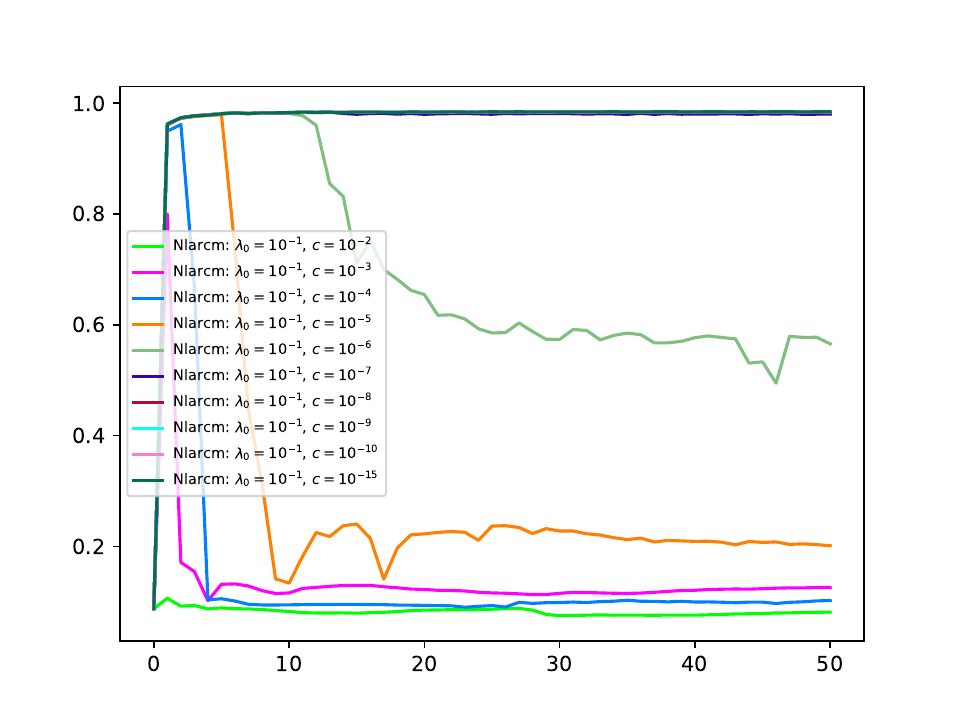}\hspace{-0.5mm}%

  \includegraphics[width=0.22\linewidth, trim=0 10 30 0]{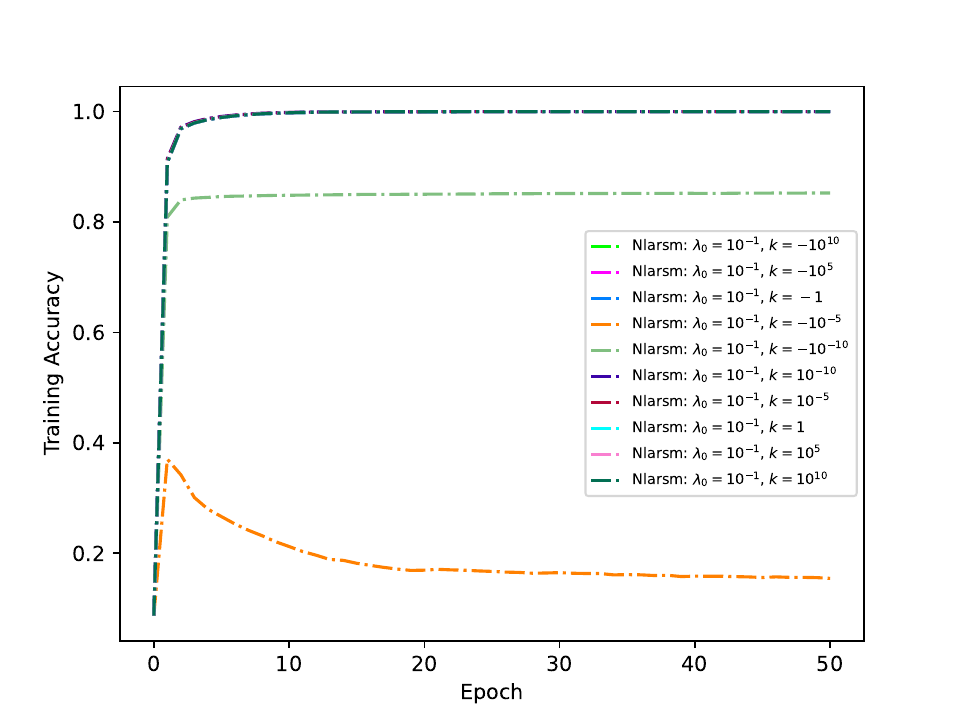}\hspace{-0.5mm}%
    \includegraphics[width=0.22\linewidth, trim=0 10 30 0]{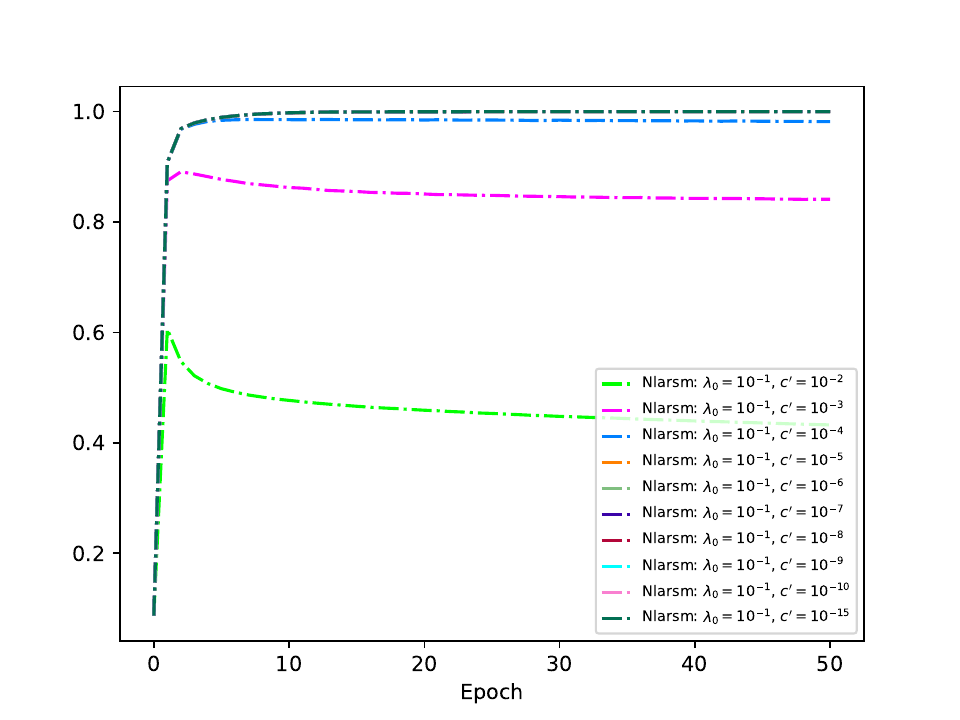}\hspace{-0.5mm}%
  \includegraphics[width=0.22\linewidth, trim=0 10 30 0]{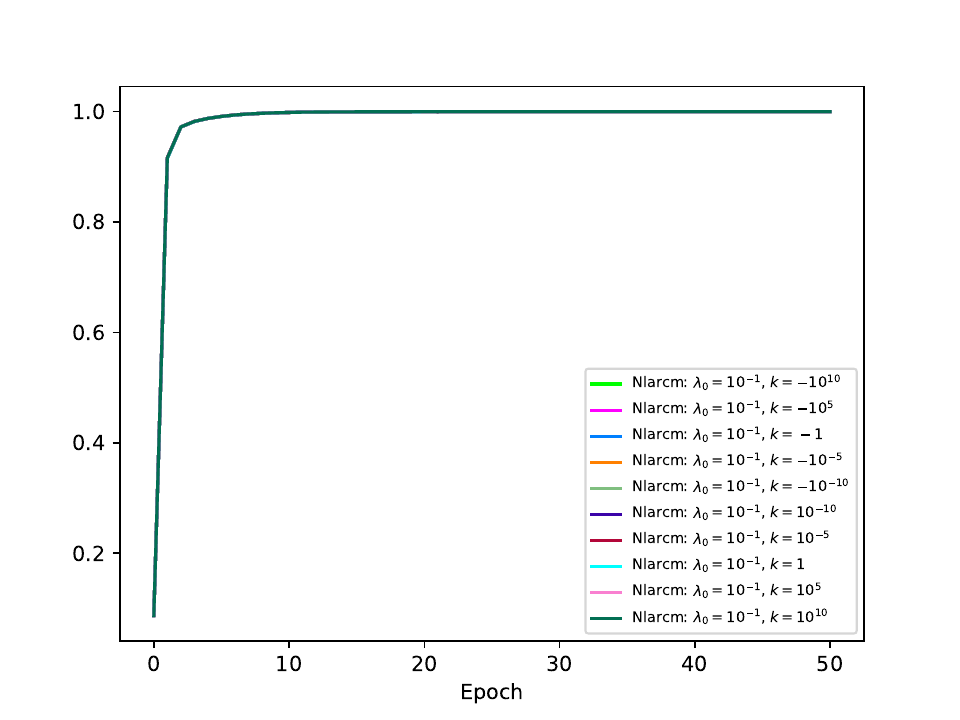}\hspace{-0.5mm}%
  \includegraphics[width=0.22\linewidth, trim=0 10 30 0]{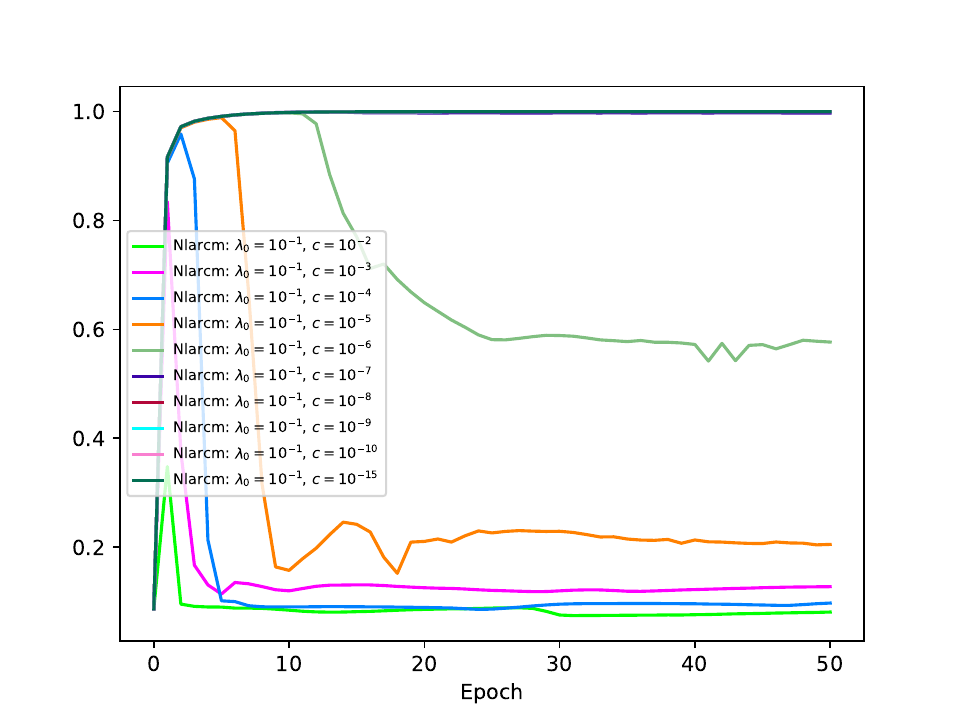}\hspace{-0.5mm}%
    \vspace{1\baselineskip}
  \caption{Sensitivity analysis of Nlarsm and Nlarcm, for a fixed $\lambda_0=0.1$, with respect to $k$, $c^\prime$, and $c$ parameters for MLP2h on the MNIST dataset.}
  \vspace{1\baselineskip}
  \label{fig:exp-mnist-mlp2h-nlar-sensitivity}
\end{figure*}

\begin{figure*}[!ht]
  \centering
  \includegraphics[width=0.22\linewidth, trim=0 10 30 0]{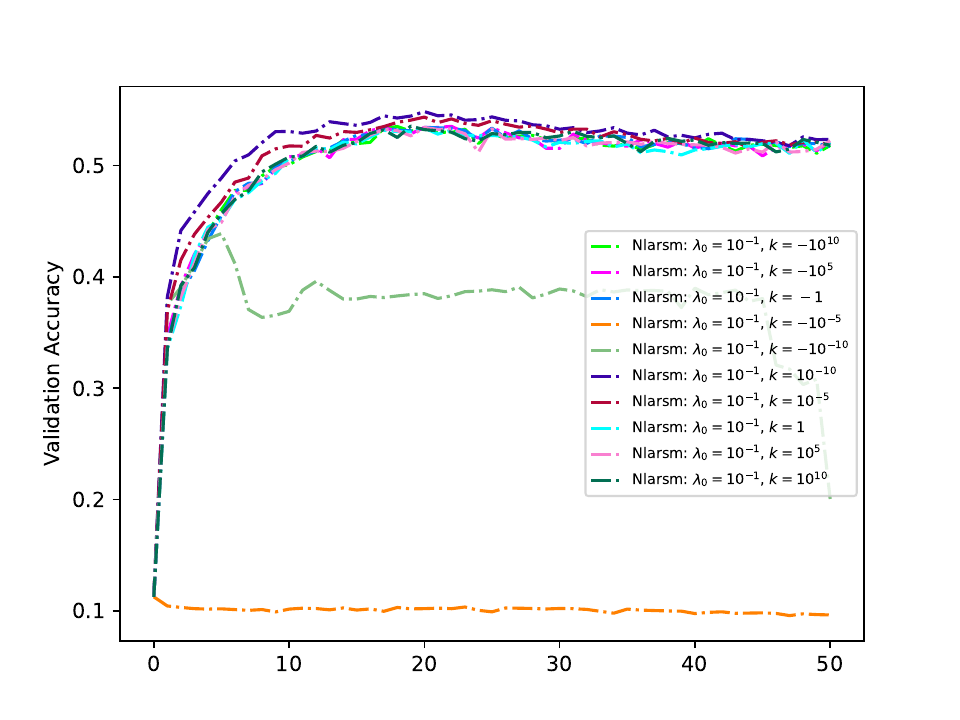}\hspace{-0.5mm}%
    \includegraphics[width=0.22\linewidth, trim=0 10 30 0]{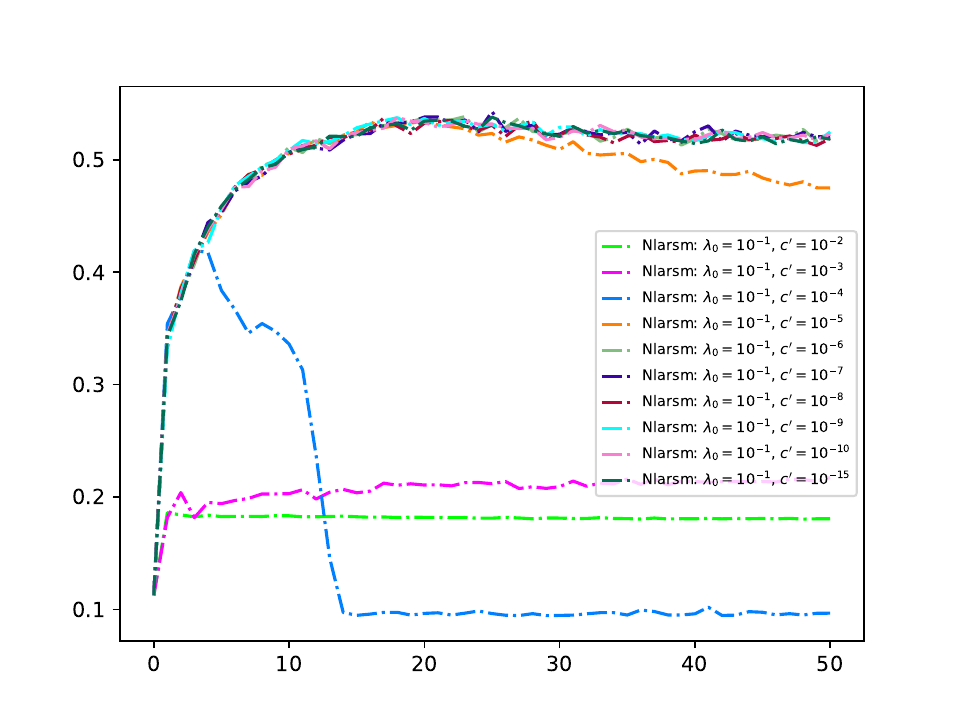}\hspace{-0.5mm}%
  \includegraphics[width=0.22\linewidth, trim=0 10 30 0]{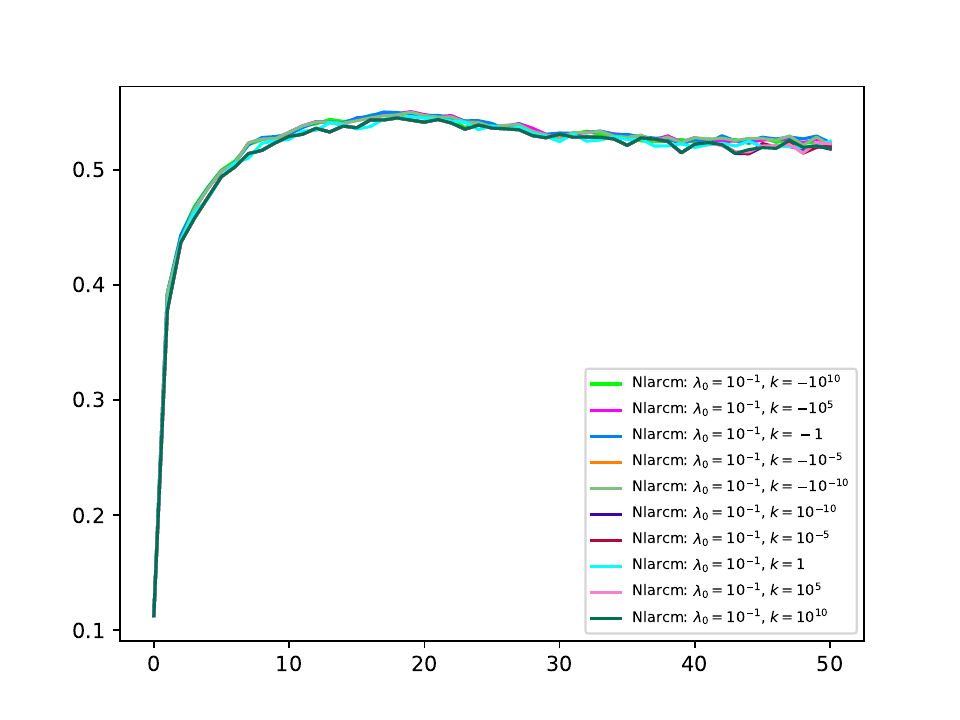}\hspace{-0.5mm}%
  \includegraphics[width=0.22\linewidth, trim=0 10 30 0]{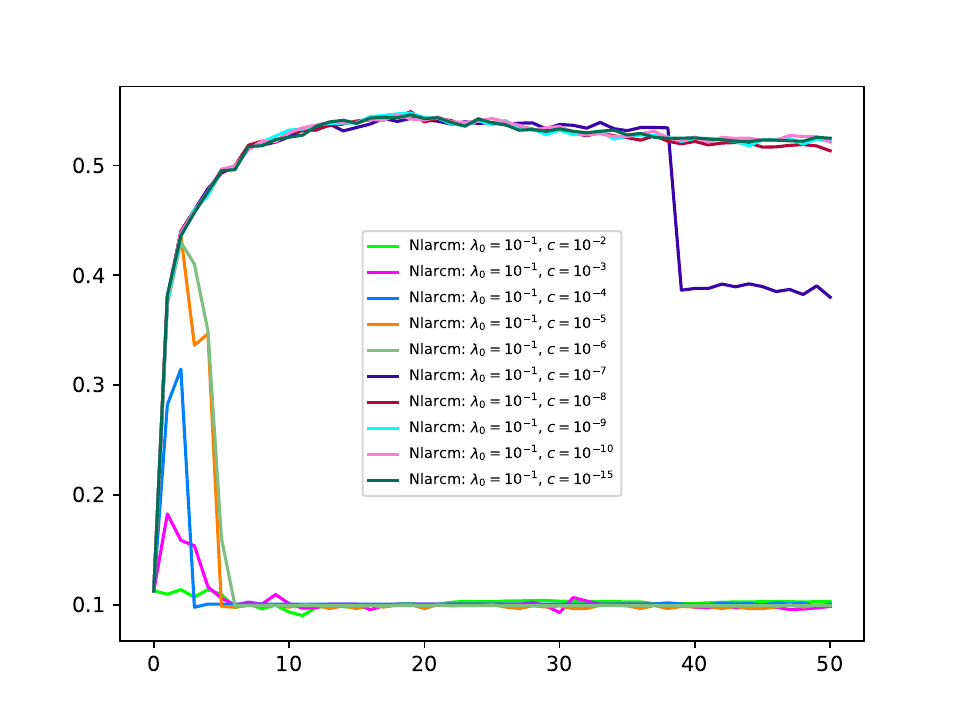}\hspace{-0.5mm}%

  \includegraphics[width=0.22\linewidth, trim=0 10 30 0]{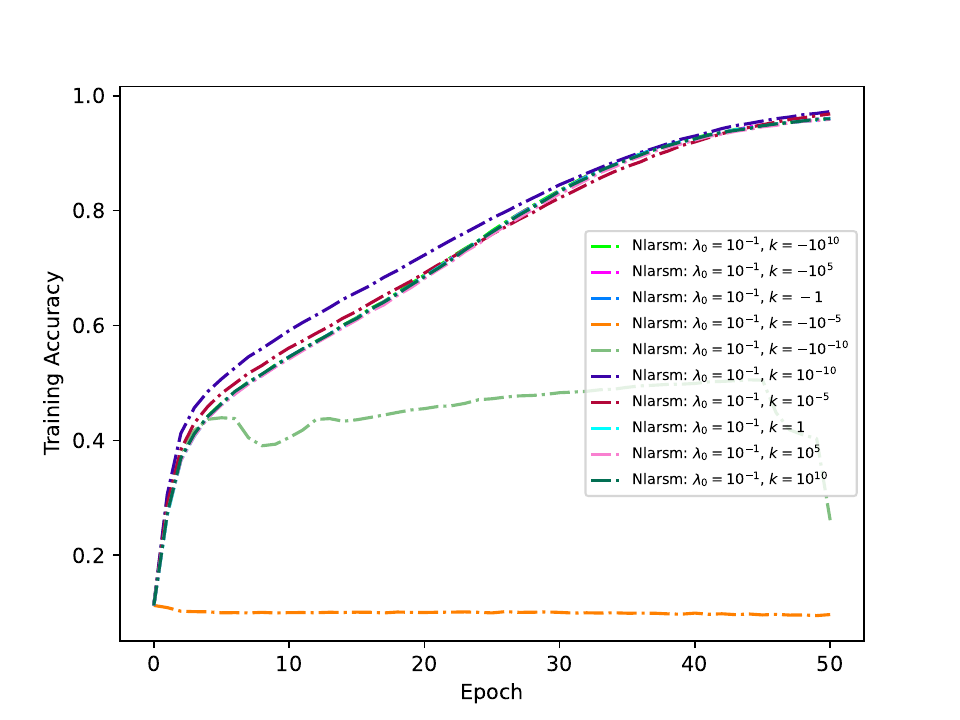}\hspace{-0.5mm}%
    \includegraphics[width=0.22\linewidth, trim=0 10 30 0]{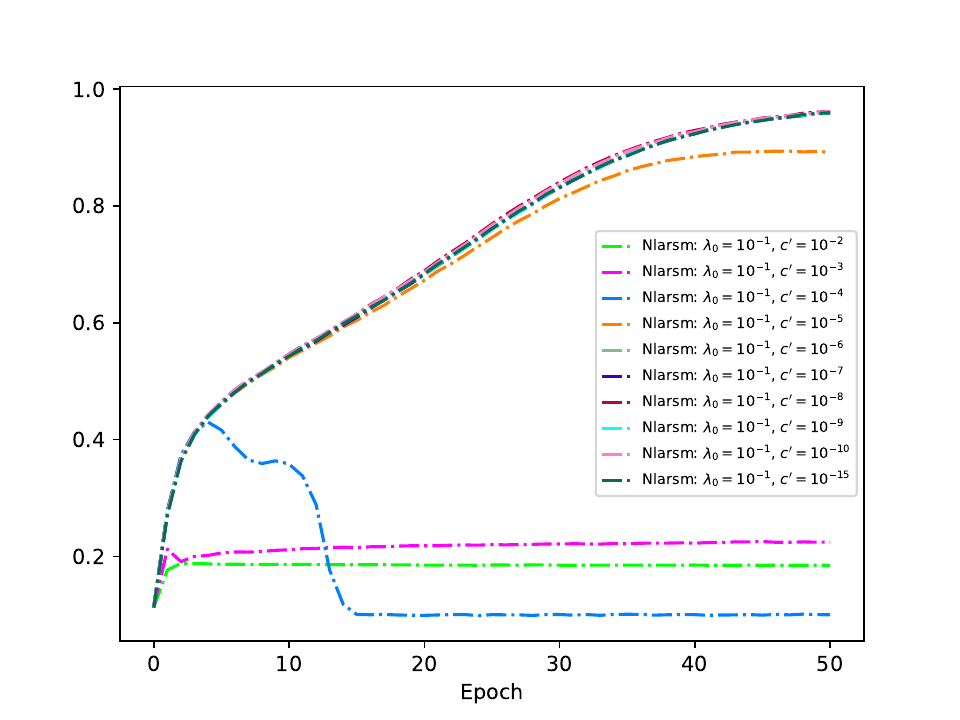}\hspace{-0.5mm}%
  \includegraphics[width=0.22\linewidth, trim=0 10 30 0]{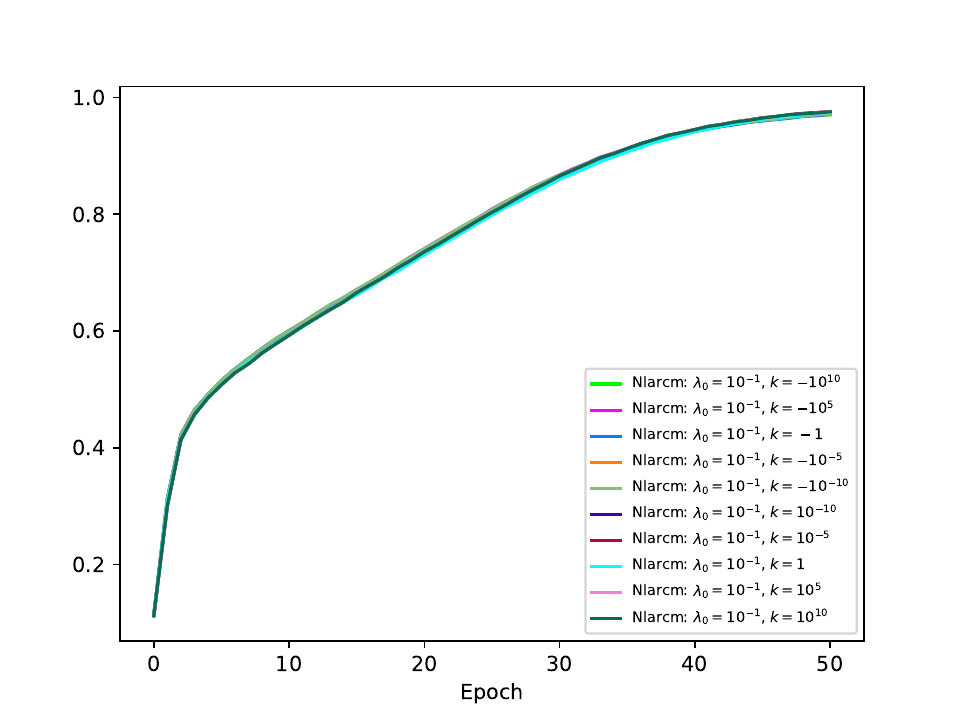}\hspace{-0.5mm}%
  \includegraphics[width=0.22\linewidth, trim=0 10 30 0]{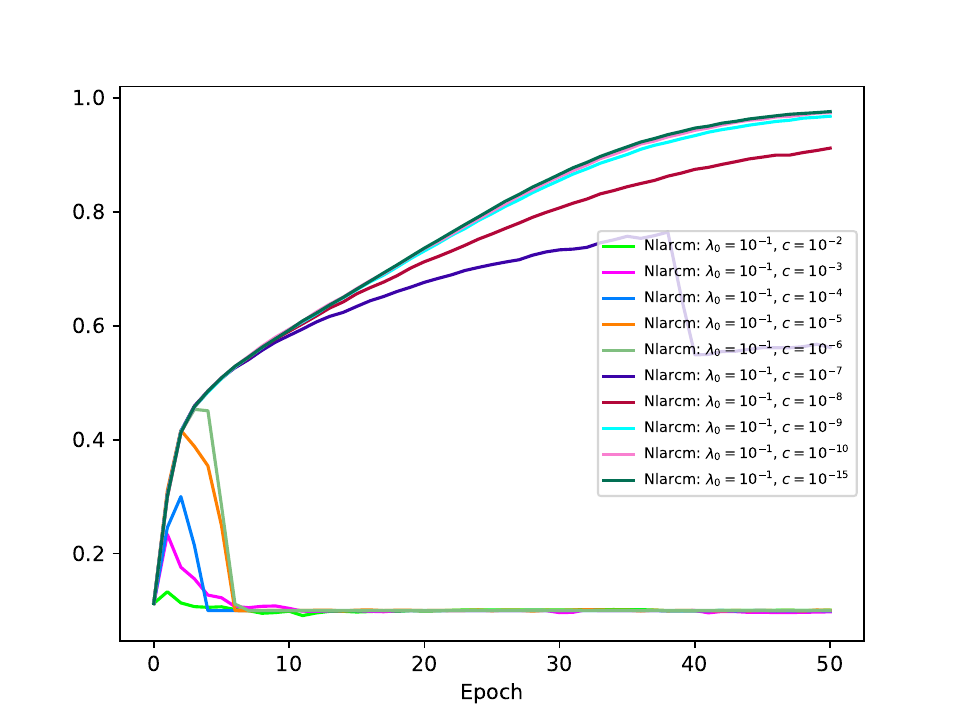}\hspace{-0.5mm}%
    \vspace{1\baselineskip}
  \caption{Sensitivity analysis of Nlarsm and Nlarcm, for a fixed $\lambda_0=0.1$, with respect to $k$, $c^\prime$, and $c$ parameters for MLP7h on the CIFAR10 dataset.}
  \vspace{1\baselineskip}
  \label{fig:exp-mnist-mlp7h-nlar-sensitivity}
\end{figure*}

\begin{figure}[htbp]
    \centering
    \begin{subfigure}[b]{0.22\textwidth}
        \centering
        \includegraphics[width=\textwidth]{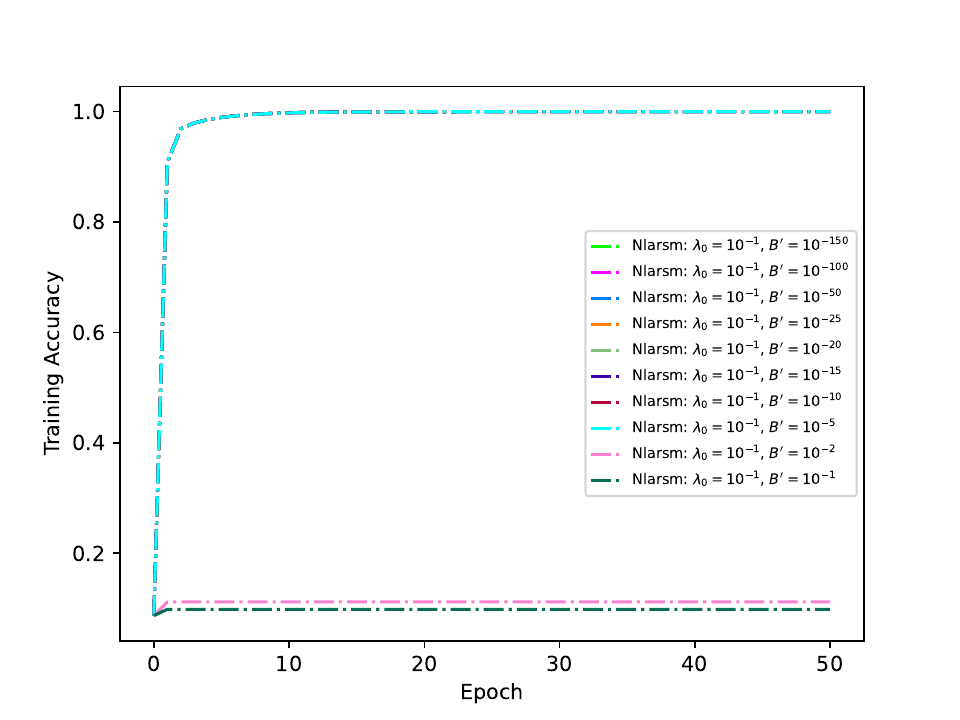}\hspace{-0.5mm}%
        \caption{\tiny MLP2h - MNIST}
    \end{subfigure}
    \hfill
    \begin{subfigure}[b]{0.22\textwidth}
        \centering
        \includegraphics[width=\textwidth]{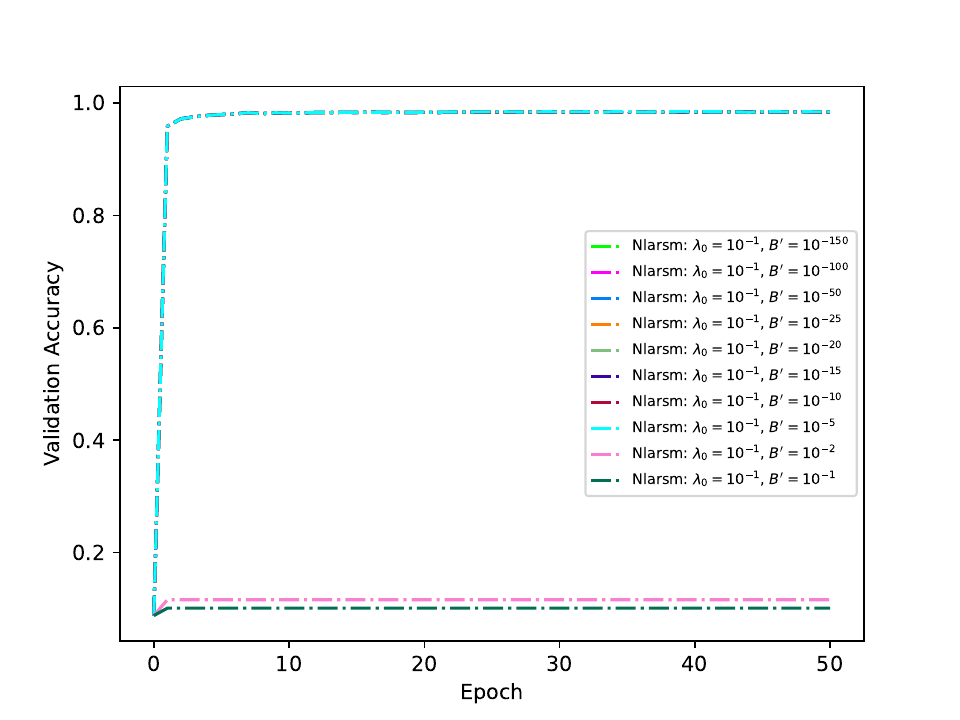}\hspace{-0.5mm}%
        \caption{\tiny MLP2h - MNIST}
    \end{subfigure}
    \hfill
    \begin{subfigure}[b]{0.22\textwidth}
        \centering
        \includegraphics[width=\textwidth]{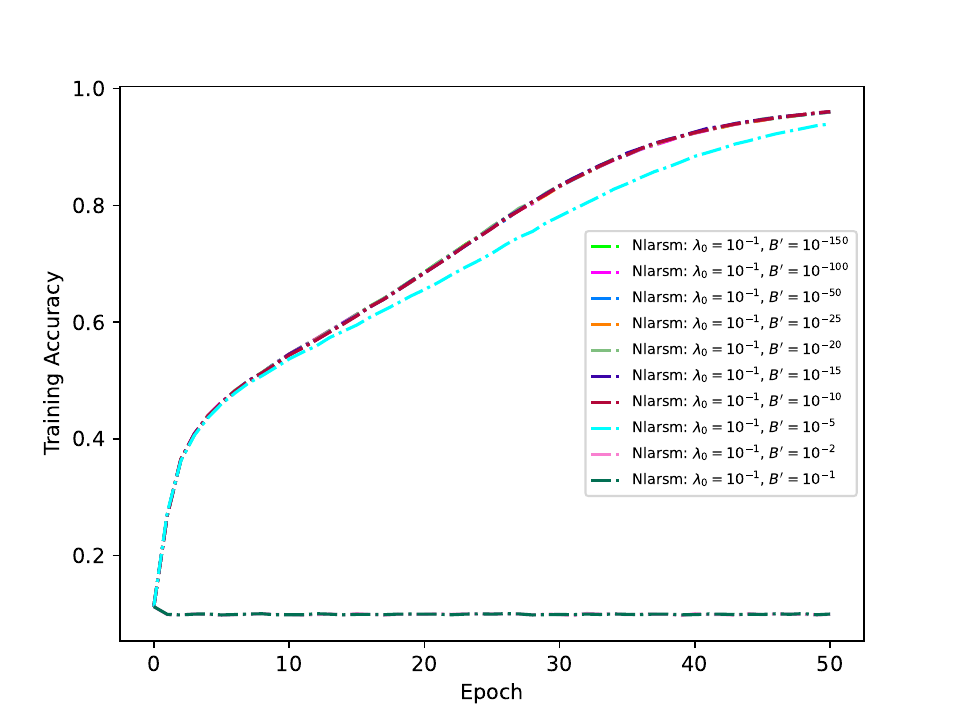}\hspace{-0.5mm}%
        \caption{\tiny MLP7h - CIFAR10}
    \end{subfigure}
    \hfill
    \begin{subfigure}[b]{0.22\textwidth}
        \centering
        \includegraphics[width=\textwidth]{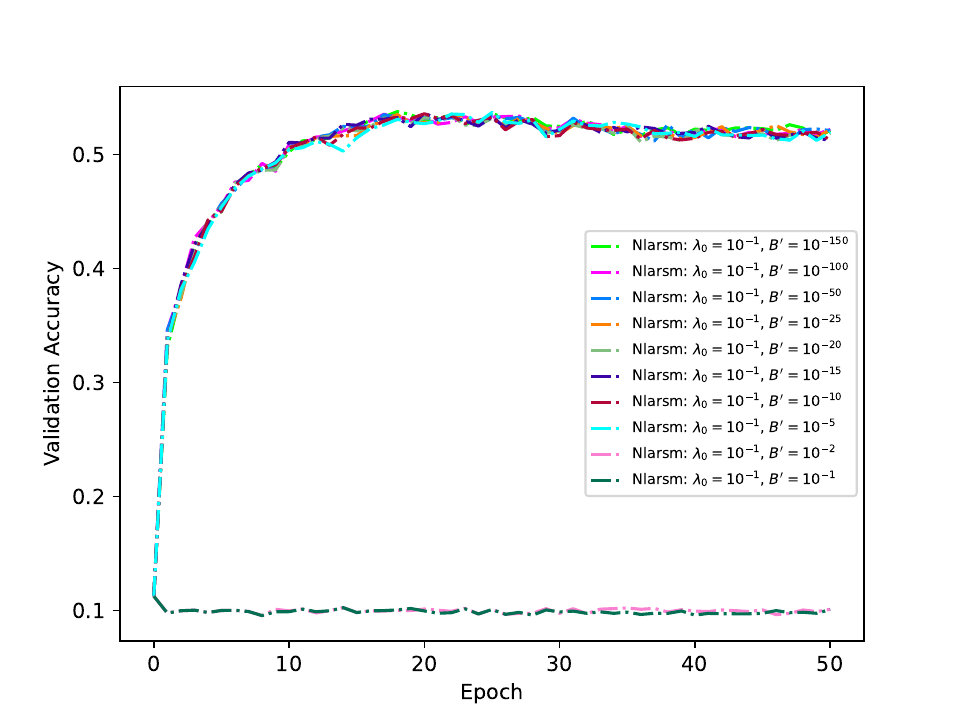}
        \caption{\tiny MLP7h - CIFAR10}
    \end{subfigure}
    \caption{Sensitivity analysis of Nlarsm, for a fixed $\lambda_0=0.1$, with respect to $B^\prime$ using MLP2h and MLP7h respectively, for the MNIST and CIFAR10 datasets.}
    \label{fig:sensitivity-four_figures}
\end{figure}

Figures \ref{fig:exp-mnist-mlp2h-nlar-sensitivity}, \ref{fig:exp-mnist-mlp7h-nlar-sensitivity}, \ref{fig:sensitivity-four_figures},  provide sensitivity analysis of Nlarsm and Nlarcm with respect to $k$, $c^\prime$, $c$, and $B^\prime$.

\subsection*{Sensitivity of Nlarsm and Nlarcm with respect to $\rho$}

\begin{figure}[htbp]
    \centering
    \begin{subfigure}[b]{0.22\textwidth}
        \centering
        \includegraphics[width=\textwidth]{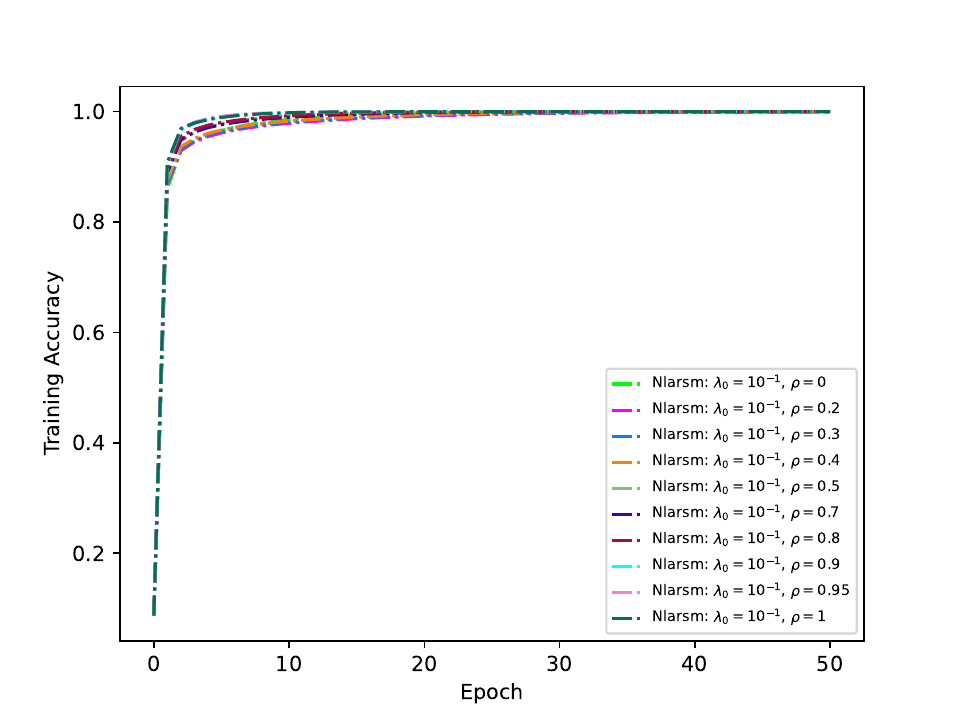}\hspace{-0.5mm}%
        \caption{\tiny MLP2h - MNIST}
    \end{subfigure}
    \hfill
    \begin{subfigure}[b]{0.22\textwidth}
        \centering
        \includegraphics[width=\textwidth]{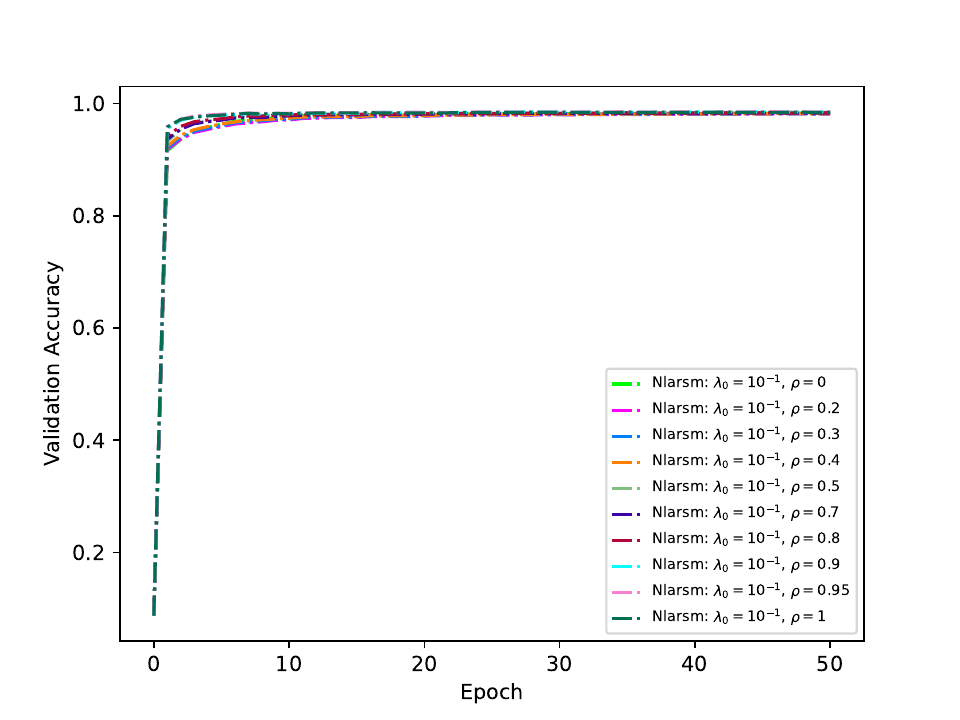}\hspace{-0.5mm}%
        \caption{\tiny MLP2h - MNIST}
    \end{subfigure}
    \hfill
    \begin{subfigure}[b]{0.22\textwidth}
        \centering
        \includegraphics[width=\textwidth]{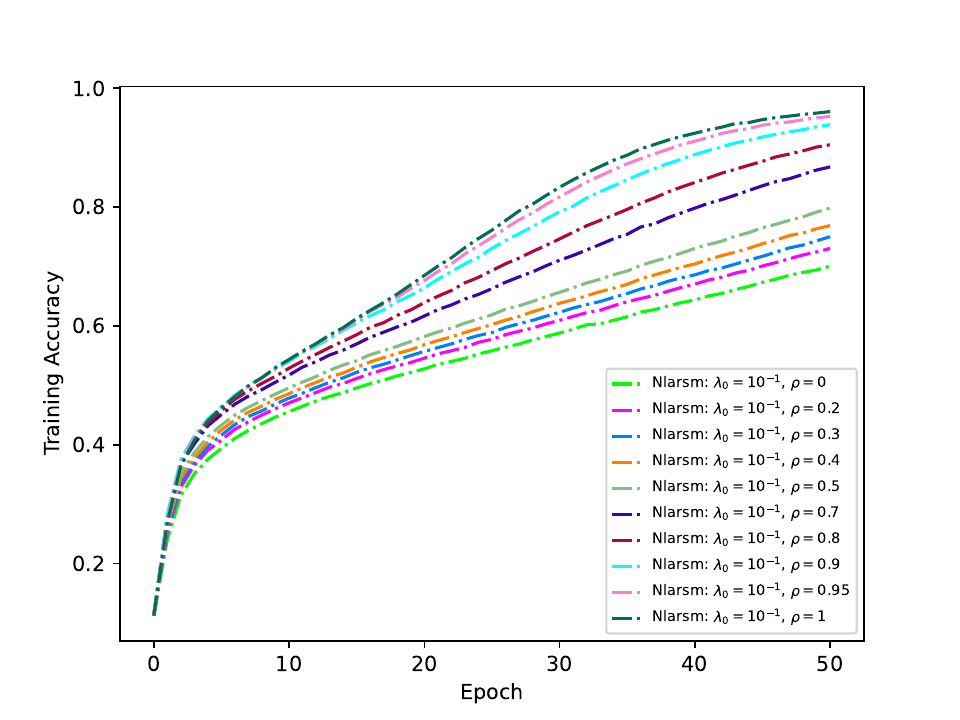}\hspace{-0.5mm}%
        \caption{\tiny MLP7h - CIFAR10}
    \end{subfigure}
    \hfill
    \begin{subfigure}[b]{0.22\textwidth}
        \centering
        \includegraphics[width=\textwidth]{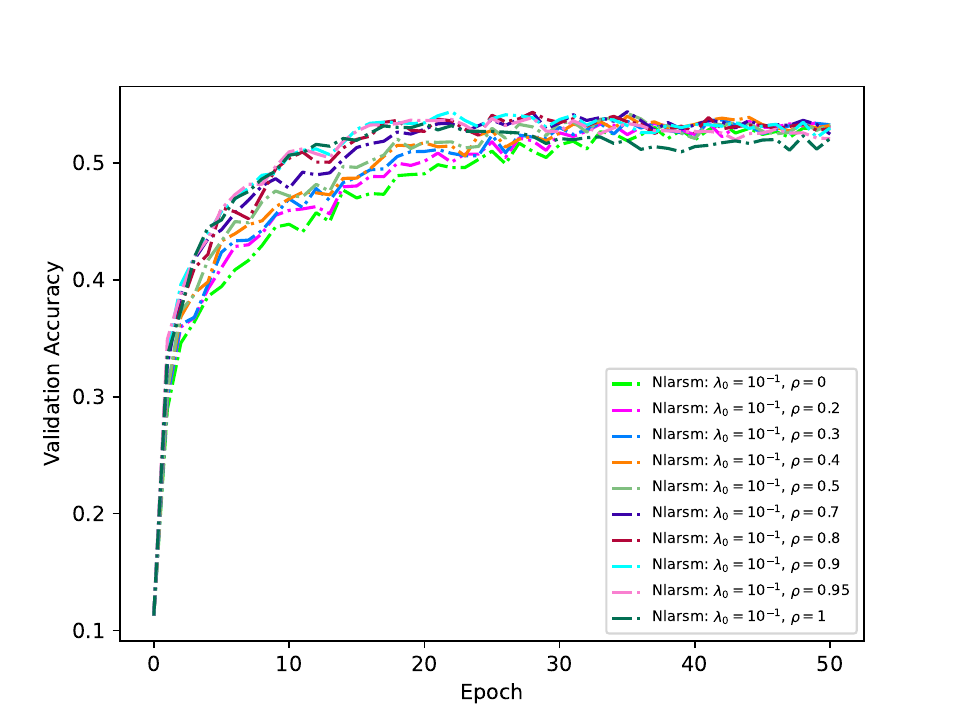}
        \caption{\tiny MLP7h - CIFAR10}
    \end{subfigure}
    \caption{Sensitivity analysis of Nlarsm, for a fixed $\lambda_0=0.1$, with respect to $\rho$ using MLP2h and MLP7h respectively, for the  MNIST and CIFAR10 datasets.}
    \label{fig:sensitivity-rho-nlarsm}
\end{figure}

\begin{figure}[htbp]
    \centering
    \begin{subfigure}[b]{0.22\textwidth}
        \centering
        \includegraphics[width=\textwidth]{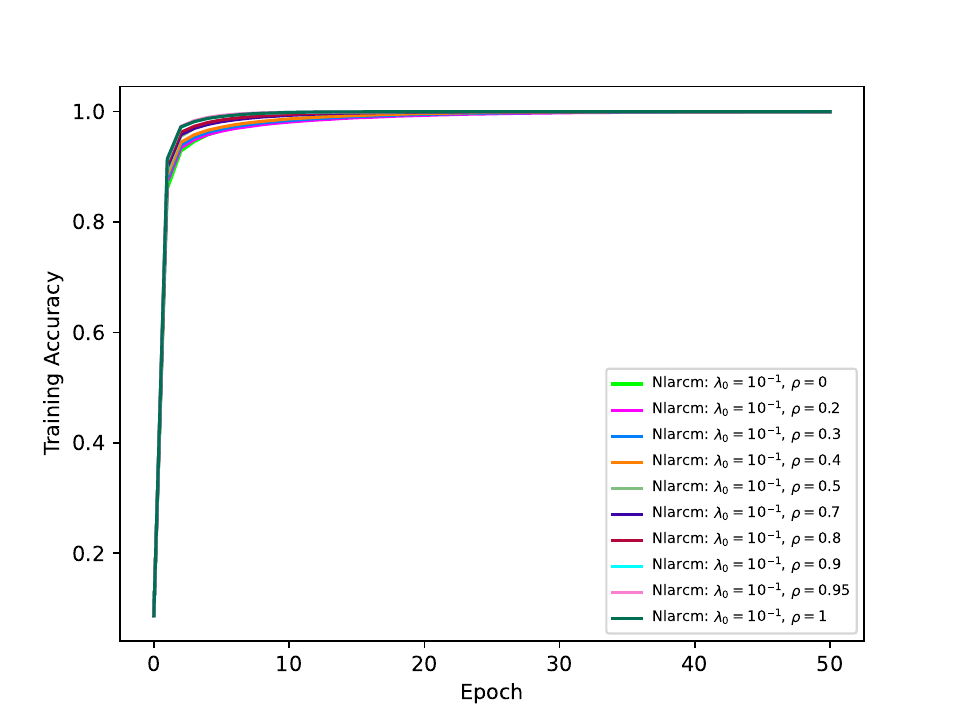}\hspace{-0.5mm}%
        \caption{\tiny MLP2h - MNIST}
    \end{subfigure}
    \hfill
    \begin{subfigure}[b]{0.22\textwidth}
        \centering
        \includegraphics[width=\textwidth]{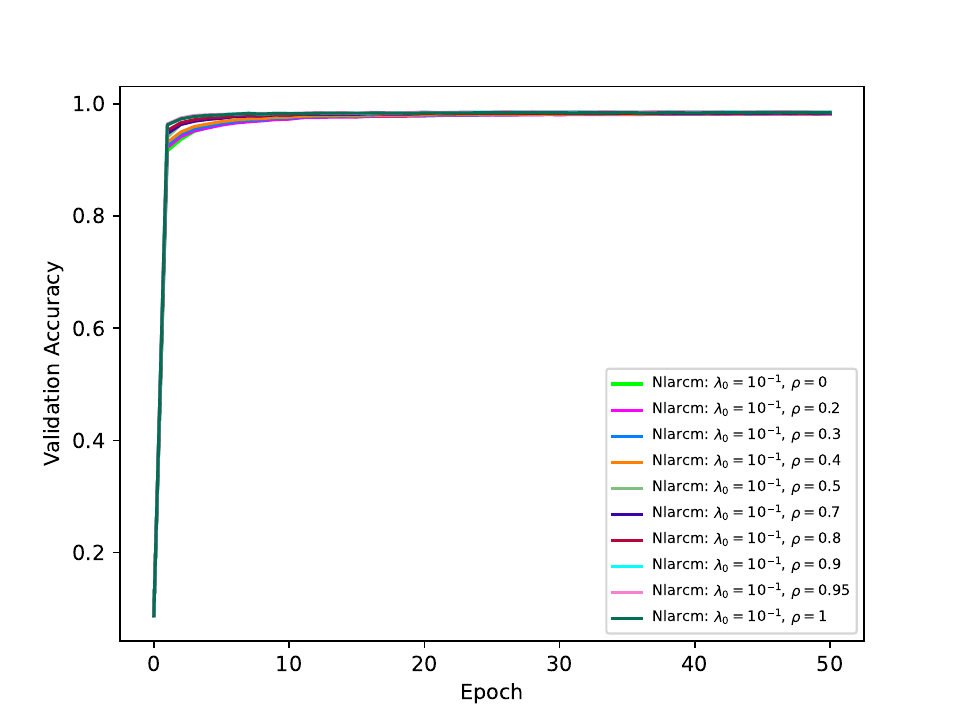}\hspace{-0.5mm}%
        \caption{\tiny MLP2h - MNIST}
    \end{subfigure}
    \hfill
    \begin{subfigure}[b]{0.22\textwidth}
        \centering
        \includegraphics[width=\textwidth]{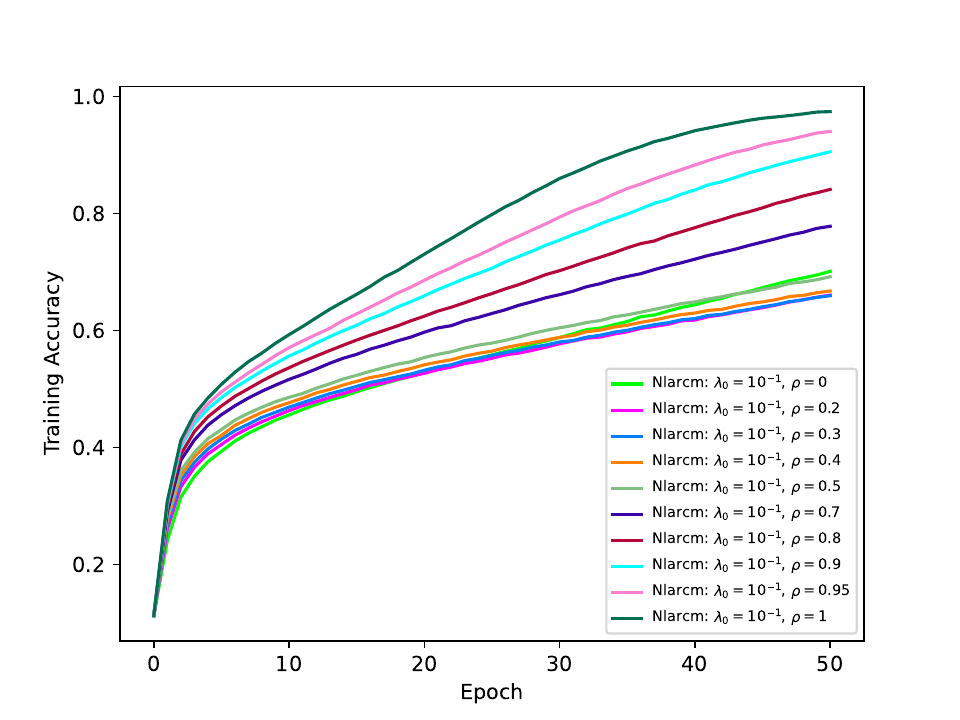}\hspace{-0.5mm}%
        \caption{\tiny MLP7h - CIFAR10}
    \end{subfigure}
    \hfill
    \begin{subfigure}[b]{0.22\textwidth}
        \centering
        \includegraphics[width=\textwidth]{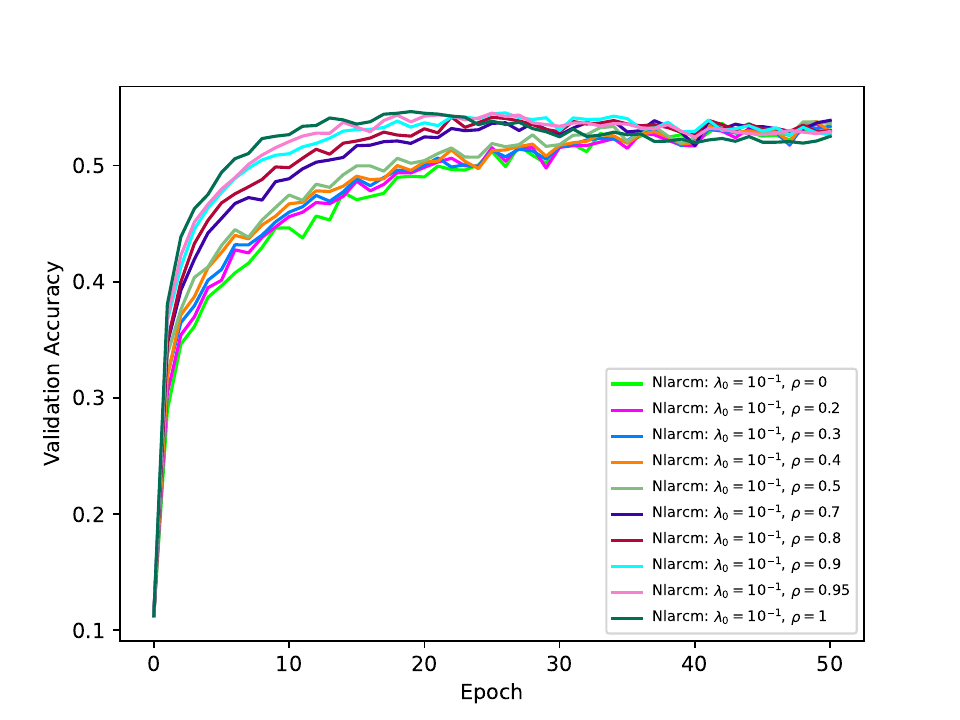}
        \caption{\tiny MLP7h - CIFAR10}
    \end{subfigure}
    \caption{Sensitivity analysis of Nlarcm, for a fixed $\lambda_0=0.1$, with respect to $\rho$ using MLP2h and MLP7h respectively, for the MNIST and CIFAR10 datasets.}
    \label{fig:sensitivity-rho-nlarcm}
\end{figure}

\begin{figure*}[!ht]
  \centering
  \includegraphics[width=0.22\linewidth, trim=0 10 30 0]{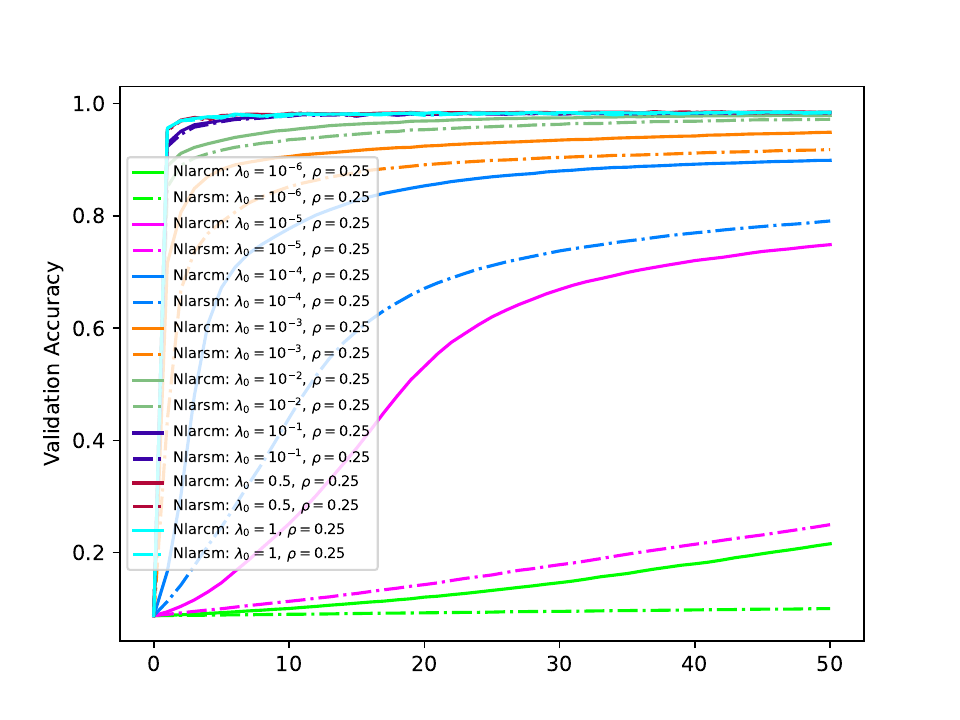}\hspace{-0.5mm}%
  \includegraphics[width=0.22\linewidth, trim=0 10 30 0]{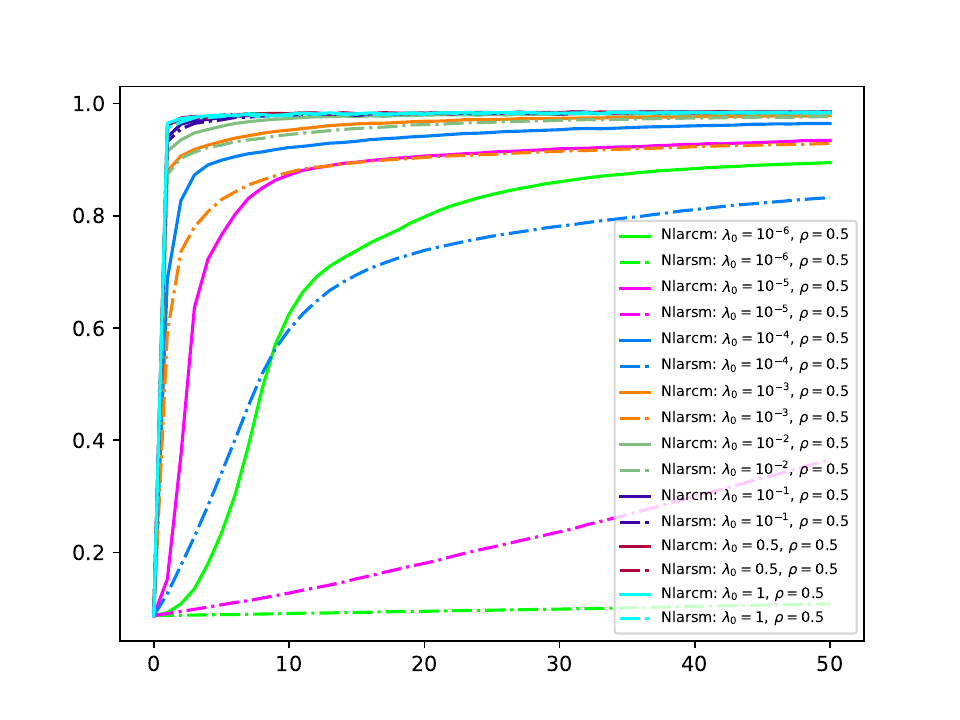}\hspace{-0.5mm}%
  \includegraphics[width=0.22\linewidth, trim=0 10 30 0]{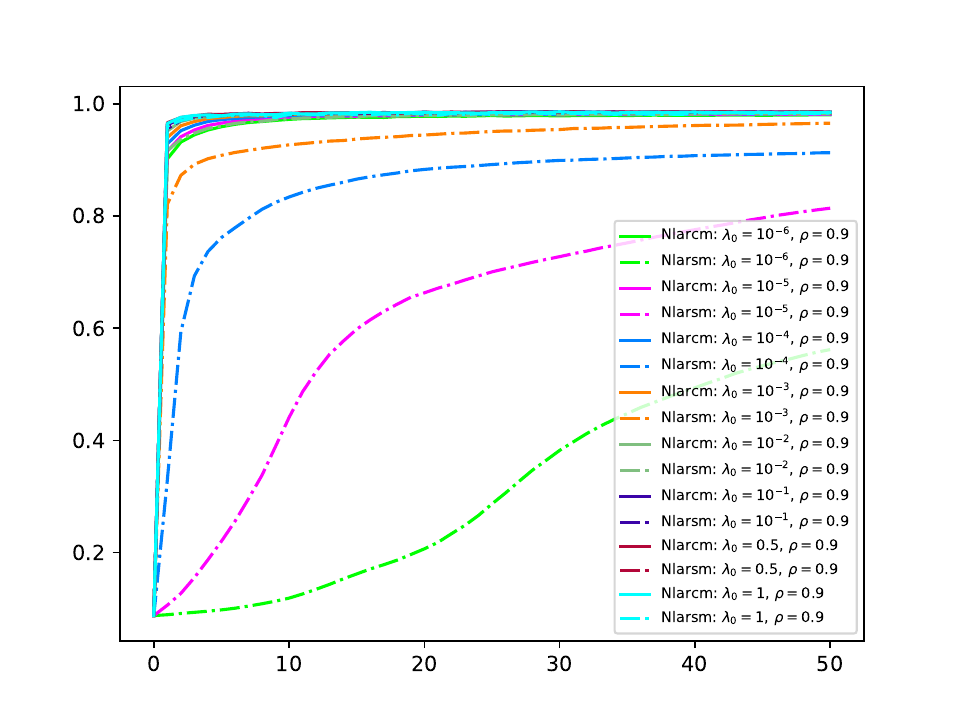}\hspace{-0.5mm}%
  \includegraphics[width=0.22\linewidth, trim=0 10 30 0]{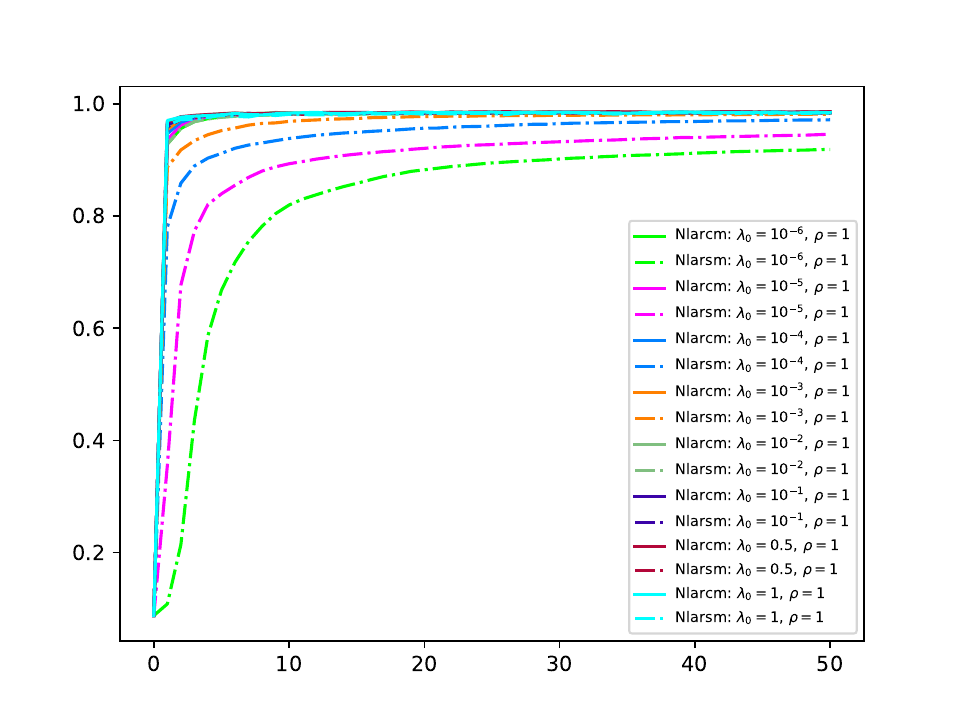}

  \includegraphics[width=0.22\linewidth, trim=0 10 30 0]{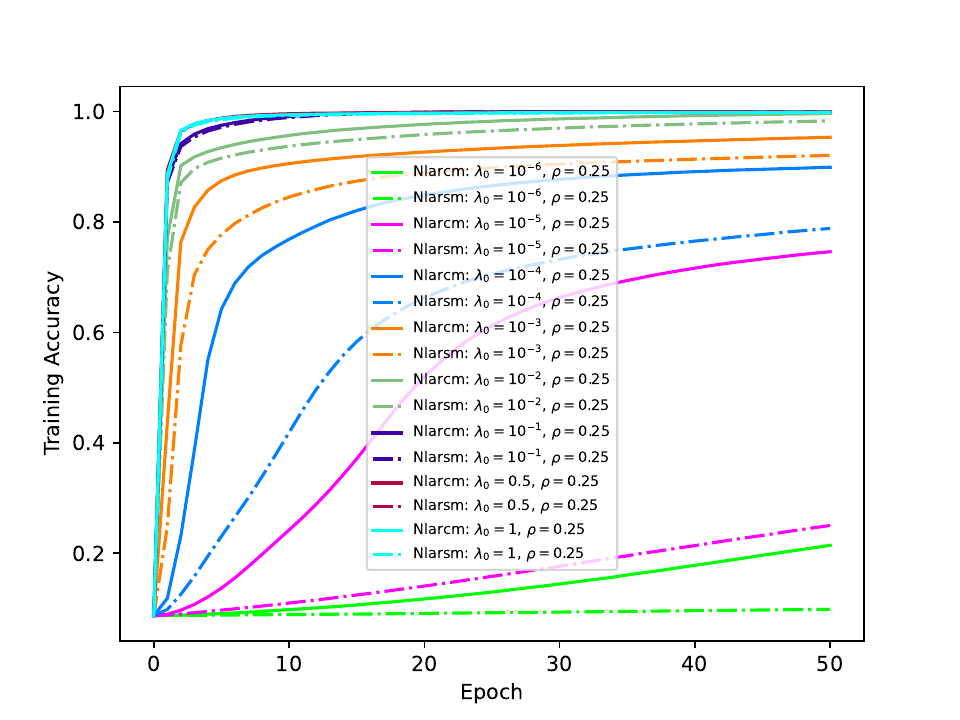}\hspace{-0.5mm}%
  \includegraphics[width=0.22\linewidth, trim=0 10 30 0]{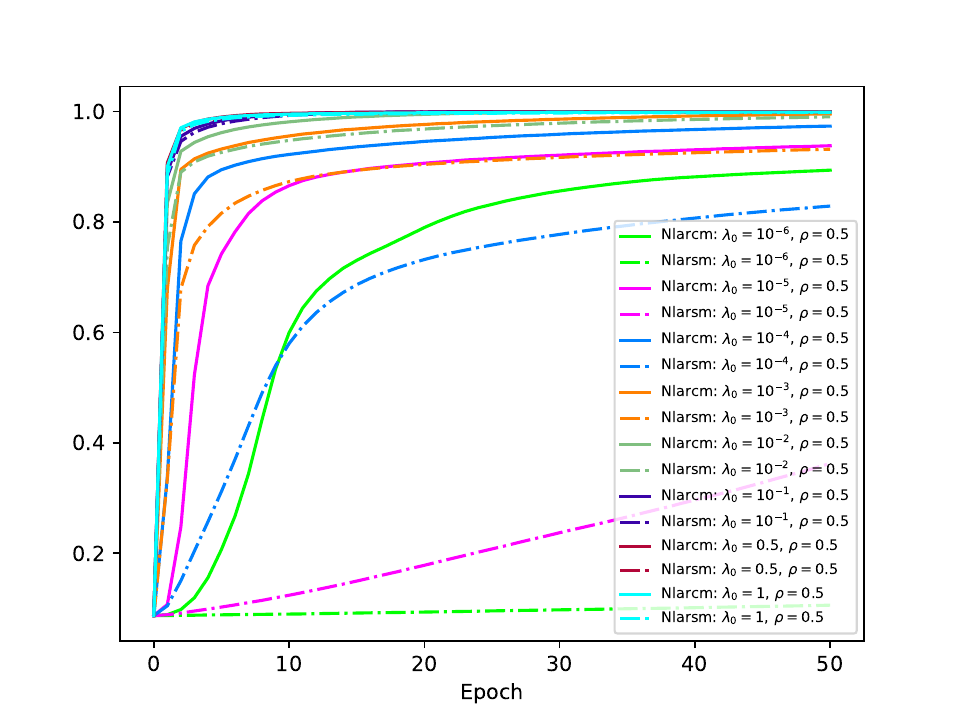}\hspace{-0.5mm}%
  \includegraphics[width=0.22\linewidth, trim=0 10 30 0]{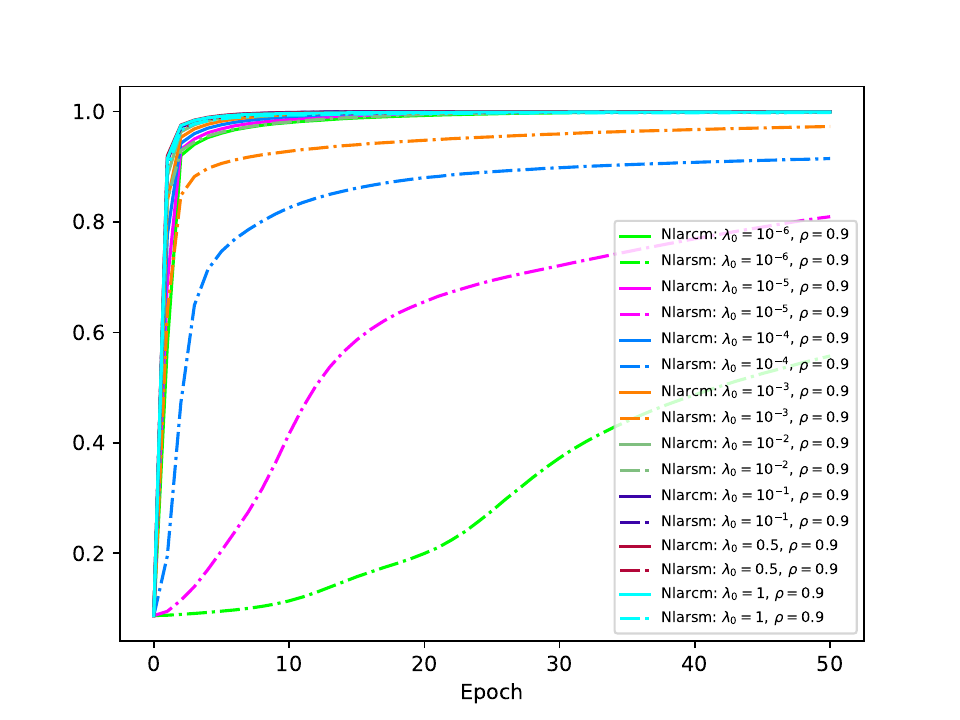}\hspace{-0.5mm}%
  \includegraphics[width=0.22\linewidth, trim=0 10 30 0]{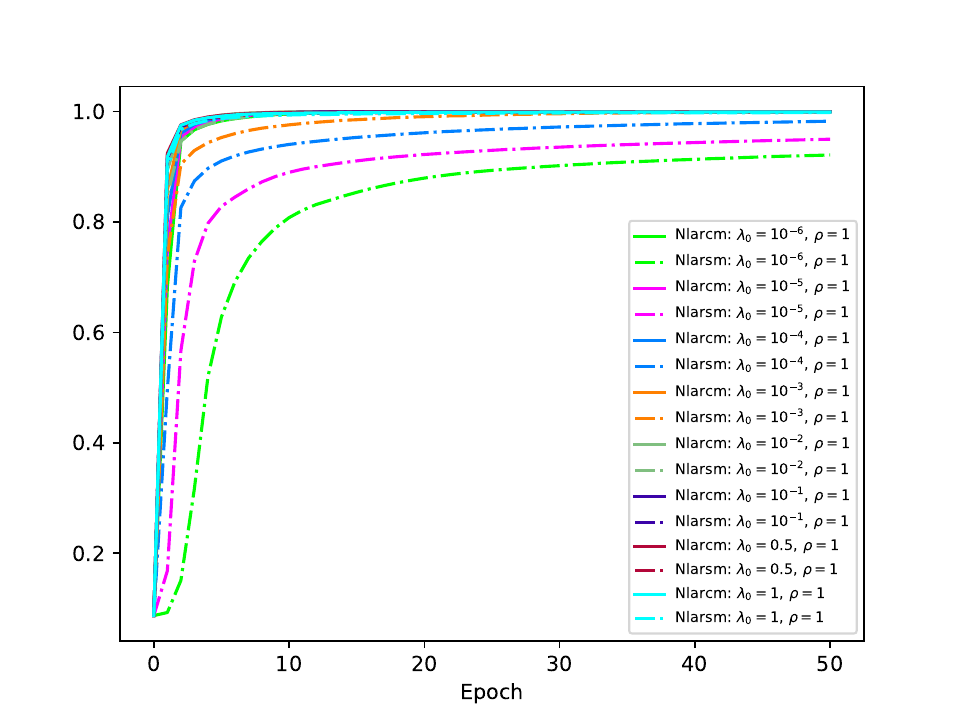}

    \vspace{1\baselineskip}
  \caption{MLP2h on the MNIST dataset: Performance comparison of Nlarsm and Nlarcm across varied momentum values and learning rates.}
  \vspace{1\baselineskip}
  \label{fig:exp-mnist-nlars-nlarc-different-rho}
\end{figure*}

\begin{figure*}[!ht]
  \centering
  \includegraphics[width=0.22\linewidth, trim=0 10 30 0]{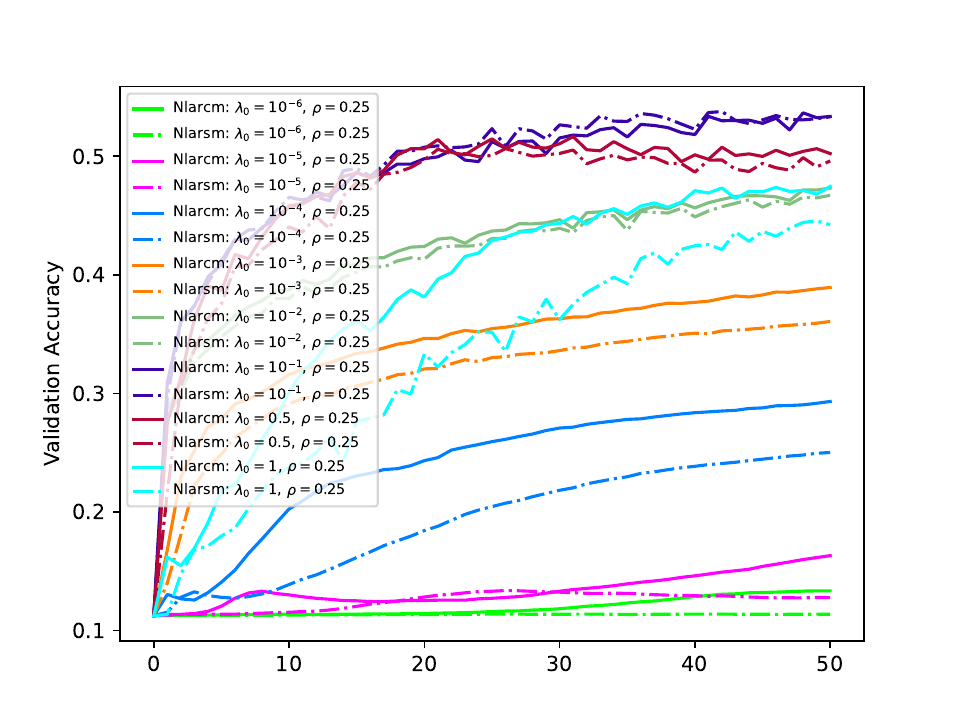}\hspace{-0.5mm}%
  \includegraphics[width=0.22\linewidth, trim=0 10 30 0]{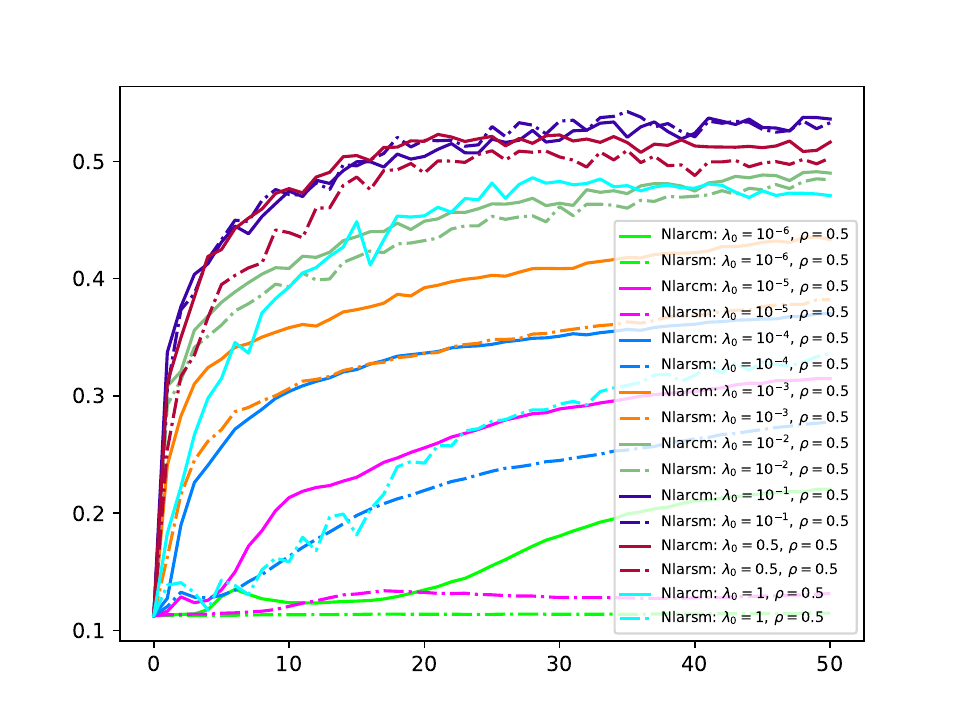}\hspace{-0.5mm}%
  \includegraphics[width=0.22\linewidth, trim=0 10 30 0]{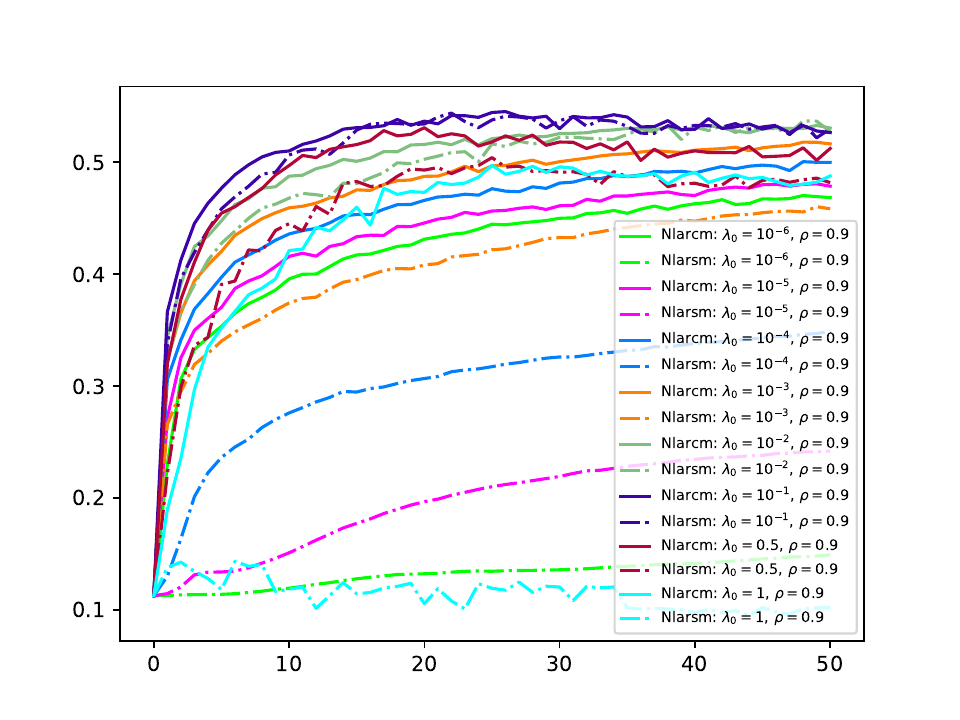}\hspace{-0.5mm}%
  \includegraphics[width=0.22\linewidth, trim=0 10 30 0]{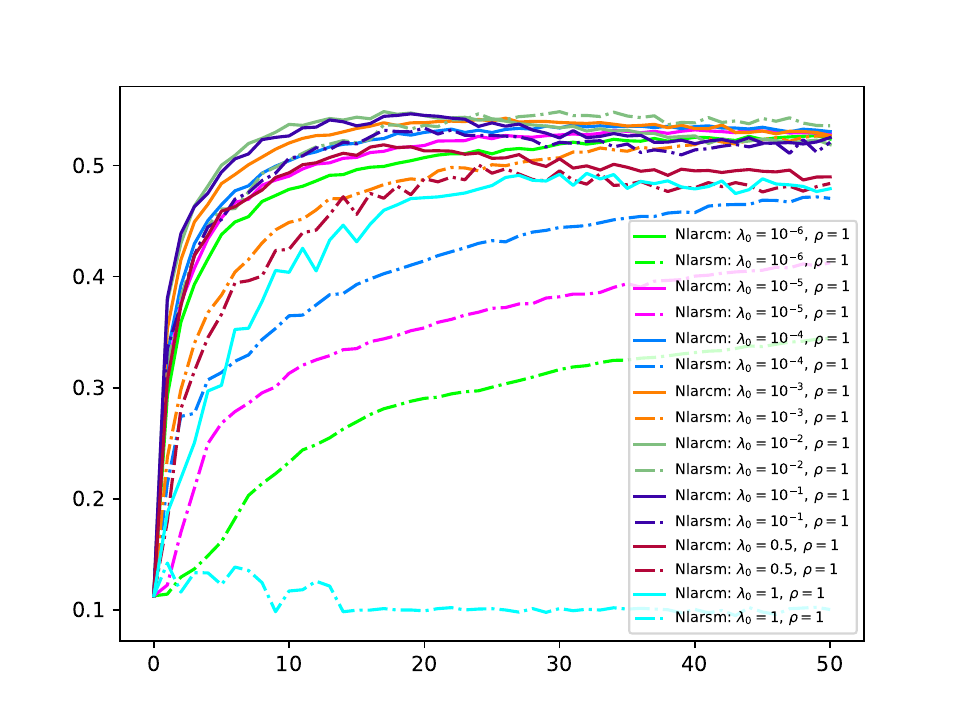}

  \includegraphics[width=0.22\linewidth, trim=0 10 30 0]{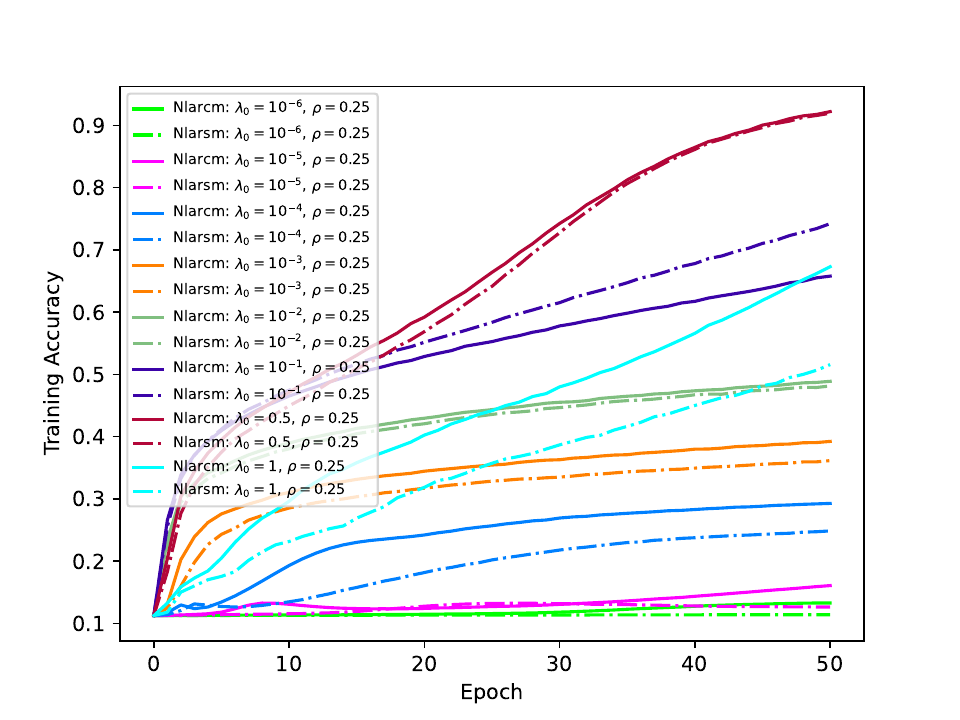}\hspace{-0.5mm}%
  \includegraphics[width=0.22\linewidth, trim=0 10 30 0]{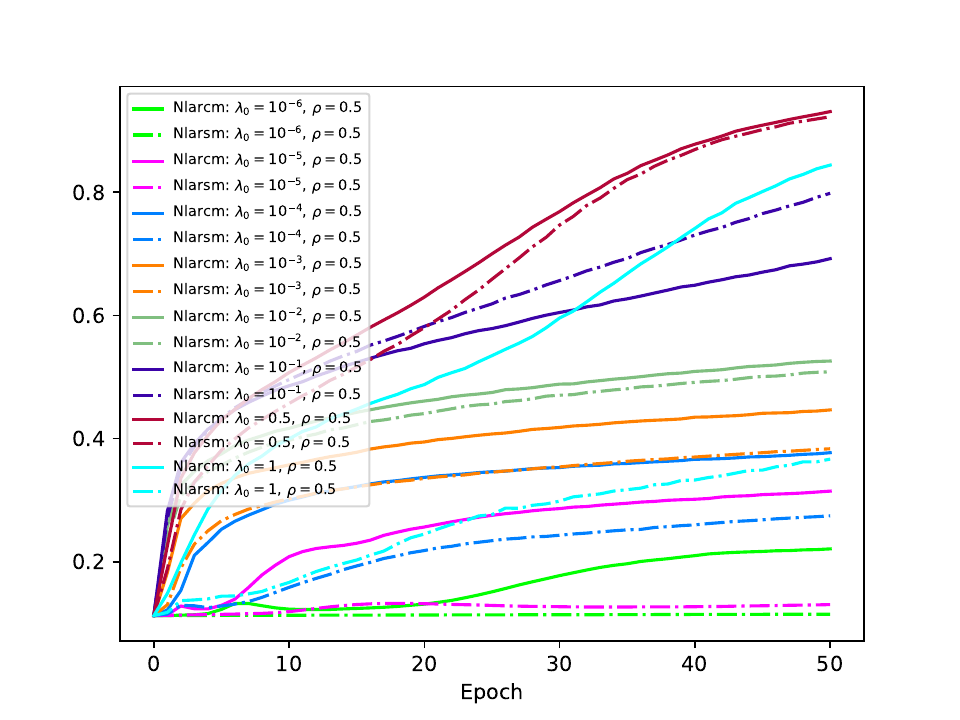}\hspace{-0.5mm}%
  \includegraphics[width=0.22\linewidth, trim=0 10 30 0]{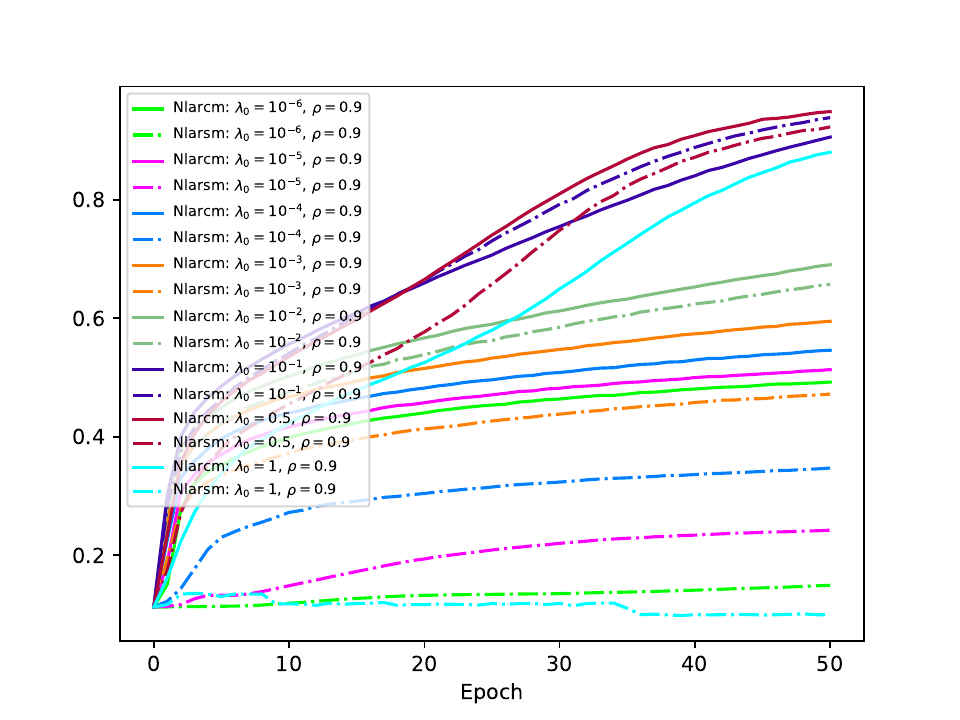}\hspace{-0.5mm}%
  \includegraphics[width=0.22\linewidth, trim=0 10 30 0]{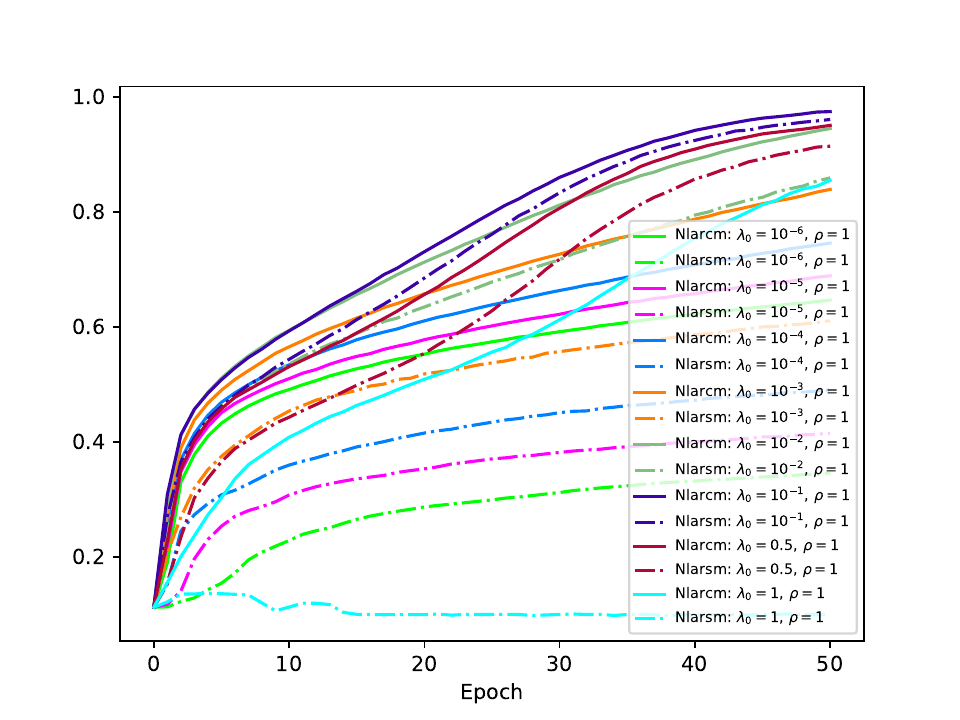}

    \vspace{1\baselineskip}
  \caption{MLP7h on the CIFAR10 dataset: Performance comparison of Nlarsm and Nlarcm across varied momentum values and learning rates.}
  \vspace{1\baselineskip}
  \label{fig:exp-cifar10-nlars-nlarc-different-rho}
\end{figure*}

Figures \ref{fig:sensitivity-rho-nlarsm} and \ref{fig:sensitivity-rho-nlarcm} show the performance of Nlarsm and Nlarcm under varying values of $\rho$ for a fixed value of $\lambda_0=0.1$. Figures \ref{fig:exp-mnist-nlars-nlarc-different-rho}, \ref{fig:exp-cifar10-nlars-nlarc-different-rho} display the performance of Nlarsm and Nlarcm under varying values of $\rho$ and $\lambda_0$.

